\documentclass[11pt,letterpaper]{article}

\usepackage[margin=1in]{geometry}
\usepackage[round,authoryear]{natbib}
\usepackage[T1]{fontenc}
\usepackage{times}
\usepackage{microtype}
\usepackage{amsmath,amssymb,amsthm,mathtools,bm,mathrsfs}
\usepackage{aliascnt}
\usepackage{booktabs,array,tabularx,multirow,longtable}
\usepackage{enumitem}
\usepackage{graphicx}
\usepackage{placeins}
\usepackage{xcolor}
\usepackage{url}
\usepackage{authblk}
\usepackage{algorithm}
\usepackage{algorithmic}
\usepackage{hyperref}
\usepackage{subcaption}
\usepackage[nameinlink,noabbrev,capitalise]{cleveref}
\hypersetup{colorlinks=true,linkcolor=blue!45!black,citecolor=blue!45!black,urlcolor=blue!45!black}
\newtheorem{theorem}{Theorem}[section]
\newaliascnt{proposition}{theorem}
\newtheorem{proposition}[proposition]{Proposition}
\aliascntresetthe{proposition}
\newaliascnt{lemma}{theorem}
\newtheorem{lemma}[lemma]{Lemma}
\aliascntresetthe{lemma}
\newaliascnt{corollary}{theorem}
\newtheorem{corollary}[corollary]{Corollary}
\aliascntresetthe{corollary}
\newaliascnt{assumption}{theorem}
\newtheorem{assumption}[assumption]{Assumption}
\aliascntresetthe{assumption}
\newaliascnt{condition}{theorem}

\aliascntresetthe{condition}
\theoremstyle{definition}
\newaliascnt{definition}{theorem}
\newtheorem{definition}[definition]{Definition}
\aliascntresetthe{definition}
\theoremstyle{remark}
\newaliascnt{remark}{theorem}
\newtheorem{remark}[remark]{Remark}
\aliascntresetthe{remark}
\crefname{theorem}{Theorem}{Theorems}
\Crefname{theorem}{Theorem}{Theorems}
\crefname{proposition}{Proposition}{Propositions}
\Crefname{proposition}{Proposition}{Propositions}
\crefname{lemma}{Lemma}{Lemmas}
\Crefname{lemma}{Lemma}{Lemmas}
\crefname{corollary}{Corollary}{Corollaries}
\Crefname{corollary}{Corollary}{Corollaries}
\crefname{assumption}{Assumption}{Assumptions}
\Crefname{assumption}{Assumption}{Assumptions}
\crefname{condition}{Condition}{Conditions}
\Crefname{condition}{Condition}{Conditions}
\crefname{definition}{Definition}{Definitions}
\Crefname{definition}{Definition}{Definitions}
\crefname{remark}{Remark}{Remarks}
\Crefname{remark}{Remark}{Remarks}
\newcommand{\E}{\mathbb{E}}
\newcommand{\R}{\mathbb{R}}

\newcommand{\cS}{\mathcal{S}}
\newcommand{\cA}{\mathcal{A}}

\newcommand{\cK}{\mathcal{K}}

\newcommand{\cG}{\mathcal{G}}

\newcommand{\KL}{D_{\mathrm{KL}}}
\newcommand{\op}{\mathrm{op}}
\newcommand{\Cov}{\operatorname{Cov}}
\newcommand{\Var}{\operatorname{Var}}
\newcommand{\argmin}{\operatorname*{arg\,min}}
\newcommand{\grad}{\nabla}
\newcommand{\wtO}{\widetilde{\mathcal O}}
\newcommand{\ip}[2]{\left\langle #1,#2\right\rangle}
\newcommand{\norm}[1]{\left\lVert #1\right\rVert}
\newcommand{\abs}[1]{\left\lvert #1\right\rvert}

\newcommand{\Id}{I}
\newcommand{\piRef}{\pi_{\mathrm{ref}}}
\newcommand{\eps}{\varepsilon}

\newcommand{\Fisher}{\mathcal F}
\newcommand{\dbar}{\overline d}

\setlist[itemize]{leftmargin=1.35em,itemsep=.15em,topsep=.25em}
\setlist[enumerate]{leftmargin=1.55em,itemsep=.15em,topsep=.25em}

\newcommand{\cO}{\mathcal{O}}

\newcommand{\Bgeom}{\mathcal B_{\mathrm{geom}}}

\usepackage{pifont}

\newcommand{\Prb}{\mathbb{P}}

\crefname{algorithm}{Algorithm}{Algorithms}
\Crefname{algorithm}{Algorithm}{Algorithms}

\newcommand{\AQ}{\mathcal A_Q}
\newcommand{\CritSet}{\mathcal K_Q}

\title{Improved Sample Complexity Analysis For First Order Bilevel Reinforcement Learning}
\author[1]{Mudit Gaur}
\author[2]{Naman Saxena}
\author[2]{Vaneet Aggarwal}
\affil[1]{Department of Electrical and Computer Engineering, Ohio State Univeristy}
\affil[2]{ Edwardson School of Industrial Engineering, Purdue University}
\date{}

\begin{document}
\maketitle
\begin{abstract}
Bilevel reinforcement learning optimizes an upper-level objective through
the policy response to a lower-level reinforcement learning problem.
Such problems arise when learning a reward model while adapting a policy
to that reward. However, a gap remains between sample-complexity
guarantees and practical actor--critic training. Tabular analyses restrict
applicability to finite spaces, while continuous-state analyses often
impose a uniform Polyak--\L{}ojasiewicz (PL) condition. We show that this
condition need not hold even for smooth positive policies with exactly
representable values. We develop a single-loop, fully first-order method
for continuous states and finite actions with specified multilayer neural
actors and a persistently trained nonlinear critic built from fixed
features. Under explicit initialization, regularity, feature-excitation,
and value-realizability conditions, fixed reference-KL regularization
yields local curvature, and we prove containment of the coupled iterates.
Shared trajectories control sampling noise, while paired critic analysis
controls normalized critic-error differences as the response perturbation
shrinks. A joint descent and tracking argument establishes an improved
$\widetilde{\mathcal O}(\varepsilon^{-2})$ sample complexity for
$O(\varepsilon)$ expected squared gradient norm of the upper objective
induced by the selected regularized policy response, including critic
training.

\end{abstract}

\section{Introduction}
\label{sec:main-introduction}
Bilevel reinforcement learning (BRL) couples reward learning with policy
adaptation: changing a reward changes both the responding policy and the
trajectories used to evaluate the upper objective. This dependence underlies
reward design and reinforcement learning from human feedback. Establishing
convergence requires justifying the policy-optimization geometry and
controlling value estimation while rewards, policies, and sampling
distributions evolve together.

Tabular analyses exploit finite policy and value representations
\citep{shen2025principled,yang2025bilevel}. In particular,
\citet{yang2025bilevel} use soft Bellman structure to obtain first-order
upper-gradient calculations for a global regularized response.
\citet{zeng2026hessianfree} establish a single-loop first-order tabular
actor--critic guarantee under regularization-dependent PL and additional
Hessian conditions. Extending these guarantees beyond finite state--action
representations requires controlling actor and critic approximation and
learning errors.

Beyond tabular formulations, PARL \citep{chakraborty2024parl} studies policy
alignment with parameterized policies and accounts for the dependence of
the upper sampling distribution on the lower response. Its formulation
accommodates continuous states, but its A-PARL algorithm uses implicit
hypergradient calculations involving mixed derivatives and an inverse
lower-level Hessian. It is therefore not fully first order, and its
convergence analysis assumes access to the required derivative oracles.
\citet{gaur2025sample} establish a first-order
$\widetilde{\mathcal O}(\varepsilon^{-3})$ sample-complexity bound for
continuous state--action spaces under a uniform lower-level
Polyak--\L{}ojasiewicz (PL) condition.

Uniform PL does not follow from smooth policy parameterization alone.
Our counterexample in \cref{prop:PL-counterexample} has smooth positive
action probabilities and exactly representable values, yet admits a
suboptimal stationary point, violating PL. This motivates specifying an
actor construction, deriving its required local geometry, and proving
that the learning dynamics remain in the region where that geometry holds. Additionally, PL conditions in general do not hold for arbitrary RL setups. Thus works which rely on PL assumptions are not applicable for arbitrary actor and critic classes.

Additionally, critic learning introduces a distinct difficulty. The value targets change
with both the reward and the policy, and the response correction uses
differences between nearby policies divided by their separation. Separate
bounds on the two critic errors do not directly control their difference
after this normalization as the separation shrinks. A coupled guarantee
must therefore establish persistent critic stability and track both the
individual value errors and their normalized endpoint differences. Thus we have the following question.

\emph{Can we obtain improved sample-complexity guarantees for single-loop,
first-order bilevel reinforcement learning beyond tabular settings,
using explicit neural actors and nonlinear critics, while deriving the
required local geometry and controlling their coupled learning errors?}

We answer this question in the affirmative and address these challenges with explicit tied multilayer actors and a
nonlinear critic built from fixed features. An initialization covariance
condition, architectural derivative bounds, and fixed reference-policy KL
regularization yield local curvature; a joint containment proof makes
this geometry valid throughout training with high probability. Shared
trajectories control the noise in the response correction. Under feature
excitation and realizability of the regularized value family, our paired
critic analysis establishes stability and tracking of the required
normalized errors. Residual-controlled actor solves complete the finite,
first-order implementation. Together, these ingredients yield
$\widetilde{\mathcal O}(\varepsilon^{-2})$ sample complexity for
$O(\varepsilon)$ expected squared upper-gradient norm, including persistent
critic training, preference sampling, and metric initialization. The
guarantee covers continuous states and finite actions.  under the stated
architecture, initialization, regularity, and value-representation
conditions.

\begin{itemize}
\item \textbf{Variance control for coupled response estimation.}
We estimate the policy-response correction by evaluating nearby
policies on shared, appropriately weighted trajectories.
This coupling keeps sampling variance controlled as the policies
being compared become closer, allowing the response approximation
to become more accurate without amplifying sampling noise.
Together with paired critic-error control and persistent
actor--critic tracking, this mechanism supports our improved
sample complexity relative to \citet{gaur2025sample}.

\item \textbf{Improved interaction complexity with single-loop first-order learning.}
We establish the $\widetilde{\mathcal O}(\varepsilon^{-2})$ interaction
bound while reward, actor, and critic updates proceed together. Shared
trajectory sampling controls response-estimation noise as the perturbation
shrinks, and a combined descent and tracking potential accounts for all
three learning processes and their sampling costs.

\item \textbf{Explicit actor--critic architectures with coupled learning guarantees.}
We specify multilayer neural actors and a nonlinear critic built
from fixed features. We show that smooth policies and exact value
realizability do not imply uniform PL, and derive local actor
curvature from initialization identifiability and fixed reference-KL
regularization. We prove containment of the coupled iterates and
stability of persistent critic learning, controlling critic errors as rewards, policies
change. A continuous-state construction verifies the actor
and critic conditions (Appendix~\ref{crex:example}).
\end{itemize}

\begin{table}[!htbp]
\centering
\footnotesize
\setlength{\tabcolsep}{3pt}
\renewcommand{\arraystretch}{1.12}
\begin{tabularx}{\linewidth}{@{}>{\raggedright\arraybackslash}p{0.22\linewidth}
 c >{\raggedright\arraybackslash}X c c c c@{}}
\toprule
Work & \shortstack{Continuous\\State Space} & Lower geometry & \shortstack{Nonlinear\\A--C analysis} & SL & FO & \shortstack{Sample\\complexity} \\
\midrule
\citet{chakraborty2024parl} & $\checkmark$ & PL; Hessian conditions & $\times$  & $\times$  & $\times$  & $\times$  \\
\citet{shen2025principled} & $\times$  & Policy-space structure & $\times$  & $\times$  & $\checkmark$ & $\times$  \\
\citet{yang2025bilevel}$^{\dagger}$ & $\times$  & Soft Bellman structure & $\times$  & $\times$  & $\checkmark$ & $\wtO(\eps^{-3.5})$ \\
\citet{gaur2025sample} & $\checkmark$  & Uniform PL & $\times$  & $\times$  & $\checkmark$ & $\wtO(\eps^{-3})$ \\
\citet{zeng2026hessianfree}$^{\ddagger}$ &$\times$ & Reg.-dependent PL; Hessian conditions & $\times$  & $\checkmark$ & $\checkmark$ & $\wtO(\eps^{-3})$ \\
This work & $\checkmark$ & Derived local curvature & $\checkmark$ & $\checkmark$ & $\checkmark$ & $\wtO(\eps^{-2})$ \\
\bottomrule
\end{tabularx}
\captionsetup{font=footnotesize}
\caption{Analyzed BRL guarantees. SL: single-loop
stochastic tracking; FO: first-order implementation. Nonlinear A--C analysis
requires explicit nonlinear actors and critics with coupled training and
critic-error control. Dashes mark unestablished guarantees or unreported
comparable sample complexitys. Rates use $O(\eps)$ squared stationarity
under distinct targets and assumptions.
$^{\dagger}$Stoc-SoBiRL; its deterministic model-based variant uses $O(1)$ inner updates.
$^{\ddagger}$Fixed regularization, Theorem~2 of version~6; best-iterate output.
Our uniform-output bound concerns a selected local regularized response,
globally optimal on $\mathcal D$ under additional PL
(\cref{cor:gpl-global-response}).
}
\label{tab:main-related-comparison}
\end{table}
\section{Problem Formulation}
\label{sec:main-setup}

\paragraph{MDP and sampling access.}
For a reward parameter $x\in\cO_x=\R^{d_x}$, let
$\mathcal M_x=(\cS,\cA,P,r_x,\rho,\gamma)$ be a discounted Markov
decision process: $\cS$ is compact and possibly continuous, $\cA$ is
finite, $P(ds'\mid s,a)$ is the transition law, $r_x(s,a)$ is the reward,
$\rho$ is the reset distribution, and $\gamma\in(0,1)$ is the discount.
Only the reward and upper sample loss depend explicitly on $x$.
We can independently reset from $\rho$, generate trajectories under
the current policies, and query feedback on trajectory tuples.
The reward and its first derivatives are evaluable; the transition
kernel need not be known.

\paragraph{Multilayer neural actor.}
Let $L$ be the hidden depth, $\mathbf m=(m_1,\ldots,m_L)$ the hidden
widths, and $\alpha\in\R^r$ the trainable coordinates. A scalar score
$F_{\mathbf m,\alpha}$ tilts a fixed strictly positive reference policy:
\begin{equation}
 \pi_{\mathbf m,\alpha}(a\mid s)
 =\frac{\piRef(a\mid s)e^{F_{\mathbf m,\alpha}(s,a)}}
 {\sum_{b\in\cA}\piRef(b\mid s)e^{F_{\mathbf m,\alpha}(s,b)}}.
 \label{eq:main-policy}
\end{equation}
Two coupled smooth multilayer networks define the score. Their weights
vary through a fixed map of the same $r$ coordinates around initialization;
we call this map the retained actor chart. Depth $L$ and dimension $r$
remain fixed as widths vary, and $F_{\mathbf m,0}=0$ gives
$\pi_{\mathbf m,0}=\piRef$. Appendix~\ref{app:actor-construction}
specifies the chart, normalization, inputs, and activations.

\paragraph{Lower and upper objectives.}
Let $\mathbb P_{\mathbf m,\alpha}$ be the trajectory law under reset
$s_0\sim\rho$, actions $a_t\sim\pi_{\mathbf m,\alpha}(\cdot\mid s_t)$,
and transitions $P$; write $\E_{\mathbf m,\alpha}$ for its expectation.
Fix tuple size $I$ and trajectory length $K$ independently of accuracy.
The tuple $\boldsymbol\xi=(\xi^1,\ldots,\xi^I)$ contains independent
length-$K$ state--action trajectories, with product law
$(\mathbb P^K_{\mathbf m,\alpha})^{\otimes I}$.
Feedback follows a fixed conditional law $\mathcal D(\cdot\mid\boldsymbol\xi)$,
independent of $(x,\alpha)$ given the tuple. For sample loss
$L(x;\boldsymbol\xi,Y)$ and fixed Kullback--Leibler (KL) regularization coefficient $\tau>0$, define
\begin{align}
 g_{\tau,\mathbf m}(x,\alpha)
 &=(1-\gamma)\E_{\mathbf m,\alpha}\sum_{t\ge0}\gamma^t
 \left[-r_x(s_t,a_t)+\tau\log
 \frac{\pi_{\mathbf m,\alpha}(a_t\mid s_t)}{\piRef(a_t\mid s_t)}\right],
 \label{eq:main-lower}\\
 f_{\mathbf m}(x,\alpha)
 &=\E_{\boldsymbol\xi\sim(\mathbb P^K_{\mathbf m,\alpha})^{\otimes I},\,
 Y\sim\mathcal D(\cdot\mid\boldsymbol\xi)}L(x;\boldsymbol\xi,Y).
 \label{eq:main-upper}
\end{align}
Thus lower minimization maximizes regularized return, while the upper
loss depends on the policy through its trajectory distribution.
For Bradley--Terry comparisons, $I=2$ and $Y\in\{0,1\}$, with $Y=1$
favoring $\xi^1$. Writing $(s_t^i,a_t^i)$ for trajectory $i$, the loss is
\begin{equation}
 \Delta_x=\sum_{t=0}^{K-1}[r_x(s_t^1,a_t^1)-r_x(s_t^2,a_t^2)],\qquad
 L(x;\boldsymbol\xi,Y)=\log(1+e^{\Delta_x})-Y\Delta_x.
 \label{eq:main-BT}
\end{equation}
Correct specification of the preference model is unnecessary
(Appendix~\ref{app:support-preference}).

\paragraph{Bilevel problem and stationarity target.}
Let $\Bgeom=\{\alpha:\|\alpha\|_2<R_{\rm geom}\}$.
For the fixed $\tau$ and radius supplied by Section~\ref{sec:main-result},
we study
\begin{equation}
 \begin{aligned}
 \min_{x\in\R^{d_x}}\quad
 &\Phi_{\tau,\mathbf m}(x)
 :=f_{\mathbf m}(x,\alpha_{0,\mathbf m}^*(x)),\\
 \text{where}\quad
 &\alpha_{0,\mathbf m}^*(x)\in
 \argmin_{\alpha\in {\Bgeom}}g_{\tau,\mathbf m}(x,\alpha).
 \end{aligned}
 \label{eq:main-local-response-root}
\end{equation}
The later conditions make this minimizer unique, interior, and smooth
in $x$, with $\nabla_\alpha g_{\tau,\mathbf m}(x,\alpha_{0,\mathbf m}^*)=0$.
The ball selects a local response in the full actor space $\R^r$;
subscript $0$ denotes the unperturbed response.
For accuracy $\eps>0$, our goal is an output satisfying
$\E\|\nabla\Phi_{\tau,\mathbf m}(x_{\rm out})\|^2=O(\eps)$.

\paragraph{Value function and nonlinear critic.}
Hereafter suppress actor widths, write $g=g_\tau$, and use $h_x,h_\alpha$
for partial gradients of a function $h$. Define
$q_{x,\alpha}=-r_x+\tau\log(\pi_\alpha/\piRef)$ and the unnormalized
discounted action value
\[
 Q^{\rm true}_{x,\alpha}(s,a)
 =\E_\alpha^{s,a}\!\left[\sum_{t\ge0}\gamma^tq_{x,\alpha}(s_t,a_t)\right],
\]
where $\E_\alpha^{s,a}$ starts the process at $(s_0,a_0)=(s,a)$.
For a fixed bounded feature map $\varphi_Q:\cS\times\cA\to\R^{d_Q}$,
coordinates $\theta\in\R^{d_Q}$, and scale $m_Q\ge1$, use
\begin{equation}
 Q_{m_Q,\theta}(s,a)=\sqrt{m_Q}\tanh\!\left(
             \varphi_Q(s,a)^\top\theta/\sqrt{m_Q}\right).
 \label{eq:main-trained-critic}
\end{equation}
The scale controls nonlinearity at fixed dimension. Section~\ref{sec:main-method}
trains this critic persistently; Appendix~\ref{crex:example} jointly
verifies the actor and critic assumptions in a continuous-state instance.

\paragraph{Response-chain correction.}
With response Jacobian $D_x\alpha_0^*(x)\in\R^{r\times d_x}$,
\begin{equation}
 \nabla\Phi_\tau(x)=f_x(x,\alpha_0^*(x))
 +D_x\alpha_0^*(x)^\top f_\alpha(x,\alpha_0^*(x)).
 \label{eq:main-chain-rule}
\end{equation}
The second term accounts for policy adaptation. The next section estimates
this total gradient.
\FloatBarrier

\section{Proposed Approach}
\label{sec:main-method}

\textbf{Response correction through paired actors.}
Following first-order penalty constructions \citep{kwon2024penalty}, let $\alpha_c^*(x)$ solve $g_\alpha+cf_\alpha=0$ locally in $\Bgeom$ for perturbation $0<c\le\min\{c_0,1\}$ and sufficiently small fixed $c_0$ (Section~\ref{sec:main-result}). Define
\begin{equation}
 S_c^*(x):=\frac{\alpha_c^*(x)-\alpha_0^*(x)}{c}.
 \label{eq:main-response-expansion}
\end{equation}
Persistent coordinates $U,S\in\R^r$ track $\alpha_0^*(x),S_c^*(x)$; the actors are $U,U+cS$. For $p=(x,U,S)$, set $a(p)=g_\alpha(x,U)$ and
\begin{align}
 R_c(p)&=\frac{g_\alpha(x,U+cS)-g_\alpha(x,U)}{c}+f_\alpha(x,U+cS),
 \label{eq:main-R}\\
 V_c(p)&=\frac{g_x(x,U+cS)-g_x(x,U)}{c}+f_x(x,U+cS).
 \label{eq:main-V}
\end{align}
At these response targets, $R_c=0$ and $V_c=\nabla\Phi_\tau(x)+O(c)$ under the conditions of Section~\ref{sec:main-result} (Appendix~\ref{app:main-step-actor}). Thus gradient differences recover policy adaptation using first-order evaluations.

\textbf{Paired critic and certified neighborhoods.}
We learn two critics together through coordinates $z=(u,v)$:
the base critic has parameter $u$, and the perturbed critic has
parameter $u+cv$. Thus $v$ tracks their separation relative to
the perturbation scale $c$.
To keep critic updates bounded, we project onto
$\mathcal K_Q$, the set of pairs satisfying
$\|u\|\le R_u$ and $\|v\|\le R_v$.
The projection $\Pi_{\mathcal K_Q}^{W}$ uses the weighted distance
induced by $\|z\|_W^2=M\|u\|^2+\|v\|^2$.
Appendix~\ref{app:main-step-td} constructs the positive radii
and weight.

For the actors, let $s>0$ bound the normalized separation and
$r_{\rm prox}>0$ be the radius of each local actor solve.
The analysis uses a prospective neighborhood, defined by
$\|S\|<s$ and
$\|U\|<R_{\rm geom}-c_0s-r_{\rm prox}/2$,
within which proposed updates can be analyzed.
The algorithm checks the stricter condition
\[
 \mathsf{Safe}(U,S):\qquad
 \|S\|<s/2,\qquad
 \|U\|<R_{\rm geom}-c_0s-r_{\rm prox}.
\]
The gap between these neighborhoods provides room for local
actor updates before the next safety check.
Appendix~\ref{app:main-step-containment} specifies the radii
and proves that the iterates satisfy this test throughout
training with high probability.


\textbf{Finite actor updates.}
Each actor update balances following an estimated descent direction
with keeping policy changes small.
Set the actor step $\eta_y=r_Q\beta$ and upper step
$\eta_x=q\eta_y$, where $\beta>0$ is the critic step and
$r_Q,q>0$ are fixed ratios.
Draw $n$ independent reset states from $\rho$, separately from
training data, and denote their empirical distribution by
$\widehat\rho_n$.
Choose $n$ for rejection budget $\delta_{\rm met}$ using
\eqref{klfin:eq:metric-size}, and check the initialization covariance
using~\eqref{klfin:eq:accept}.
This empirical certificate supplies the local KL geometry needed
for the updates; rejection returns $x_0$.
Store the accepted states and reuse them throughout training
for both actors.

For a frozen actor parameter $b$ and estimated gradient $h$, define
\begin{equation}
 \ell_{b,h}(\alpha)=\frac1n\sum_{i=1}^n
 \KL\bigl(\pi_b(\cdot\mid s_i)\|\pi_\alpha(\cdot\mid s_i)\bigr)
 +\eta_y h^\top(\alpha-b).
 \label{eq:main-finite-KL}
\end{equation}
The KL term discourages large policy changes on the stored states,
while the linear term encourages descent along $-h$.
The map $\widetilde{\mathsf M}(b,h)$ minimizes this objective
approximately by projected gradient descent, starting at $b$
and remaining within distance $r_{\rm prox}$.
Under the certified conditions, the stopping rule in
\cref{klfin:prop:solver} guarantees distance at most
$c\eta_y^2$ from the exact minimizer within $J_{\max}$ updates.
These numerical steps use stored states and finite action sums,
requiring no fresh environment interactions.
Appendix~\ref{app:support-prox} gives the implementation details
and constants.

\textbf{Sampling routines and timing.}
At each iteration, all sampling routines use frozen pre-update
parameters; the rollout cap $H$ is distinct from the preference
length $K$.
$\mathsf{LowerPair}$ uses the paired critics and preference tuples
to estimate $(a,R_c)$, with
$\widehat C=\widehat a+c\widehat R$ exactly.
$\mathsf{OuterPair}$ estimates $V_c$ using reward derivatives
and separate preference tuples.
$\mathsf{TDPair}$ estimates the paired temporal-difference (TD)
field in~\eqref{eq:algorithm-TD}, reusing each base-policy
transition for both endpoint residuals.
The sampling blocks are conditionally independent given the
pre-draw history, which includes the metric sample.
Preference tuples and labels come from the old perturbed actor,
and the updated critic enters the next iteration.
Appendix~\ref{app:support-sampling} specifies the estimators and
their interaction costs.
Algorithm~\ref{alg:deep} gives a detailed expansion of one
iteration of Algorithm~\ref{alg:main-trained}, spelling out the
sampling batches, estimator construction, and parameter updates.

\begin{algorithm}[!t]
\caption{Single-loop bilevel RL with a persistent paired critic}
\label{alg:main-trained}
\small
\begin{algorithmic}[1]
\REQUIRE Initial $x_0\in\R^{d_x}$; iteration count $T$; $c,\tau,m_Q,H,\delta_{\rm met}$; batch sizes; step sizes
 $\beta,\eta_y=r_Q\beta,\eta_x=q\eta_y$; the specified geometry and radii.
\STATE Fix the chart; draw $n$ independent reset states to form $\widehat\rho_n$ and certify it using \eqref{klfin:eq:metric-size} and \eqref{klfin:eq:accept}.
\STATE Set $U_0=S_0=0$ and deterministic $z_0\in\mathcal K_Q$; \textbf{return} $x_0$ if the metric certificate or $\mathsf{Safe}(U_0,S_0)$ fails.
\FOR{$t=0,\ldots,T-1$}
 \STATE Freeze $p_t=(x_t,U_t,S_t)$ and $z_t$.
 \STATE $(\widehat a_t,\widehat R_t)\gets\mathsf{LowerPair}(p_t,z_t;H)$;
 $\widehat C_t\gets\widehat a_t+c\widehat R_t$ exactly.
 \STATE $U_{t+1}\gets\widetilde{\mathsf M}(U_t,\widehat a_t)$;
 $\alpha^c_{t+1}\gets\widetilde{\mathsf M}(U_t+cS_t,\widehat C_t)$.
 \STATE $S_{t+1}\gets(\alpha^c_{t+1}-U_{t+1})/c$.
 \STATE $\widehat V_t\gets\mathsf{OuterPair}(p_t;H)$;
 $x_{t+1}\gets x_t-\eta_x\widehat V_t$.
 \STATE $\widehat F_t\gets\mathsf{TDPair}(p_t,z_t;H)$;
 $z_{t+1}\gets\Pi_{\mathcal K_Q}^{W}(z_t-\beta\widehat F_t)$.
 \STATE \textbf{return} $x_0$ if $\mathsf{Safe}(U_{t+1},S_{t+1})$ fails.
\ENDFOR
\STATE Draw an independent uniform index $J\in\{0,\ldots,T-1\}$; \textbf{return} $x_J$.
\end{algorithmic}
\end{algorithm}

\textbf{Why shared samples matter.}
The response correction divides differences between nearby policy
evaluations by $c$, which can amplify noise from independent
endpoint estimates. Shared, appropriately weighted trajectories
make the centered endpoint difference shrink with $c$, keeping
the normalized sampling variance controlled as the deterministic
approximation bias decreases. Steps~3 and~6 of the proof outline in
Section~\ref{sec:main-result} establish this variance
control and explain its role in the sample complexity.
\section{Theoretical Analysis}
\label{sec:main-result}
Fix $A>0$, $\overline B_A=\{\alpha\in\R^r:\|\alpha\|_2\le A\}$, and $0<R_{\rm geom}\le A/2$. Derivatives use multilinear operator norms; order zero denotes the function. Bounds are uniform in admissible actor widths. After fixing the construction, global constants are independent of $x,c,m_Q$ and accuracy unless stated otherwise. First we give the needed assumption as follows.

\begin{assumption}[A1: primitive local regularity]
\label{ass:regularity}\label{ass:main-primitive}
For every compact $Q\Subset\cO_x$ and $\alpha\in\overline B_A$, require:
\begin{enumerate}[label=(\roman*),itemsep=0pt,topsep=2pt]
\item \textbf{Reward:} $r_x(s,a)$ is $C^4$ in $x$, with $\sup_{x\in Q,s,a}\|D_x^k r_x(s,a)\|\le C_{r,k}(Q)$ for $0\le k\le4$.
\item \textbf{Upper sample loss:} $L(x;\boldsymbol\xi,Y)$ is $C^4$ in $x$, with $\sup_{x\in Q,\boldsymbol\xi,Y}\|D_x^kL\|\le C_{L,k}(Q)$ for $0\le k\le4$.
\item \textbf{Actor score:} the normalized tied construction satisfies $\sup_{\mathbf m,\alpha\in\overline B_A,s,a}\|D_\alpha^kF_{\mathbf m,\alpha}\|\le C_{F,k}(A)$ for $1\le k\le4$.
\end{enumerate}
All envelopes are finite and uniform in sample arguments and admissible widths; state or action differentiability is unnecessary.
\end{assumption}
\noindent\emph{Context.} Related regularity appears in \citet[Assumption~2]{marbach2001} and \citet[Assumptions~2 and~7(2)]{kwon2024penalty}. Fourth-order bounds control paired KL-update remainders; the specified architecture verifies item~(iii).

\begin{assumption}[A2: initialization identifiability]
\label{ass:gap}\label{ass:main-population}
In the retained $r$-dimensional chart, let $Z_{\mathbf m,0}=D_\alpha F_{\mathbf m,0}$. For a width-independent $\lambda_{\rm eff,L}>0$, require
\begin{equation}
 \cG_{\mathbf m,0}:=\E_{s\sim\rho}
 \Cov_{a\sim\piRef(\cdot\mid s)}(Z_{\mathbf m,0}(s,a))
 \succeq\lambda_{\rm eff,L}I_r.
 \label{eq:main-population-gap}
\end{equation}
\end{assumption}
\noindent\emph{Context, verification, and derived geometry.}
Unlike the domain-wide Fisher conditions of \citet[Assumption~2.1]{liu2020} and \citet[Assumption~3]{fatkhullin2023}, A2 concerns only initialization. It admits a finite-sample certificate for an independently fixed chart (\cref{red:prop:pilot-certification}) and analytical verification in Appendix~\ref{crex:example}. Architectural derivative bounds and global regularity in G below propagate this gap locally. Sufficiently strong fixed reference-KL regularization then gives
\begin{equation}
 \nabla_\alpha^2[g_{\tau,\mathbf m}(x,\alpha)+cf_{\mathbf m}(x,\alpha)]
 \succeq\mu_\tau I_r,
 \label{eq:main-derived-actor-curvature}
\end{equation}
uniformly over $x$, $\alpha\in\Bgeom$, and $0\le c\le c_0$, for $\mu_\tau>0$ (\cref{glob:geometry}). This yields unique smooth interior responses; under C1--C2 and the stated schedule, containment of the coupled iterates is proved (\cref{critic:containment}).


\paragraph{Initialization identifiability and derived geometry.}
\label{ex:main-PL}
\citet{gaur2025sample} imposes uniform lower-level PL, which
need not hold even for smooth positive policies
(\cref{prop:PL-counterexample}).
Our analysis instead starts from the initialization covariance
condition in Assumption~\ref{ass:gap}, verified analytically
for our explicit construction in Appendix~\ref{crex:example}.
Together with regularity and fixed reference-KL regularization,
this condition yields local response curvature
(\cref{glob:geometry}).
The coupled-update analysis then establishes containment
within this region (\cref{critic:containment}).

\begin{assumption}[G: global upper regularity for preference losses]
\label{ass:global-upper}
Use Section~\ref{sec:main-setup}'s fixed-dynamics, finite-action reset setting and objective~\eqref{eq:main-upper} on $\cO_x=\R^{d_x}$, with fixed tuple size $I$ and trajectory length $K$. Tuples and labels are freshly sampleable. The conditional feedback law is independent of $(x,\alpha)$ given the trajectories, and $L$ has no additional explicit actor dependence. For finite constants, uniformly in $x$ and sample arguments,
\begin{equation}
 \begin{array}{lll}
 |r_x|\le B_r,&\|\nabla_xr_x\|\le G_r,&\|\nabla_x^2r_x\|_{\op}\le L_r,\\
 |L|\le B_L,&\|\nabla_xL\|\le G_L,&\|\nabla_x^2L\|_{\op}\le L_L.
 \end{array}
 \label{eq:main-global-envelopes}
\end{equation}
These bounds hold for each realized label; A1's local $C^4$ regularity is retained.
\end{assumption}
\noindent\emph{Context and role.}
Related global reward and loss envelopes appear in
\citet[Assumptions~7.2--7.3]{yang2025bilevel}.
For Bradley--Terry, the reward bounds imply the loss envelopes
(\cref{glob:BT}).
These bounds remain uniform as $x$ changes. Together with A1--A2,
they allow sufficiently strong fixed reference-KL regularization
to yield actor curvature uniformly in $x$, without restricting
the upper iterates to a prescribed compact set.

\paragraph{Critic assumptions.}
The next two assumptions concern feature excitation and value
representation. Let $\nu_U$ denote the normalized discounted
state--action occupancy under $\pi_U$ with reset distribution
$\rho$. For a state--action function $h$, define
$K_\alpha h(s,a)=\E[h(s',a')\mid s,a]$, where
$s'\sim P(\cdot\mid s,a)$ and
$a'\sim\pi_\alpha(\cdot\mid s')$.
In C1--C2, $(U,S)$ ranges over the prospective neighborhood
in Section~\ref{sec:main-method}; C2 covers every interpolation
$\alpha=U+tS$, $0\le t\le c_0$, including the prospective
update segments of Appendix~\ref{app:main-step-containment}.

\begin{assumption}[C1: fixed bounded critic features and excitation]
\label{ass:critic-features}
The map $\varphi_Q:\cS\times\cA\to\R^{d_Q}$ and dimension $d_Q$ are fixed independently of $x$, actors, $m_Q$, and accuracy, with $\sup_{s,a}\|\varphi_Q(s,a)\|\le F_Q<\infty$. Every base coordinate $U$ in the stated neighborhood satisfies $\E_{\nu_U}[\varphi_Q\varphi_Q^\top]\succeq\lambda_QI_{d_Q}$ for a uniform $\lambda_Q>0$.
\end{assumption}
\noindent\emph{Context and role.}
C1 rules out unidentifiable critic coefficient directions under
discounted reset occupancy, supporting stable TD learning.
Related feature conditions appear in linear TD
\citep[Section~2]{bhandari2018td} and neural TD
\citep[Assumption~3.1]{ke2024td}.

\begin{assumption}[C2: realizability of the regularized lower value]
\label{ass:critic-realizability}
Choose sufficiently large finite $\tau$ for actor geometry and initialization as in \cref{glob:geometry,critic:containment}, then keep it fixed before selecting critic constants. For every $x\in\R^{d_x}$ and actor $\alpha$ specified above, require
\begin{equation}
 Q^{\mathrm{true}}_{x,\alpha}
 =(\mathrm{Id}-\gamma K_\alpha)^{-1}q_{x,\alpha}
 \in\operatorname{span}\varphi_Q,
 \label{eq:regularized-value}
\end{equation}
where $q_{x,\alpha},Q^{\rm true}_{x,\alpha}$ are in Section~\ref{sec:main-setup}, $\mathrm{Id}$ is the identity operator, and $\operatorname{span}\varphi_Q=\{\varphi_Q^\top b:b\in\R^{d_Q}\}$.
\end{assumption}
\noindent\emph{Context and role.}
C2 requires the fixed feature span to represent the regularized
value family, including its actor-dependent KL cost.
Related representation conditions appear in
\citet[Assumption~A and Proposition~2.3]{jin2020linear};
neural TD retians projection error
\citep[Theorem~3.7]{ke2024td}.

\paragraph{Paired accuracy of the nonlinear critic.}
At the paired population TD target $(u^*,v^*)=Z(p)$
from~\eqref{critic:Z}, with $p=(x,U,S)$, define the value errors
$e_0^*(p)=Q_{m_Q,u^*}-Q^{\rm true}_{x,U}$ and
$e_c^*(p)=Q_{m_Q,u^*+cv^*}-Q^{\rm true}_{x,U+cS}$.
Appendix~\ref{app:main-step-td} proves stability and
\begin{equation}
 \mathcal A_Q:=\sup_p\left\{
 \|e_0^*(p)\|_\infty^2+
 \left\|\frac{e_c^*(p)-e_0^*(p)}{c}\right\|_\infty^2
 \right\}\le Cm_Q^{-2},
 \label{eq:main-paired-critic-floor}
\end{equation}
uniformly over the prescribed domain, with $C$ independent of
admissible $c,m_Q$.
The key point is that the critic-error difference is
$O(c/m_Q)$: division by $c$ preserves the approximation scale.
With $c^2\asymp\eps$ and $m_Q\asymp\eps^{-1/2}$, the paired
floor is $O(\eps)$, whereas separate endpoint bounds yield
only $O(1)$.
This is a population approximation bound; sampling and
moving-target errors from persistent training are controlled
jointly with actor tracking, starting from any deterministic
$z_0\in\mathcal K_Q$
(Appendix~\ref{app:main-step-tracking}).(Appendix~\ref{app:main-step-tracking}).

\subsection{Main Result}
Now with the assumptions in place the main theorem is stated as follows,

\begin{theorem}[Single-loop stationarity with a trained paired critic]
\label{thm:main}\label{thm:main-headline}
Fix $\gamma,I,K$, actor depth and retained dimension, and critic feature dimension. Assume A1, A2, G, and C1--C2 for the specified actor and critic~\eqref{eq:main-trained-critic}, at the selected fixed $\tau$ and neighborhoods. Initialize $U_0=S_0=0$ and any deterministic $z_0\in\mathcal K_Q$.
There exist fixed projection radii, actor buffers, positive step ratios $r_Q,q$, and positive constants independent of $\eps$ such that, for sufficiently small $\eps>0$, use
\begin{equation}
 \begin{gathered}
 \beta=k_\beta\eps,\qquad \eta_y=r_Q\beta,\qquad
 \eta_x=q\eta_y,\qquad c^2=k_c\eps,\qquad
 T=\lceil k_T\eps^{-2}\rceil,\\
 H=\Theta\!\left((1-\gamma)^{-1}\log(1/\eps)\right),\qquad
 m_Q\ge\max\{m_{Q,0},C_m\eps^{-1/2}\}.
 \end{gathered}
 \label{eq:main-schedule}
\end{equation}
Choose constants as in Appendix~\ref{app:main-step-containment}, with squared truncation biases summing to $o(\eps)$, positive TD/evaluation batches at most polynomial in $H$, and fixed preference batches as in Appendix~\ref{app:support-sampling}.
Run Algorithm~\ref{alg:main-trained} with $\delta_{\rm met}=\eps/2$, its certified metric, and finite KL solves. Total certificate-failure probability is at most $\eps$.
The output $x_{\rm out}$, uniform among training iterates on success and equal to $x_0$ otherwise, satisfies
\begin{equation}
 \E\|\nabla\Phi_{\tau,\mathbf m}(x_{\rm out})\|^2=O(\eps),\qquad
 N_{\rm seg},N_{\rm trans}=\widetilde{\mathcal O}(\eps^{-2}),
 \label{eq:main-headline-bound}
\end{equation}
where $N_{\rm seg},N_{\rm trans}$ count all sampled trajectory segments and environment transitions, including critic, evaluation, preference, and initialization data.  The tilde suppresses logarithmic factors in $1/\eps$.
\end{theorem}

\paragraph{Interpretation of the guarantee.}
The theorem establishes stationarity of the upper objective induced
by the selected policy response at fixed reference-KL regularization.
The sample complexity includes all training and initialization data,
including persistent critic learning from potentially inaccurate
initial value estimates.
Finite KL solves reuse stored metric states, with their numerical
work accounted for separately
(Appendices~\ref{app:support-prox} and~\ref{app:main-step-stationarity}).
Appendix~\ref{subsec:target-comparison} compares our optimization
target and sample complexity with prior guarantees.
Under an additional domain PL condition, the selected response is a
global lower minimizer as shown in
(\cref{cor:gpl-global-response}).

\subsection{Proof outline}
\label{subsec:main-proof-outline}

Write $\Phi=\Phi_{\tau,\mathbf m}$. The derived bounds
$\|\nabla\Phi\|\le G_\Phi$ and $\Phi\ge\underline\Phi$ hold globally
in $x$, with $\underline\Phi=0$ for Bradley--Terry and $-B_L$
otherwise (\cref{critic:actor-field-bounds}). Conditional
expectations $\E_t$ use the pre-draw history, including the
accepted metric sample. Appendix~\ref{app:main-results} follows
the six steps below.

\paragraph{1. Derived actor geometry and first-order response correction.}
(Appendix~\ref{app:main-step-actor}.)
A2, the reset component of occupancy, and the architectural
derivative bounds yield local reference-KL curvature.
Sufficiently strong fixed $\tau$ dominates the reward and
perturbation curvature, giving unique smooth interior base and
perturbed responses uniformly in $x$
(\cref{glob:geometry}). Their normalized difference yields
$V_c(x,\alpha_0^*(x),S_c^*(x))=\nabla\Phi(x)+O(c)$
(\cref{glob:secant}).
This identifies the first-order population direction to be
tracked, with squared response-approximation bias $O(c^2)$
and without assuming uniform lower-level PL.

\paragraph{2. Stable paired targets and normalized approximation errors.}
(Appendix~\ref{app:main-step-td}.)
The reset identity and Jensen's inequality give, for
square-integrable $h$,
\begin{equation}
 \E_{\nu_U}[h(h-\gamma K_Uh)]
 \ge(1-\sqrt\gamma)\|h\|_{L^2(\nu_U)}^2.
 \label{eq:outline-coercivity}
\end{equation}
C1 transfers this coercivity to linear TD; the $O(c)$ actor
and $O(m_Q^{-1})$ nonlinear field and derivative perturbations
preserve stability. This gives an interior paired target
$Z(p)$ with two bounded derivatives
(\cref{critic:TD-stability,critic:pair-global}).
Under C2, comparing the TD roots and their actor derivatives
with the realizable value family gives $O(m_Q^{-1})$ control
of both the value error and its derivative along the actor
path. Integrating this derivative between the endpoints yields
$\|e_c^*(p)-e_0^*(p)\|_\infty=O(c/m_Q)$.
Together with $\|e_0^*(p)\|_\infty=O(m_Q^{-1})$, this establishes
$\mathcal A_Q=O(m_Q^{-2})$
in~\eqref{eq:main-paired-critic-floor}, uniformly in the response
perturbation (\cref{critic:paired-floor}).
Thus shrinking $c$ does not amplify the normalized approximation
error of the population critic targets. The next step accounts
for stochastic tracking of these targets.

\paragraph{3. Shared samples, finite solves, and moving-target tracking.}
(Appendix~\ref{app:main-step-tracking}.)
Shared, appropriately weighted trajectories control endpoint
fluctuations before division by $c$. On the certified
neighborhoods, the centered endpoint difference has per-sample
second moment $O(c^2)$, so the normalized correction has
variance $O(B^{-1})$ for a batch of $B$ shared trajectories,
uniformly in $c$. Uniform geometric-length moments permit
fixed positive batches (\cref{critic:oracle}).
Evaluation uses the old critic and
$\widehat C=\widehat a+c\widehat R$, while a conditionally
independent TD block tracks $Z(p)$
(\cref{critic:recursions}).
For moving targets, conditional independence and centering
cancel linear noise cross terms, leaving the $1/\beta$ penalty
only on the mean target drift (\cref{critic:moving}).
The empirical metric and finite KL solves give endpoint error
at most $c\eta_y^2$, hence normalized error $O(\eta_y^2)$.
These solves reuse stored metric states and require no
additional environment samples
(\cref{klfin:prop:solver,klfin:prop:transfer}).

\paragraph{4. Joint tracking contraction.}
(Appendix~\ref{app:main-step-contraction}.)
Define the critic tracking error
$\mathcal C_t=\|z_t-Z(p_t)\|_W^2$.
Let $\mathcal T_t$ be the nonnegative actor tracking potential
in~\eqref{critic:actor-potential}, controlling errors in
$U_t,S_t$. For a fixed positive weight $\lambda$, set
\begin{equation}
 \mathcal D_t=\mathcal T_t+\lambda r_Q\mathcal C_t.
 \label{eq:outline-D}
\end{equation}
Sufficiently small fixed step ratios absorb the mutual
actor--critic error terms and give, for fixed
$\rho_D,C_D>0$ (\cref{critic:fast}),
\begin{equation}
 \begin{aligned}
 \E_t\mathcal D_{t+1}
 &\le(1-\rho_D\eta_y)\mathcal D_t
   +C_D\eta_y(qG_\Phi^2+\omega+\beta),\\
 \omega
 &=c^2+\mathcal A_Q+b_{\rm ev}^2+b_{\rm TD}^2,
 \end{aligned}
 \label{eq:outline-drift}
\end{equation}
where $b_{\rm ev},b_{\rm TD}$ bound truncation biases.
This inequality holds before exit and through the prospective
next step in the larger analytic neighborhood.
The term $qG_\Phi^2$ bounds target movement for containment;
Step~6 retains the gradient term explicitly to avoid introducing
this constant into the stationarity floor.

\paragraph{5. Proving containment of the coupled iterates.}
(Appendix~\ref{app:main-step-containment}.)
Choose actor geometry and fixed $\tau$, then critic radii and
potential weights, followed by $r_Q$ to accommodate the permitted
initial critic error and $q$ to control target drift.
Buffered prospective-step estimates justify the local recursions
before containment is known. A stopped martingale bound then
controls $\mathcal D_t$: under~\eqref{eq:main-schedule},
fluctuations of order
$\sqrt{\eta_y\log(T/\eps)}\operatorname{poly}(H)$
fit inside the actor buffers.
This excludes a first actor exit with conditional probability
at least $1-\eps/2$ for every accepted metric realization
(\cref{critic:D-max,critic:containment}).
Projection contains the critic, and G supplies global bounds
in the upper variable.

\paragraph{6. Upper stationarity and sample complexity.}
(Appendix~\ref{app:main-step-stationarity}.)
With fixed positive weights and constants, and
$\Delta\Psi_t=\Psi_{t+1}-\Psi_t$, upper descent and tracking give
(\cref{critic:master})
\begin{equation}
 \begin{aligned}
 \Psi_t
 &=\Phi(x_t)-\underline\Phi
   +a_\Psi q\mathcal T_t+b_\Psi qr_Q\mathcal C_t,\\
 \E_t\Delta\Psi_t
 &\le-a_*\eta_x
   \bigl(\|\nabla\Phi(x_t)\|^2+\mathcal T_t+\mathcal C_t\bigr) +C_*\eta_x(\omega+\beta+\eta_y+\eta_x).
 \end{aligned}
 \label{eq:outline-Psi}
\end{equation}
Telescoping with predictable pre-exit indicators and terminal
nonnegativity avoids conditioning on future containment.
Uniform output on success and fallback on failure yield
\begin{equation}
 \E\|\nabla\Phi(x_{\rm out})\|^2
 \le C\left[
 \frac{\Psi_0}{\eta_xT}+\omega+\beta+\eta_y+\eta_x
 \right]+\eps G_\Phi^2.
 \label{eq:outline-final-rate}
\end{equation}
The initial potential includes critic error.
Under~\eqref{eq:main-schedule}, $c^2=\Theta(\eps)$ makes the
squared response-secant bias $O(\eps)$, while
$m_Q^{-2}=O(\eps)$ gives the same order for the paired critic
approximation floor. Truncation ensures
$b_{\rm ev}^2+b_{\rm TD}^2=o(\eps)$, hence $\omega=O(\eps)$.
The choices $\eta_x,\eta_y,\beta=\Theta(\eps)$ and
$T=\Theta(\eps^{-2})$ also give
$\Psi_0/(\eta_xT)=O(\eps)$, so every term in the bound is
$O(\eps)$. The uniform sampling bounds from Step~3 allow these choices
without a polynomial increase in batch size as $c$ shrinks.
Counting TD, response-evaluation, preference, and initialization
samples therefore gives
$\widetilde{\mathcal O}(\eps^{-2})$ sample complexity.
Appendix~\ref{app:gaur-rate-comparison} explains the
$\widetilde{\mathcal O}(\eps^{-3})$ sample complexity established
by \citet{gaur2025sample} and the source of our improved bound.
\section{Empirical Results}
\label{sec:empirical-results}
We focus on interaction scaling in the exact continuous-state construction
and end-to-end performance on Walker Walk.

\begin{figure}[h]
\centering
\includegraphics[width=\linewidth]{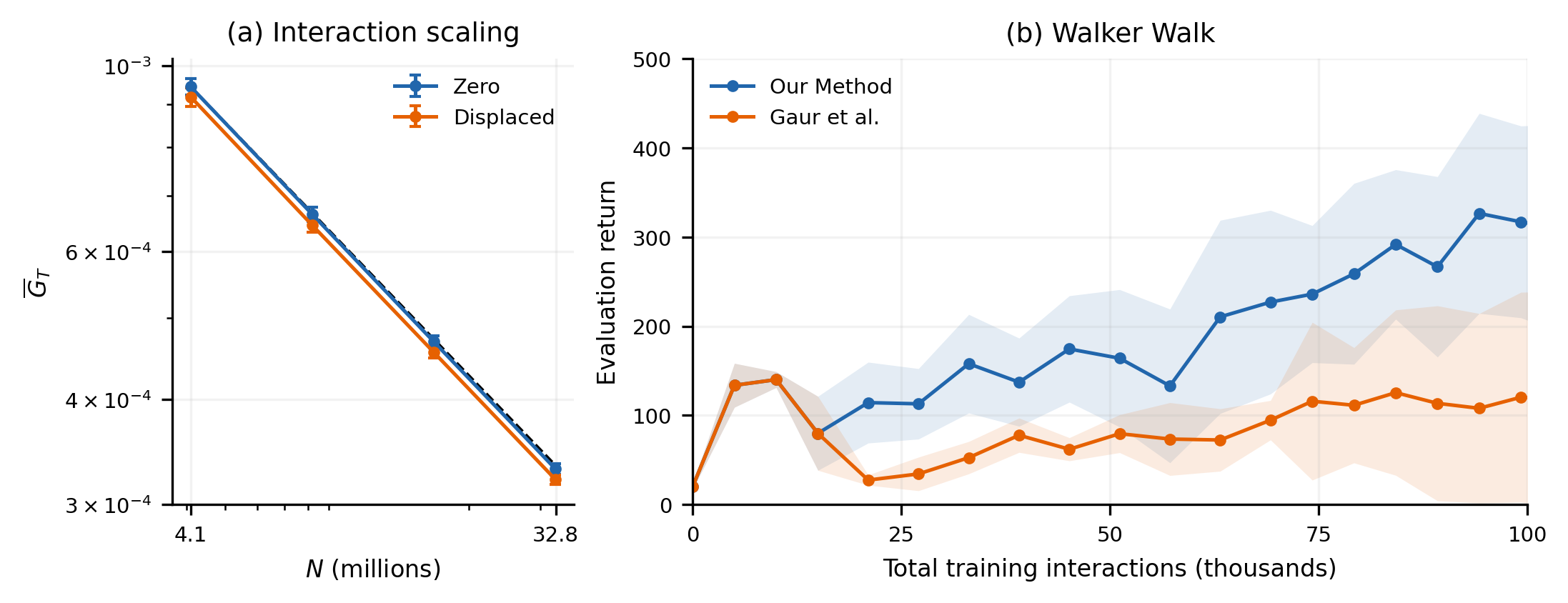}
\caption{\textbf{Interaction scaling and continuous control.}
(a) Average squared upper-gradient norm,
$\overline G_T=T^{-1}\sum_{t<T}|\Phi'(x_t)|^2$, versus total training
cost $N$, for zero and displaced critic initialization. Error bars show
95\% $t$ intervals over ten seeds; the dashed line is an $N^{-1/2}$ reference.
(b) Walker Walk evaluation returns: means and one-standard-deviation bands
across five matched training seeds. 
}
\label{fig:continuous-headline}
\end{figure}

We use the exact continuous-state, two-action construction in
Appendix~\ref{crex:example}, with a tied multilayer actor and a
fixed-feature nonlinear critic. Training uses sampled trajectories;
exact reference calculations provide the upper-gradient evaluation metric.
Figure~\ref{fig:continuous-headline}(a) shows similar decay of the average
squared upper-gradient norm for an initially exact zero critic and a
displaced critic with inaccurate value estimates, consistent with the
predicted $N^{-1/2}$ trend and
$\widetilde{\mathcal O}(\varepsilon^{-2})$ interaction complexity.
The prior works in Table~\ref{tab:main-related-comparison} do not report
this direct check of the predicted stationarity scaling against total
sampled interactions.
On Walker Walk, we learn rewards from trajectory preferences and
incorporate the response correction into reward updates while training a
continuous-control policy. Figure~\ref{fig:continuous-headline}(b) shows
higher mean returns than the adapted method of \citet{gaur2025sample}.
Appendices~\ref{app:continuous-scaling-theory} and
\ref{app:continuous-initialization-comparison} explain the scaling and
initializations; Appendices~\ref{app:continuous-validation} and
\ref{app:end-to-end-details} give protocols and implementation qualifications for the exact and end-to-end cases resepctively.

\section{Conclusion}
\label{sec:main-interpretation}
We developed a single-loop, fully first-order bilevel RL method with
explicit neural actors and a persistently trained nonlinear critic.
We proved $\widetilde{\mathcal O}(\varepsilon^{-2})$ sample complexity for
$O(\varepsilon)$ expected squared upper-gradient norm under the stated
assumptions.

\FloatBarrier
\subsection*{AI use statement}
Generative AI tools were used for writing assistance. The authors verified all mathematical statements, proofs, citations, and experimental results, and take full responsibility for the content of the submission.

\subsection*{Reproducibility Statement}
The supplementary material includes the experiment code, dependencies, configurations, and seeds, with setup instructions in the README in order to rerun the experiments from scratch, generate all given figures.

\bibliography{references}
\clearpage

\appendix
%
%
%

\section*{Contents of the Appendix}

\begingroup
\small
\makeatletter
\c@tocdepth=2\relax
\parskip=0pt\relax

\ExplSyntaxOn
\cs_set_eq:NN \__brl_appendix_original_contentsline:nnnn \contentsline
\cs_set_protected:Npn \contentsline #1#2#3#4
  {
    \regex_match:nnT
      { \A (?: appendix | section | subsection ) \. [A-Z]+ (?: \. | \Z ) }
      { #4 }
      { \__brl_appendix_original_contentsline:nnnn {#1}{#2}{#3}{#4} }
  }
\ExplSyntaxOff

\@starttoc{toc}
\makeatother
\endgroup

\section{Related Work}
\label{sec:main-related}

\paragraph{Standard reinforcement learning and actor--critic methods.}
Classical reinforcement learning (RL) methods optimize a single
policy objective, with policy-gradient methods providing a framework
for direct policy optimization under function approximation
\citep{williams1992reinforce,sutton1999policygrad}.
Actor--critic algorithms combine policy improvement with value-function
estimation, with early convergence analyses establishing their
properties under two-timescale stochastic approximation
\citep{konda2000actor,bhatnagar2009natural,pmlr-v202-gaur23a}.
Subsequent work has developed finite-sample guarantees for temporal
difference learning and actor--critic methods under function
approximation and Markovian sampling
\citep{bhandari2018td,xu2020actorcritic,agarwal2021policy,pmlr-v235-gaur24a}.
Neural policy-gradient analyses have further established convergence
and optimality guarantees under suitable overparameterization and
representation assumptions \citep{wang2020neuralpg}.
These developments motivate our use of first-order policy updates
and persistent critic learning. However, standard RL analyses
typically optimize a fixed policy objective rather than an upper
objective induced by an implicitly defined policy response.
In bilevel RL, the lower-level policy changes with the upper-level
parameter, introducing additional response-sensitivity and estimation
errors. In particular, ordinary critic accuracy guarantees do not
directly control the normalized differences of critic errors required
by our response correction. Our analysis addresses this additional
coupling through local response geometry, shared sampling, and
paired critic tracking for the specified actor--critic architecture.

\paragraph{Bilevel optimization and RL.}
Implicit bilevel methods control Hessian inverses or iterative hypergradients
\citep{franceschi2018bilevel,grazzi2020iteration,ji2021bilevel}; penalty methods
permit first-order stochastic updates \citep{kwon2024penalty}.
In BRL, PARL uses implicit response calculations \citep{chakraborty2024parl},
penalty approaches prescribe lower-solution accuracy \citep{shen2025principled},
and soft Bellman structure yields tabular hypergradients \citep{yang2025bilevel}.
\citet{gaur2025sample} analyze continuous state--action spaces under uniform
lower PL, with value-estimation error controlled by an approximation assumption
and an imported actor--critic bound. \citet{zeng2026hessianfree} give a
single-loop tabular actor--critic guarantee under regularization-dependent PL
and Hessian conditions. Table~\ref{tab:main-related-comparison} compares assumptions;
Section~\ref{subsec:target-comparison} distinguishes targets.

\paragraph{Critic construction and coupled learning.}
Prior BRL implementations already use neural critics
\citep{shen2025principled,yang2025bilevel}. We analyze an explicit nonlinear
critic whose persistent paired updates track moving values and control the
normalized difference of endpoint errors as the response perturbation
shrinks. Separate endpoint accuracy need not provide this control after
division by the perturbation. The proof requires fixed bounded features,
excitation, and realizability of the regularized value family
(Appendices~\ref{app:main-step-td} and~\ref{crex:example}); it does not
establish unrestricted feature learning or nonlinear expressive superiority.

\paragraph{Policy geometry and TD.}
Natural policy gradients and trust regions motivate Fisher geometry
\citep{kakade2001natural,schulman2015trpo,agarwal2021policy}.
Our initialization certificate yields local response curvature, in contrast
to domain-wide Fisher nondegeneracy \citep{liu2020,fatkhullin2023}.
Feature-conditioned TD stability \citep{bhandari2018td,ke2024td} supports
our nonlinear paired tracking proof. The context after A1--C2 details the restrictions.

\paragraph{Bilevel optimization.}
Bilevel optimization couples an upper objective to the solution of a
parameterized lower problem. With smooth, strongly convex lower
objectives, the unique lower response supports convergence analyses
based on implicit differentiation or differentiation through lower
updates \citep{grazzi2020iteration,ji2021bilevel}.
For convex lower objectives with multiple minimizers, bilevel descent
aggregation incorporates the upper objective when selecting among
lower solutions \citep{liu2020aggregation}.
Nonconvex lower problems introduce further difficulties: stationary
points can be suboptimal, and the response reached by optimization
can depend on initialization.
Approaches to this setting include value-function reformulations
\citep{liu2021} and selection maps that explicitly specify which
lower critical point defines the upper objective
\citep{arbel2022selection}.
Our work develops this local-response perspective for BRL, deriving
response curvature for explicit neural actors and controlling the
coupled learning of policies and persistent nonlinear critics.
\paragraph{Comparison with lower bounds.}
\citet{ji2026lower} establish an
$\Omega(\delta^{-4})$ stochastic oracle lower bound
for zero-respecting first-order methods on
nonconvex-strongly-convex bilevel problems, where $\delta$ denotes
expected gradient-mapping norm accuracy.
With fixed problem constants and $\delta=\sqrt{\varepsilon}$,
this yields an $\Omega(\varepsilon^{-2})$ benchmark for squared
stationarity accuracy, matching the accuracy exponent of our
$\widetilde{\mathcal O}(\varepsilon^{-2})$ interaction bound.
This comparison does not establish minimax optimality for our
structured RL setting, since the problem classes and oracle
models differ; matching condition-number dependence also
remains open.

\subsection{Optimization target and comparison with prior guarantees}
\label{subsec:target-comparison}
The target $\Phi_{\tau,\mathbf m}$ uses fixed reference-KL regularization:
$c$ is an estimation perturbation with vanishing bias, whereas $\tau$
remains part of the lower problem. \citet{yang2025bilevel} use a global
entropy-regularized tabular response; \citet{gaur2025sample} global lower
responses in their parameter class under uniform PL;
\citet{shen2025principled} a penalty reformulation linked to lower optimality;
and \citet{zeng2026hessianfree} fixed and vanishing regularization under
their geometric conditions. Our explicit structure yields local curvature
and joint containment. We can show that under similar assumptions to \cite{gaur2025sample} we can attain the same stationarity guarantee for a selected globally optimal lower response.

\begin{assumption}[Optional PL on a prescribed actor domain]
\label{ass:gpl-domain}
Fix the finite actor architecture $\mathbf m$ and $\tau$ of
\cref{thm:main-headline}, and a compact convex $\mathcal D\subset\R^r$
with $\overline B_{R_{\rm geom}}\subset\operatorname{int}\mathcal D$,
independent of $x,c,m_Q,\eps$. With
$g^*_{\tau,\mathbf m,\mathcal D}(x)=\min_{\alpha\in\mathcal D}
 g_{\tau,\mathbf m}(x,\alpha)$, assume a constant $\mu_{\rm PL}>0$
independent of $x,\eps$ such that, for all $x\in\R^{d_x}$ and
$\alpha\in\mathcal D$,
\begin{equation}
 \tfrac12\|\nabla_\alpha g_{\tau,\mathbf m}(x,\alpha)\|^2
 \ge\mu_{\rm PL}\bigl[g_{\tau,\mathbf m}(x,\alpha)
                    -g^*_{\tau,\mathbf m,\mathcal D}(x)\bigr].
 \label{eq:gpl-domain-pl}
\end{equation}
The gradient is the ambient Euclidean gradient, including at the boundary.
\end{assumption}

\begin{corollary}[Selected global regularized response]
\label{cor:gpl-global-response}
Under \cref{thm:main-headline}'s conditions and \cref{ass:gpl-domain},
$\alpha^*_{0,\mathbf m}(x)\in\argmin_{\alpha\in\mathcal D}
 g_{\tau,\mathbf m}(x,\alpha)$ for every $x$.
Thus $\Phi^{\rm sel}_{\tau,\mathbf m,\mathcal D}(x)
 :=f_{\mathbf m}(x,\alpha^*_{0,\mathbf m}(x))\equiv\Phi_{\tau,\mathbf m}(x)$
is a selected global-response objective, and the same output satisfies
\eqref{eq:main-headline-bound} with unchanged label and numerical costs.
\end{corollary}
Appendix~\ref{app:gpl-global-response} proves the corollary, gives selection-independence
conditions, and rules out full-space PL for this bounded architecture.
Initialization, fixed $\tau$, and upper stationarity are unchanged.

\subsubsection{Source of the sample-complexity improvement}
\label{app:gaur-rate-comparison}

We compare the sample-complexity exponents for squared-gradient
stationarity, retaining the distinctions between optimization targets
and assumptions discussed above.

\paragraph{The published analysis.}
In the notation of \citet[Theorem~1]{gaur2025sample}, let $\sigma$
denote the perturbation, $T_{\rm G},K_{\rm G}$ the outer and inner
iteration counts, $n_{\rm G},B_{\rm G}$ the sampling budgets, and
$H_{\rm G}$ the rollout horizon.
Writing $\mathcal S_{\rm G}$ for their averaged squared-stationarity
criterion and $\varepsilon_{\rm approx,G}$ for their approximation
term, their bound has the form
\begin{equation}
 \mathcal S_{\rm G}
 \le \widetilde{\mathcal O}\!\left(
 \frac{1}{T_{\rm G}}
 +\frac{e^{-K_{\rm G}}}{\sigma^2}
 +\frac{1}{\sigma^2 n_{\rm G}}
 +\frac{\gamma^{2H_{\rm G}}}{\sigma^2 B_{\rm G}}
 +\sigma^2+\varepsilon_{\rm approx,G}
 \right).
 \label{eq:gaur-comparison-bound}
\end{equation}
Their stated schedule uses, up to logarithmic factors,
\begin{equation}
 \begin{gathered}
 \sigma^2\asymp\varepsilon,\qquad
 T_{\rm G}\asymp\varepsilon^{-1},\qquad
 n_{\rm G},B_{\rm G}\asymp\varepsilon^{-2},\\
 K_{\rm G},H_{\rm G}=O(\log(1/\varepsilon)).
 \end{gathered}
 \label{eq:gaur-comparison-schedule}
\end{equation}
This gives
$\mathcal S_{\rm G}
 =\widetilde{\mathcal O}(\varepsilon+\varepsilon_{\rm approx,G})$
and the sample count
\begin{equation}
 N_{\rm G}
 =n_{\rm G}K_{\rm G}T_{\rm G}
  +B_{\rm G}K_{\rm G}H_{\rm G}T_{\rm G}
  +B_{\rm G}H_{\rm G}T_{\rm G}
 =\widetilde{\mathcal O}(\varepsilon^{-3}).
 \label{eq:gaur-comparison-cost}
\end{equation}
The term $1/(\sigma^2n_{\rm G})$ exposes the estimation cost:
with $\sigma^2\asymp\varepsilon$, taking
$n_{\rm G}\asymp\varepsilon^{-2}$ makes this contribution
$O(\varepsilon)$.

\paragraph{Joint control of normalized response errors.}
Our analysis controls endpoint differences jointly before dividing
by the response perturbation. Let $\bar\zeta_0,\bar\zeta_c$ denote
the centered endpoint errors averaged over $B$ shared trajectories.
On the prescribed actor and critic neighborhoods,
\begin{equation}
 \E_t\|\bar\zeta_c-\bar\zeta_0\|^2
 \le \frac{Cc^2}{B},
 \qquad
 \E_t\left\|
 \frac{\bar\zeta_c-\bar\zeta_0}{c}
 \right\|^2
 \le \frac{C}{B},
 \label{eq:gaur-comparison-coupling}
\end{equation}
with $C$ independent of $c$ and $B$
(\cref{critic:oracle}).
The shared-trajectory construction and its weighting therefore
control the sampling variance uniformly as the perturbation shrinks.

The critic approximation requires a corresponding joint estimate.
The paired population TD targets satisfy
\begin{equation}
 \|e_0^*(p)\|_\infty^2+
 \left\|\frac{e_c^*(p)-e_0^*(p)}{c}\right\|_\infty^2
 \le C m_Q^{-2},
 \label{eq:gaur-comparison-critic}
\end{equation}
uniformly over the prescribed neighborhood
(\cref{critic:paired-floor}).
Thus neither centered sampling noise nor the normalized critic
approximation floor introduces an inverse power of $c$.
The moving-target analysis additionally accounts for the error
incurred while learning these critics persistently
(Appendix~\ref{app:main-step-tracking}).

\paragraph{Propagation to the final sample complexity.}
The joint descent and tracking bound
\eqref{eq:outline-final-rate} combines these estimates with the
initial-error contribution $\Psi_0/(\eta_xT)$.
Under~\eqref{eq:main-schedule},
\begin{equation}
 \begin{gathered}
 c^2=\Theta(\varepsilon),\qquad
 m_Q^{-2}=O(\varepsilon),\\
 \eta_x,\eta_y,\beta=\Theta(\varepsilon),\qquad
 T=\Theta(\varepsilon^{-2}).
 \end{gathered}
 \label{eq:gaur-comparison-our-schedule}
\end{equation}
The squared secant bias, paired critic approximation floor,
step-size-weighted stochastic terms, and initial-error contribution
are then all $O(\varepsilon)$. The prescribed truncation and
metric-certificate accuracies control the remaining terms.

Uniform sampling moments allow fixed or logarithmically growing
batches. Consequently, counting all TD, response-evaluation,
preference, and initialization samples gives
\begin{equation}
 N_{\rm ours}
 =\widetilde{\mathcal O}(T)
 =\widetilde{\mathcal O}(\varepsilon^{-2}),
 \label{eq:gaur-comparison-our-cost}
\end{equation}
including critic training.
The improved exponent follows from joint control of errors in the
divided response correction together with persistent actor--critic
tracking. This comparison concerns the guarantees established by
the two analyses under their respective assumptions.

\section*{Organization of the Appendix}

The appendix separates the proof of the main theorem, its supporting
constructions and estimates, and the experimental evidence.
Appendix~\ref{app:main-results} gives the main proof sequence;
Appendix~\ref{app:supporting-results} supplies its technical ingredients.
Appendices~\ref{app:continuous-validation}--\ref{app:empirical-requirements}
describe the experiments.

\paragraph{Appendix~\ref{app:main-results}: Main results.}
This section proves Theorem~\ref{thm:main-headline} in the order of the
six-step proof outline in the main text.
Section~\ref{app:main-step-actor} establishes the actor geometry and
response correction.
Section~\ref{app:main-step-td} derives stable paired-TD targets and their
approximation accuracy, while Section~\ref{app:main-step-tracking}
controls shared-sample estimation and persistent tracking.
Sections~\ref{app:main-step-contraction} and
\ref{app:main-step-containment} construct a contracting potential and
exclude a first exit from the admissible region.
Section~\ref{app:main-step-stationarity} then proves upper-objective
stationarity and accounts for the data used in training and metric
estimation.

\paragraph{Appendix~\ref{app:supporting-results}: Supporting results.}
Sections~\ref{app:support-actor}--\ref{app:support-response} develop the
multilayer actor construction, initialization certificates, preference-loss
derivatives, and response calculus.
Section~\ref{app:support-prox} establishes the common-KL geometry and
implements the actor updates through an empirical metric and a finite
local solver, including the effect of numerical errors.
Sections~\ref{app:support-critic}--\ref{app:support-tools} provide the
critic, sampling, conditional-tracking, and concentration estimates used
in the main proof.
Sections~\ref{app:support-verification} and \ref{app:pl-strengthened}
discuss assumption verification and counterexamples to uniform
lower-level PL guarantees.
Section~\ref{crex:example} constructs a continuous-state instance
with an explicit response, verified critic-feature conditions, and a
nonzero chain correction; it also clarifies the interpretation of the
on-policy preference objective.
Section~\ref{app:gpl-global-response} proves the optional global-response
corollary, states selection-independence conditions, and explains the
obstruction to uniform PL on the full actor space.

\paragraph{Appendix~\ref{app:continuous-validation}: Controlled continuous-state experiments.}
This section specifies the actor and nonlinear critic, sampled training
protocol, initialization comparisons, and reference calculations.
It reports response-secant bias, shared versus independent sampling
variance, critic approximation and tracking, and stationarity versus
charged interactions. It also records the controls, uncertainty
estimates, numerical safeguards, pilot history, and qualifications of
the finite-budget scaling results.

\paragraph{Appendix~\ref{app:end-to-end-details}: End-to-end benchmarks.}
This section documents Walker Walk and Highway-fast, including the SAC
implementation, preference collection, baselines, architectures,
hyperparameters, learning curves, and checkpoint and per-seed returns.
It also gives the uncertainty estimates and accounting of main and
auxiliary training interactions.
\section{Main Results}
\label{app:main-results}
This section proves Theorem~\ref{thm:main-headline} under A1, A2, G and
C1--C2, following the six steps of the main-text proof outline. The target
is the selected fixed-reference-KL response objective; the dynamics are
fixed, actions are finite, and preferences use fixed finite tuples.
The proofs of the actor construction and auxiliary calculations are in
Appendix~\ref{app:supporting-results}.

\paragraph{Notation and sampling history.}
Suppress actor-width subscripts and write $g=g_\tau$, $\Phi=\Phi_{\tau,\mathbf m}$,
$p=(x,U,S)$, $\alpha_0=U$, and $\alpha_c=U+cS$.
All norms are Euclidean or induced operator norms unless a subscript is given.
Let $\mathcal F_t$ be the history before the current data blocks, including
the old actors, upper parameter, critic, and independent initial metric
sample. On acceptance, the common-KL law is fixed at $\widehat\rho_n$
throughout training, with the certificate of \cref{klfin:prop:Fisher}. Write $\E_t=\E[\,\cdot\mid\mathcal F_t]$.
The lower, outer and TD blocks are independent conditional on this history;
the two preference blocks are separate and use the old perturbed actor.
If needed, $\mathcal G_t$ additionally reveals the lower block and actor
updates; outer samples remain fresh under that conditioning. The new
critic is first used at the next iteration.

Constants denoted $C$ may change between inequalities. They may depend on
the fixed discount, tuple sizes, actor depth and retained dimensions,
feature conditioning, and the sufficiently large but fixed $\tau$.
They are independent of $x$, accuracy, $0<c\le c_0$, $m_Q\ge m_{Q,0}$,
and admissible actor widths. Uniformity under later reductions of step
ratios is stated where needed. $L$ denotes actor depth, $K$ preference
trajectory length, $H$ the truncation cap and $N$ a geometric prefix length.
\paragraph{Policy transition and reset occupancy.}
For bounded measurable $h$, the policy transition operator and normalized
discounted state--action occupancy used in C1--C2 are
\begin{equation}
 \begin{gathered}
 (K_\alpha h)(s,a)=\E_{s'\sim P(\cdot\mid s,a)}
       \sum_{a'\in\cA}\pi_\alpha(a'\mid s')h(s',a'),\\
 \rho_U(ds,da)=\rho(ds)\pi_U(da\mid s),\qquad
 \nu_U=(1-\gamma)\sum_{k\ge0}\gamma^k\rho_UK_U^k.
 \end{gathered}
 \label{eq:critic-occupancy}
\end{equation}
Thus $\nu_U=(1-\gamma)\rho_U+\gamma\nu_UK_U$; its state marginal is
$d_U^\gamma$.

\subsection{Actor geometry and response correction}
\label{app:main-step-actor}
Write $j(x,\alpha)=(1-\gamma)\E_{\pi_\alpha}\sum_{t\ge0}\gamma^t r_x(s_t,a_t)$
and $\mathcal K(\alpha)=(1-\gamma)\E_{\pi_\alpha}\sum_{t\ge0}\gamma^t
\log(\pi_\alpha/\piRef)(a_t\mid s_t)$, so $g=-j+\tau\mathcal K$.
Put $h_c=g+cf$. The construction in Appendix~\ref{app:support-actor}
supplies actor derivative bounds $C_k=C_{k,L}$ and the KL Hessian
Lipschitz constant $L_K$, independent of $x$, width, and $\tau$.
Let $a_0=(1-\gamma)\lambda_{\rm eff,L}$ and $\kappa=a_0/4$, and choose
\begin{equation}
 0<R=R_A=R_{\rm geom}\le
 \min\{A/2,a_0/(2L_K),\sqrt{\lambda_{\rm eff,L}}/(2C_2)\}.
 \label{eq:R}
\end{equation}
Omit a restriction with zero denominator. Write $B_R$ for the open ball
and $\Theta_R=\overline B_R$ for its closure. Define
$J_q=B_r A_q(\gamma)$, $G_f=B_Lb_1(IK)$ and $L_f=B_Lb_2(IK)$
using the likelihood polynomials in Appendix~\ref{app:support-preference}.
Fix a separation radius $s=S_{\rm tube}>0$ independently of $\tau$;
reduce the preliminary cap $c_0$ so $c_0s\le R/8$.
The common-KL radius $r_{\rm prox}\le R/8$ is chosen in
Appendix~\ref{app:support-prox}. All actor bounds below are used where
$\|S\|\le s$ and $U+c\overline B_s\subset B_R$, with the additional
local-update buffer specified in Step~\ref{app:main-step-containment}.
\begin{proposition}[One certified actor neighborhood for every $x$]
\label{glob:geometry}
For a fixed perturbation cap $c_0$, choose a finite $\tau\ge1$ satisfying
\begin{equation}
 \tau\ge\frac{4(J_2+c_0L_f)}{a_0},\qquad
 \frac{J_1+c_0G_f}{\kappa\tau}<\frac R4.
 \label{glob:tau-choice}
\end{equation}
Then $h_c=g_\tau+cf$ has actor Hessian at least
$\mu_\tau I$, $\mu_\tau=\kappa\tau$, on $B_R$ for every $x$ and
$0\le c\le c_0$. Its unique root on this ball obeys
\begin{equation}
 \norm{\alpha_c^\star(x)}\le\frac{J_1+cG_f}{\mu_\tau}<R/4,
 \qquad
 \norm{\alpha_c^\star(x)-\alpha_0^\star(x)}
       \le\frac{cG_f}{\mu_\tau}.
 \label{glob:root-bounds}
\end{equation}
The locally defined roots agree on overlaps and give a globally defined
selected response on $\R^{d_x}$, smooth to the orders used below.
\end{proposition}
\begin{proof}
Lemma~\ref{lem:reference} and the KL Hessian Lipschitz bound give
$\mathcal K_{\alpha\alpha}\succeq(a_0/2)I$ on $B_R$.
Lemma~\ref{glob:mixed} yields
$h_{c,\alpha\alpha}\succeq(\tau a_0/2-J_2-c_0L_f)I
\succeq\mu_\tau I$. Since the KL gradient vanishes at initialization,
$\norm{h_{c,\alpha}(x,0)}\le J_1+cG_f$. On any sphere of radius
strictly between $(J_1+cG_f)/\mu_\tau$ and $R$, strong monotonicity
makes the gradient point outward. A minimizer on the closed smaller
ball is therefore interior. Strong convexity gives uniqueness and the
first distance bound. Subtract the root equations for $c$ and $0$ and
use strong monotonicity to obtain the second bound. The implicit-function
theorem applies because the actor Hessian is invertible; uniqueness
identifies the local branches wherever they overlap. This asserts a
selected local neural response, not an unrestricted global optimum.
\end{proof}

Let $G_{\alpha x}$, $G_{\alpha xx}$, $G_{\alpha\alpha x}$, and
$G_{\alpha\alpha\alpha}$ denote the corresponding uniform derivative
bounds for $g_\tau$, and write $F_x,F_\alpha,F_{xx},F_{x\alpha},
F_{\alpha\alpha}$ for those of $f$. Define
\begin{align}
 a_1&=\mu_\tau^{-1}G_{\alpha x},\notag\\
 a_2&=\mu_\tau^{-1}\bigl(G_{\alpha xx}
          +2G_{\alpha\alpha x}a_1
          +G_{\alpha\alpha\alpha}a_1^2\bigr).
 \label{glob:response-constants}
\end{align}

\begin{corollary}[Global gradient, Hessian, and value bounds]
\label{glob:Phi}
The selected response satisfies $\norm{D_x\alpha_0^\star}\le a_1$ and
$\norm{D_x^2\alpha_0^\star}\le a_2$. Consequently,
\begin{align}
 \norm{\nabla\Phi_\tau(x)}
 &\le F_x+F_\alpha a_1=:G_\Phi,\label{glob:gradient-bound}\\
 \norm{\nabla^2\Phi_\tau(x)}_\op
 &\le F_{xx}+2F_{x\alpha}a_1+F_{\alpha\alpha}a_1^2
               +F_\alpha a_2=:L_\Phi.\label{glob:Hessian-bound}
\end{align}
Also $-B_L\le\Phi_\tau(x)\le B_L$, with lower bound $0$ for
\eqref{glob:BT-loss}. The gradient is globally Lipschitz, and the usual
quadratic upper Taylor bound holds along every upper-update segment.
\end{corollary}
\begin{proof}
Set $a(x)=\alpha_0^\star(x)$, $G_0=g_{\alpha\alpha}(x,a(x))$, and
$B=g_{\alpha x}(x,a(x))$. Differentiating $g_\alpha(x,a(x))=0$ gives
$D_xa=-G_0^{-1}B$. In unit directions $u,v$, differentiating once more
gives
\begin{align*}
 G_0 D_x^2a[u,v]=-\bigl(&g_{\alpha xx}[u,v]
   +g_{\alpha\alpha x}[D_xa[u],v]
   +g_{\alpha\alpha x}[D_xa[v],u]\\
   &+g_{\alpha\alpha\alpha}[D_xa[u],D_xa[v]]\bigr).
\end{align*}
The inverse bound $\norm{G_0^{-1}}\le\mu_\tau^{-1}$ proves
\eqref{glob:response-constants}. Applying the chain rule twice to
$f(x,a(x))$ proves the two bounds. Integrating the bounded Hessian
along a segment in $\R^{d_x}$ proves the final statement.
\end{proof}

\begin{lemma}[Uniform consistency of the response correction]
\label{glob:secant}
With the population field $V_c$ in \eqref{eq:main-V},
\begin{equation}
 \norm{V_c(x,\alpha_0^\star(x),S_c^\star(x))
           -\nabla\Phi_\tau(x)}\le C_{\rm sec}c,
 \qquad S_c^\star=(\alpha_c^\star-\alpha_0^\star)/c,
 \label{glob:secant-consistency}
\end{equation}
uniformly over $x$, width, and $0<c\le c_0$. On the actor tube, the
field is uniformly Lipschitz in $(U,S)$, with no inverse power of $c$.
\end{lemma}
\begin{proof}
For fixed $x$, implicit differentiation in $c$ gives
$\partial_c\alpha_c^\star=-h_{c,\alpha\alpha}^{-1}
f_\alpha(x,\alpha_c^\star)$, hence a uniform bound $G_f/\mu_\tau$.
A second derivative satisfies
\[
 \partial_c^2\alpha_c^\star=-h_{c,\alpha\alpha}^{-1}
 \left[h_{c,\alpha\alpha\alpha}
    [\partial_c\alpha_c^\star,\partial_c\alpha_c^\star]
        +2f_{\alpha\alpha}\partial_c\alpha_c^\star\right],
\]
which is uniformly bounded by Lemma~\ref{glob:mixed}. Taylor expansion
in $c$ of $g_x(x,\alpha_c^\star)$ and the Lipschitz bound for
$f_x(x,\cdot)$ show that $V_c$ differs by $O(c)$ from
$f_x-g_{x\alpha}g_{\alpha\alpha}^{-1}f_\alpha$, evaluated at the
base response. This is the exact response-chain gradient. Finally,
\[
 V_c(x,U,S)=\int_0^1g_{x\alpha}(x,U+\theta cS)S\,d\theta
                    +f_x(x,U+cS).
\]
Differentiating this integral in $U,S$ uses only the bounded blocks
$g_{x\alpha},g_{x\alpha\alpha},f_{x\alpha}$ and bounded $S$.
\end{proof}

\paragraph{Actor tracking potential.}
For $c>0$, define the scaled potential
\begin{equation}
 \mathcal H_c(x,U,S)=
 \frac{g(x,U+cS)-g(x,U)-c\langle g_\alpha(x,U),S\rangle}{c^2}
 +\frac{f(x,U+cS)-f(x,U)}c.
 \label{eq:scaled-objective}
\end{equation}
Its $S$ gradient is $R_c$ from \eqref{eq:main-R}. Let
$\bar S_c(x,U)$ be its unique interior minimizer over $\overline B_s$;
Appendix~\ref{app:support-response} proves existence for sufficiently
large fixed $\tau$ and the integral extension at $c=0$.
Set $\Delta_0(x,U)=g(x,U)-g(x,\alpha_0^*(x))$ and
$\Gamma_c(x,U,S)=\mathcal H_c(x,U,S)-\mathcal H_c(x,U,\bar S_c(x,U))$.
For the fixed positive weight $\lambda_S$ selected in
Lemma~\ref{lem:tau-scaling}, the nonnegative actor potential is
\begin{equation}
 \mathcal T(x,U,S)=\Delta_0(x,U)+\lambda_S\Gamma_c(x,U,S).
 \label{critic:actor-potential}
\end{equation}
\begin{lemma}[Certified actor geometry and local update bounds]
\label{critic:actor-input}
For the radii just selected, there are constants $\kappa,E_*,J_1,G_f>0$. A sufficient finite fixed $\tau$ yields $\mu_\tau=\kappa\tau$ and, uniformly in $x$ and sufficiently small $c$,
\begin{align}
 \norm{\alpha_0^\star(x)}&\le J_1/\mu_\tau,
 &\norm{S_c^\star(x)}&\le G_f/\mu_\tau,\notag\\
 \norm{U-\alpha_0^\star(x)}^2+\norm{S-S_c^\star(x)}^2
 &\le E_*\mathcal T(x,U,S)/\mu_\tau,\notag\\
 \mathcal T(x,0,0)&\le\frac{J_1^2+\lambda_SG_f^2}{2\mu_\tau}.
 \label{critic:actor-cert}
\end{align}
Here $E_*,\lambda_S$ can be fixed independently of sufficiently large $\tau$. The needed gradients and full $(x,U,S)$ Hessian of $\mathcal T$ are uniformly bounded on the prospective actor tube. Moreover,
\begin{equation}
 \begin{gathered}
 \norm{\nabla\Phi}\le G_\Phi,\qquad
 \norm{\nabla^2\Phi}\le L_\Phi,\qquad
 \Phi\ge\underline\Phi,\\
 \norm{a}^2+\norm{R_c}^2\le C\mathcal T,\qquad
 \norm{V_c}^2\le C(\norm{\nabla\Phi}^2+\mathcal T+c^2).
 \end{gathered}
 \label{critic:actor-field-bounds}
\end{equation}
For preference cross-entropy take $\underline\Phi=0$; for a bounded general loss take $-B_L$.

Write $P^0=\Fisher_{\mathbf m,\widehat\rho_n}(U)^{-1}$, $P^c=\Fisher_{\mathbf m,\widehat\rho_n}(U+cS)^{-1}$ and $K_c=(P^0-P^c)/c$. On metric acceptance these operators obey the same fixed bounds by Proposition~\ref{klfin:prop:Fisher}. For compatible samples satisfying $\widehat C=\widehat a+c\widehat R$, the finite actor updates admit
\begin{align}
 \Delta U&=-\eta_y P^0\widehat a+\mathfrak r_U,\notag\\
 \Delta S&=-\eta_y P^c\widehat R+\eta_y K_c\widehat a+\mathfrak r_S,
 \qquad \Delta x=-\eta_x\widehat V.
 \label{critic:actor-expansion}
\end{align}
Here $\mathfrak r_U=\rho^U+e_U$ and $\mathfrak r_S=\rho^S+e_S$, where $\rho^U,\rho^S$ are the exact reference-map remainders and the finite-solver errors satisfy $\|e_U\|\le c\eta_y^2$, $\|e_S\|\le2\eta_y^2$ samplewise. The predictable operators are uniformly bounded and the noiseless $(U,S)$ direction decreases $\mathcal T$ by at least a fixed multiple of $\mathcal T$. The combined remainders have conditional first, second and fourth moments $O(\eta_y^2)$, $O(\eta_y^4)$ and $O(\eta_y^8)$, respectively. Deterministic actor increments are $\eta_y\operatorname{poly}(H)$ on sufficiently small prospective steps.
\end{lemma}

\begin{proof}
Proposition~\ref{glob:geometry} gives curvature and the two response
displacements. Lemmas~\ref{glob:tracking-regularity} and
\ref{lem:tau-scaling} prove the tracking self-bounds, full Hessian bound,
distance control and initialization estimate without any compact upper
domain. Corollary~\ref{glob:Phi} gives the objective bounds, and
Lemma~\ref{lem:tracked-field} gives the corrected-field estimate.
For the local updates, Lemmas~\ref{lem:prox-expansion} and
\ref{lem:pair-prox} prove the exact reference expansion, including its
inverse-metric correction. Proposition~\ref{klfin:prop:solver} and
Lemma~\ref{klfin:lem:paired-error} give the stated implementation errors;
no derivative of the numerical stopping rule is required.
Lemma~\ref{lem:population-coupled-descent}
proves descent of its population direction. The moment assertions follow
from Lemma~\ref{lem:conditional-movement} and the sampled-field bounds
in Lemma~\ref{critic:oracle}. These are moment conditions derived from
the sampling protocol, and do not assume accurate critic values.
\end{proof}

\subsection{Stable paired TD targets and derived accuracy}
\label{app:main-step-td}
For $p=(x,U,S)$, write $\alpha_0=U$, $\alpha_c=U+cS$ and
$z=(u,v)$, so the critics have coordinates $u$ and $u+cv$.
Let $q_{x,\alpha}$ and $Q^{\rm true}_{x,\alpha}$ have the convention in
\eqref{eq:regularized-value}: the true value is an unnormalized
discounted cost. The normalized state--action occupancy obeys
\begin{equation}
 \nu_U=(1-\gamma)\rho_U+\gamma\nu_UK_U.
 \label{critic:reset}
\end{equation}
Define the untruncated TD mean field (distinct from the cap $H$) by
\begin{align}
 \mathsf H(x,U,\alpha,\theta)
 &=\E_{\nu_U}\bigl[(Q_{m_Q,\theta}-q_{x,\alpha}
       -\gamma K_\alpha Q_{m_Q,\theta})\nabla_\theta Q_{m_Q,\theta}\bigr],\notag\\
 F(p,z)&=\begin{pmatrix}
 \mathsf H(x,U,U,u)\\
 [\mathsf H(x,U,U+cS,u+cv)-\mathsf H(x,U,U,u)]/c
 \end{pmatrix}.
 \label{critic:TD}
\end{align}
The operator $K_\alpha$ includes the next-state and next-action expectations. This is a TD semigradient field.
The sample field $\widehat{\mathsf H}$, with exact finite next-action sums,
is defined in \eqref{eq:algorithm-TD}; $U$ denotes its base sampling law
and has no additional pathwise role in that integrand.

The projection set and weighted norm are
$\mathcal K_Q=\overline B_{R_u}\times\overline B_{R_v}$ and
$\|z\|_W^2=M\|u\|^2+\|v\|^2$, with $W=\operatorname{diag}(MI_{d_Q},I_{d_Q})$.
The radii and $M>0$ are constructed below; their definition precedes
the projected update in \eqref{eq:algorithm-critic-update}.
Let $\ell_{\max}$ bound $|\log(\pi_\alpha/\piRef)|$ on the prospective
actor neighborhood and put $q_{\max}=B_r+\tau\ell_{\max}$.
\begin{proposition}[Global TD coercivity and interior roots]
\label{critic:TD-stability}
Set $\mu_0=(1-\sqrt\gamma)\lambda_Q$. Choose
\begin{equation}
 R_Q>4q_{\max}F_Q/\mu_0,
 \qquad \gamma F_Q^2L_\pi c_0s\le\mu_0/4,
 \label{critic:critic-choices}
\end{equation}
where $L_\pi$ is a uniform policy $\ell_1$-Lipschitz constant. There is a finite $m_{Q,0}$ such that, for every $m_Q\ge m_{Q,0}$, $\mathsf H(x,U,\alpha,\cdot)$ is strongly monotone on $\overline B_{R_Q}$ with modulus $\mu_Q=\mu_0/2$, and has a unique root $\theta^*(x,U,\alpha)$ with norm strictly less than $R_Q/2$. The bounds are global in $x$ and uniform in $m_Q,c$.
\end{proposition}
\begin{proof}
Lemma~\ref{lem:critic-calculus} gives the nonlinear critic derivative and linearization bounds in \eqref{critic:neural-bounds}.

For every scalar $v$,~\eqref{critic:reset} and Jensen imply
$\norm{K_Uv}_{L^2(\nu_U)}\le\gamma^{-1/2}\norm v_{L^2(\nu_U)}$.
Consequently
\begin{equation}
 \E_{\nu_U}v(v-\gamma K_Uv)
 \ge(1-\sqrt\gamma)\norm v_{L^2(\nu_U)}^2.
 \label{critic:coercivity}
\end{equation}
For a unit $b\in\R^{d_Q}$ set $v=b^\top\nabla Q_{m_Q,\theta}$. Differentiating the TD field shows that its quadratic Jacobian form equals
$\E v(v-\gamma K_\alpha v)$ plus the TD residual times $b^\top\nabla^2Qb$. The residual has absolute value at most
$D_Q=(1+\gamma)F_Q R_Q+q_{\max}$. Replacing $K_\alpha$ by $K_U$ costs at most $\gamma F_Q^2L_\pi c_0s$. Also
\[
 \E_{\nu_U}\nabla Q\nabla Q^\top
 \succeq\operatorname{sech}^4(F_Q R_Q/\sqrt{m_Q})\lambda_Q I.
\]
It follows that
\begin{equation}
 \operatorname{sym} D_\theta\mathsf H\succeq
 \left[\mu_0\operatorname{sech}^4(F_Q R_Q/\sqrt{m_Q})
 -\frac{2D_QF_Q^3R_Q}{m_Q}
 -\gamma F_Q^2L_\pi c_0s\right]I.
 \label{critic:mu-m}
\end{equation}
Choose $m_{Q,0}$ so the first two deviations from $\mu_0$ total at most $\mu_0/4$. Integration along line segments proves strong monotonicity with modulus $\mu_Q$.

At zero, $\norm{\mathsf H(x,U,\alpha,0)}\le q_{\max}F_Q$. A variational-inequality solution exists on the closed ball by applying Brouwer to a projected field step. At the boundary its outward radial field satisfies
$\ip{\theta}{\mathsf H(\theta)}\ge\mu_QR_Q^2-R_Qq_{\max}F_Q>0$,
which excludes a boundary variational-inequality solution by testing against zero. Thus the solution is an interior root, unique by strong monotonicity. Its norm is at most $q_{\max}F_Q/\mu_Q<R_Q/2$.
\end{proof}

With $\theta^*$ the unique TD root just obtained, define its paired target
\begin{equation}
 Z(p)=\left(\theta^*(x,U,U),
       \frac{\theta^*(x,U,U+cS)-\theta^*(x,U,U)}c\right)
 \label{critic:Z}
\end{equation}

\begin{proposition}[Uniform paired roots, projection set, and target derivatives]
\label{critic:pair-global}
There are fixed radii $R_u,R_v$ and a weight $M$ for which $R_u+c_0R_v<R_Q$, both endpoint critic parameters lie in the critic ball for every $z\in\CritSet$, and
the target $Z(p)$
lies strictly inside $\CritSet$ with uniform slack. After reducing $c_0$ if needed, $F(p,\cdot)$ is strongly monotone and Lipschitz in the $W$ norm, with constants uniform in $x,m_Q,c$. Its sample field is bounded uniformly in $c$. The target is twice continuously differentiable in $p$, with
\begin{equation}
 \norm{DZ}_W\le L_Z,\qquad\norm{D^2Z}_W\le H_Z
 \label{critic:Z-bounds}
\end{equation}
uniformly over $x$ and the prospective actor tube.
\end{proposition}
\begin{proof}
The inverse of $D_\theta\mathsf H$ has norm at most $\mu_Q^{-1}$. Implicit differentiation gives
\[
 D_\alpha\theta^*=-(D_\theta\mathsf H)^{-1}D_\alpha \mathsf H,
\]
uniformly bounded by Lemma~\ref{critic:anisotropic}. Integrating this derivative along $U+tS$ bounds the second component of~\eqref{critic:Z} by a constant times $s$. Choose $R_u$ strictly between the uniform base-root bound and $R_Q$, and $R_v$ strictly above that response bound. Reduce $c_0$ to get $R_u+c_0R_v<R_Q$. The root and derivative constants can first be bounded for the preliminary cap; reduction does not invalidate them.

Keeping the sample and base law fixed gives the integral identity
\begin{equation}
 \begin{aligned}
 &\frac{\widehat{\mathsf H}(x,U,U+cS,u+cv)-\widehat{\mathsf H}(x,U,U,u)}c\\
 &\quad=\int_0^1\bigl\{
 D_\alpha\widehat{\mathsf H}(x,U,U+tcS,u+tcv)[S]\\
 &\hspace{48mm}+D_\theta\widehat{\mathsf H}(x,U,U+tcS,u+tcv)[v]\bigr\}\,dt.
 \end{aligned}
 \label{critic:pair-integral}
\end{equation}
It bounds the sample field without dividing two unrelated endpoint estimates by $c$. Its mean has block Jacobian
\[
 F_z=\begin{pmatrix}
 D_\theta \mathsf H_0&0\\ B_c&D_\theta \mathsf H_c
 \end{pmatrix},\qquad
 B_c=[D_\theta \mathsf H_c-D_\theta \mathsf H_0]/c.
\]
The same integral formula bounds $\norm{B_c}\le C_B$. For increments $(a,b)$ its weighted quadratic form is at least
$M\mu_Q\norm a^2+\mu_Q\norm b^2-C_B\norm a\norm b$.
Choosing $M\ge1+C_B^2/\mu_Q^2$ gives strong monotonicity with modulus $\mu_F=\mu_Q/2$ in the $W$ norm. Bounded blocks give Lipschitz continuity, and the equations $F=0$ identify~\eqref{critic:Z}.

Twice differentiating the mean integral~\eqref{critic:pair-integral} uses derivatives of $\mathsf H$ of total order at most three, with at most two $x$ derivatives and at least one actor or critic derivative in each third-order block. Lemma~\ref{critic:anisotropic} is exactly sufficient, including the derivatives of the base law. Thus $F_p,F_z,F_{pp},F_{pz},F_{zz}$ have uniform bounds. Applying the implicit-function theorem in the open neighborhood of each interior target gives
\begin{align}
 DZ&=-F_z^{-1}F_p,\notag\\
 D^2Z[a,b]&=-F_z^{-1}\{F_{pp}[a,b]+F_{pz}[a,DZb]
              +F_{zp}[DZa,b]+F_{zz}[DZa,DZb]\}.
 \label{critic:IFT}
\end{align}
Strong monotonicity bounds the inverse by $\mu_F^{-1}$ in the weighted norm. For bounds $L_1$ on $F_p$ and $L_2$ on the second blocks, one may take $L_Z=L_1/\mu_F$ and
$H_Z=L_2(1+L_Z)^2/\mu_F$. Uniqueness identifies the local branches on overlaps for every $x\in\R^{d_x}$.
\end{proof}

\begin{proposition}[The global paired approximation floor is derived]
\label{critic:paired-floor}
Under C2, write $(u^*,v^*)=Z(p)$ and define the endpoint errors
\[
 e_0^*(p)=Q_{m_Q,u^*}-Q^{\rm true}_{x,U},\qquad
 e_c^*(p)=Q_{m_Q,u^*+cv^*}-Q^{\rm true}_{x,U+cS}.
\]
Then the squared paired approximation floor satisfies
\begin{equation}
 \AQ:=\sup_p\left\{
 \norm{e_0^*(p)}_\infty^2+
 \norm{\frac{e_c^*(p)-e_0^*(p)}{c}}_\infty^2
 \right\}\le C m_Q^{-2}.
 \label{critic:AQ}
\end{equation}
This bound and the required projected-root coefficient derivative bounds are uniform in $x$.
\end{proposition}
\begin{proof}
Fix $x,U,S$ and vary the endpoint $\alpha_t=U+tS$, $0\le t\le c_0$, keeping $\nu_U$ fixed. The linear-feature TD field is $A_t\theta-b_t$, with
\[
 A_t=\E_{\nu_U}[\varphi_Q(\varphi_Q-\gamma K_{\alpha_t}\varphi_Q)^\top],
 \qquad b_t=\E_{\nu_U}[\varphi_Q q_{x,\alpha_t}].
\]
Equation~\eqref{critic:coercivity} and the perturbation bound give
$\operatorname{sym} A_t\succeq(\mu_0/2)I$. Consequently its root $w_t=A_t^{-1}b_t$ is uniformly bounded and lies inside $B_{R_Q/2}$. All required coefficient derivatives follow by differentiating $A_tw_t=b_t$: $A_t$ has only actor dependence, and $b_t$ uses only the available reward derivatives and actor likelihood derivatives. In particular no smooth-coefficient hypothesis is needed in addition to C2. Exact realizability makes $w_t$ the true-value coefficient, since the Bellman equation implies the TD equation and the latter has a unique root.

Let $\mathsf H_t^{m_Q}$ be the nonlinear field, with root $\theta_t^{m_Q}$. From~\eqref{critic:neural-bounds}, uniformly over the critic ball,
\[
 \begin{aligned}
 &\norm{\mathsf H_t^{m_Q}-(A_t\theta-b_t)}
 +\norm{D_\theta \mathsf H_t^{m_Q}-A_t}\\
 &\qquad+\norm{\partial_t\mathsf H_t^{m_Q}-(\dot A_t\theta-\dot b_t)}\le C/m_Q.
 \end{aligned}
\]
Strong monotonicity implies $\norm{\theta_t^{m_Q}-w_t}\le C/m_Q$. Differentiating both root equations, with $B_t=D_\theta \mathsf H_t^{m_Q}(\theta_t^{m_Q})$, gives
\[
 B_t(\dot\theta_t^{m_Q}-\dot w_t)
 =-[\partial_t\mathsf H_t^{m_Q}(\theta_t^{m_Q})-(\dot A_tw_t-\dot b_t)]
                        -(B_t-A_t)\dot w_t.
\]
The inverse, coefficient, and derivative bounds imply
$\norm{\dot\theta_t^{m_Q}-\dot w_t}\le C/m_Q$. Therefore
$e_{m_Q}(t)=Q_{m_Q,\theta_t^{m_Q}}-\varphi_Q^\top w_t$ has
$\norm{e_{m_Q}(t)}_\infty+\norm{\dot e_{m_Q}(t)}_\infty\le C/m_Q$.
Integrating $\dot e_{m_Q}$ from $0$ to $c$ bounds the normalized error difference by $C/m_Q$. Squaring proves~\eqref{critic:AQ}. All constants were global in $x$ before this calculation.
\end{proof}

\subsection{Shared samples and persistent tracking}
\label{app:main-step-tracking}
Condition on a realized accepted metric sample, included in $\mathcal F_t$.
The three sampling routines in Appendix~\ref{app:support-sampling} use
the old $p_t,z_t$ and conditionally independent data blocks. The numerical
actor solvers use only the stored states and frozen lower block;
Lemma~\ref{klfin:lem:history} preserves the conditional independence from
the current TD block. All constants below are uniform over accepted
metric realizations. Write
$b_{\rm TD}=C_{\rm TD}\gamma^H$ and
$b_{\rm ev}=C_{\rm ev}(H+1)^\ell\gamma^H$ for uniform truncation-bias
bounds, with fixed constants and integer $\ell$ supplied there. Define
\begin{equation}
 \mathcal T_t=\mathcal T(p_t),\quad
 \mathcal C_t=\|z_t-Z(p_t)\|_W^2,\quad
 X_t=\|\nabla\Phi(x_t)\|^2,\quad
 \omega=c^2+\mathcal A_Q+b_{\rm ev}^2+b_{\rm TD}^2.
 \label{critic:errors}
\end{equation}
All conditional inequalities below hold at a pre-exit point and through
its prospective next step in the larger analytic actor neighborhood.
\begin{lemma}[Uniform oracle biases and moments]
\label{critic:oracle}
The squared conditional biases of the base, normalized-response, and corrected outer fields are at most
\begin{equation}
 C(\mathcal C_t+\AQ+b_{\rm ev}^2),\qquad
 b_{\rm ev}\le C(H+1)^\ell\gamma^H
 \label{critic:oracle-bias}
\end{equation}
for a fixed integer $\ell$. For the fixed-dynamics outer estimator its critic-dependent term can be omitted. Every fixed-order conditional sample moment is uniformly bounded in $H$; deterministic envelopes are polynomial in $H$. All statements are uniform in $x,m_Q,c$. Fixed positive batch sizes already satisfy the normalized moment requirements; TD/evaluation batches may grow at most polynomially in $H$, with fixed preference batch sizes.
\end{lemma}

\begin{proof}
Lemma~\ref{lem:paired-evaluation} controls the base and normalized
critic errors before likelihood weighting, then transfers them through
the shared-prefix estimators. Lemmas~\ref{lem:balance} and
\ref{lem:weight-difference} control those weights; Proposition~\ref{pref:oracle-means}
supplies unbiased bounded preference terms. Their geometric tail and
moment estimates prove the displayed bounds. Every data block uses the
old critic, so centering is conditional on the stated pre-draw history.
For fixed dynamics the outer field contains only reward derivatives and
direct preference terms, and hence no critic error.
\end{proof}
\begin{lemma}[Moving-target critic recursion with centered target movement]
\label{critic:moving}
For sufficiently small steps, $\eta_x\le\eta_y$, and $\eta_y/\beta$ sufficiently small, there are fixed positive $k,C$ such that
\begin{equation}
 \E_t\Delta \mathcal C_t\le-k\beta \mathcal C_t
       +C\frac{\eta_y^2}{\beta}\mathcal T_t+C\frac{\eta_x^2}{\beta}X_t
       +C\beta\omega+C(\beta^2+\eta_y^2+\eta_x^2).
 \label{critic:C-rec}
\end{equation}
The constants can be fixed uniformly over all sufficiently small ratios $\eta_y/\beta$ and $\eta_x/\eta_y$.
\end{lemma}
\begin{proof}
First hold the target at the old $p_t$. Projection is nonexpansive and $Z(p_t)\in\CritSet$. Strong monotonicity, Lipschitz continuity, bounded noise second moment, and~\eqref{critic:td-bias} give
\begin{equation}
 \E_t\norm{z_{t+1}-Z(p_t)}_W^2
 \le(1-a\beta)\mathcal C_t+C\beta b_{\rm TD}^2+C\beta^2
 \label{critic:frozen}
\end{equation}
for a fixed $a>0$. Indeed expansion of the projected squared distance gives the cross term $-2\beta\ip{z_t-Z}{F}$; absorb the bias cross term by Young's inequality, and the $\beta^2$ terms by a sufficiently small absolute step.

Write $\Delta p_t=d_t+w_{t+1}$, where $d_t=\E_t\Delta p_t$ and $\E_tw_{t+1}=0$. The current TD block is conditionally independent of this entire increment. Reveal that block first, and set $\bar e=z_{t+1}-Z(p_t)$, $J=DZ(p_t)$. Then $\bar e,J,d_t$ are fixed under the enlarged conditioning and $w_{t+1}$ stays centered. Taylor expansion gives
\[
 Z(p_t+\Delta p_t)-Z(p_t)=J(d_t+w_{t+1})+R_t,
 \qquad \norm{R_t}_W\le\tfrac12H_Z\norm{\Delta p_t}^2.
\]
Both critic points defining $\bar e$ lie in $\CritSet$, so $\norm{\bar e}_W$ is bounded by its diameter. Prospective increments have a fixed deterministic bound. The remainder cross terms and its square are consequently bounded by $C\norm{d_t+w_{t+1}}^2$. Centering cancels the linear cross term involving $w_{t+1}$. Use Young's inequality only for the deterministic displacement $Jd_t$, and combine with~\eqref{critic:frozen}, to obtain
\begin{equation}
 \E_t \mathcal C_{t+1}\le(1-a'\beta)\mathcal C_t+C\beta b_{\rm TD}^2+C\beta^2
       +\frac C\beta\norm{d_t}^2+C\E_t\norm{w_{t+1}}^2.
 \label{critic:moving-raw}
\end{equation}
In particular the centered target movement has no $1/\beta$ penalty.

Lemma~\ref{lem:conditional-movement} gives the conditional mean, second
moment and fourth moment estimates in \eqref{critic:movement}. It uses
the paired reference-map remainder, the samplewise finite-solver error
from Lemma~\ref{klfin:lem:paired-error}, and the trained-critic oracle bias,
including the fact that only the deterministic target drift is penalized
by $1/\beta$ in \eqref{critic:moving-raw}.

Substitute~\eqref{critic:movement} into~\eqref{critic:moving-raw}. Choose a fixed upper bound on $(\eta_y/\beta)^2$ small enough to absorb the self-coupling $C(\eta_y^2+\eta_x^2)\mathcal C_t/\beta$ into the negative drift. Since $\eta_x\le\eta_y\le\beta\le1$, the remaining terms have the sizes in~\eqref{critic:C-rec}. The constants before absorption are uniform over ratios at most one; fixing a sufficiently small upper bound and then reducing the ratios preserves them.
\end{proof}

\begin{lemma}[Actor tracking and outer descent with the trained critic]
\label{critic:recursions}
For sufficiently small steps and $\eta_x/\eta_y$ below a fixed threshold, there are fixed positive $a,b,d,u,v,w,C$ with
\begin{align}
 \E_t\Delta\Phi_t&\le-a\eta_x X_t+b\eta_x \mathcal T_t+d\eta_x \mathcal C_t+C\eta_x \omega+C\eta_x^2,
 \label{critic:Phi-rec}\\
 \E_t\Delta \mathcal T_t&\le-u\eta_y \mathcal T_t+v\eta_x X_t+w\eta_y \mathcal C_t
                              +C\eta_y\omega+C(\eta_y^2+\eta_x^2).
 \label{critic:A-rec}
\end{align}
The constants can be fixed uniformly over subsequently smaller step-size ratios.
\end{lemma}

\begin{proof}
Use the simultaneous Taylor estimate of Lemma~\ref{lem:tracking-taylor}.
The population actor direction decreases $\mathcal T$ by
$\rho_0\mathcal T$, where $\rho_0>0$ is fixed, and
$\|\nabla\mathcal T\|\le C\sqrt{\mathcal T}$ by
Lemma~\ref{glob:tracking-regularity}. Thus lower conditional bias adds at
most $C\eta_y\sqrt{\mathcal T_t}\sqrt{\mathcal C_t+\omega}$.
Young's inequality leaves $\rho_0\eta_y\mathcal T_t/4$
and $C\eta_y(\mathcal C_t+\omega)$. The mean local-map remainders
and all quadratic Taylor terms cost $C(\eta_y^2+\eta_x^2)$.
The finite numerical error contributes at most $C\eta_y^2$ because the
potential gradient is uniformly bounded on the prospective neighborhood.
Equivalently, Proposition~\ref{klfin:prop:transfer} transfers each
one-step inequality from the virtual exact step at the same old iterate.

The upper contribution is bounded by
$C\eta_x(\mathcal T_t+X_t+\mathcal C_t+\omega)$ because
$\|V_c\|\le C(\sqrt{X_t}+\sqrt{\mathcal T_t}+c)$.
Fix an upper bound on $q=\eta_x/\eta_y$ small enough to absorb its
$\mathcal T_t$ coefficient. Since $\eta_x\le\eta_y$, the remaining
terms give \eqref{critic:A-rec}, with constants unchanged when $q$ is
subsequently reduced. This calculation uses the full Hessian of the
tracking potential and requires no smoothness of the chosen metric
as a function of the iterates.

For the outer step, global $L_\Phi$-smoothness gives
\[
 \E_t\Delta\Phi_t\le
 -\eta_x\langle\nabla\Phi(x_t),V_c(p_t)+e_{V,t}\rangle
 +\tfrac12L_\Phi\eta_x^2\E_t\|\widehat V_t\|^2.
\]
Here $e_{V,t}=\E_t\widehat V_t-V_c(p_t)$.
Lemma~\ref{lem:tracked-field} bounds the squared secant error by
$C(\mathcal T_t+c^2)$, and Lemma~\ref{critic:oracle} bounds the
additional squared bias by $C(\mathcal C_t+\mathcal A_Q+b_{\rm ev}^2)$.
Young's inequality and bounded second moments yield \eqref{critic:Phi-rec}.
The actual fixed-dynamics outer estimator has no critic dependence;
the displayed $d\eta_x\mathcal C_t$ is a harmless conservative bound.
\end{proof}

\subsection{A contracting potential for containment}
\label{app:main-step-contraction}
The step ratios $r_Q=\eta_y/\beta$ and $q=\eta_x/\eta_y$ are positive
constants at most one. We first construct a contracting actor--critic
potential; the objective-containing potential used for stationarity is
defined separately in Step~\ref{app:main-step-stationarity}.
\begin{theorem}[Combined tracking contraction]
\label{critic:fast}
Choose $\lambda>0$ with $\lambda k>w$, and reduce the admissible upper bound on $r_Q$ so that $\lambda C r_Q^2\le u/2$ in~\eqref{critic:C-rec}. Define
\begin{equation}
 \mathcal D_t=\mathcal T_t+\lambda r_Q \mathcal C_t.
 \label{critic:D}
\end{equation}
There are $\rho_D,C_D>0$, uniform over subsequently smaller $r_Q,q$, such that
\begin{equation}
 \E_t\mathcal D_{t+1}\le(1-\rho_D\eta_y)\mathcal D_t+\eta_y D_\eps,
 \qquad D_\eps=C_D(qG_\Phi^2+\omega+\beta).
 \label{critic:D-rec}
\end{equation}
\end{theorem}
\begin{proof}
Add~\eqref{critic:A-rec} to $\lambda r_Q$ times~\eqref{critic:C-rec}. After dividing by $\eta_y$, the coefficients of $\mathcal T_t,\mathcal C_t,X_t$ are
\[
 -u+\lambda C r_Q^2,\qquad w-\lambda k,
 \qquad vq+\lambda Cq^2r_Q^2.
\]
Let $\delta_C=\lambda k-w>0$ and choose
$\rho_D=\min\{u/2,\delta_C/\lambda\}$. Since $r_Q\le1$,
the negative terms dominate $-\rho_D(\mathcal T_t+\lambda r_Q\mathcal C_t)$.
The $X_t$ coefficient is at most $(v+\lambda C)q$, and $X_t\le G_\Phi^2$. The bias terms are at most $C(1+\lambda)\omega$. The divided step remainders equal, up to a fixed factor,
\[
 \eta_y+\eta_x^2/\eta_y+\lambda(\beta+\eta_y^2/\beta+\eta_x^2/\beta)
 \le C(1+\lambda)\beta.
\]
Absorb these constants into $C_D$. No inverse power of a subsequently reduced ratio enters either $\rho_D$ or $C_D$.
The finite solver adds at most $C(1+\lambda r_Q)\eta_y^2$ to the
unweighted one-step comparison in Proposition~\ref{klfin:prop:transfer}.
After division by $\eta_y$, this is $O(\eta_y)\le O(\beta)$,
already included in $D_\eps$. Thus the recurrence concerns the
implemented process, with enlarged fixed remainder constants.
\end{proof}

\subsection{Constructively excluding a first exit}
\label{app:main-step-containment}
Use the actor radius $R_A=R$ and separation radius $s$ from Step~\ref{app:main-step-actor}.
The local-solve radius $r_{\rm prox}$ obeys \eqref{eq:prox-radius}. Put
\begin{equation}
 \begin{gathered}
 S_{\max}=s/2,\quad c_0s\le R_A/8,\quad r_{\rm prox}\le R_A/8,\\
 b_{\rm cap}=\min\{R_A/4,s/8\},\quad T_{\rm cap}=b_{\rm cap}^2\mu_\tau/E_*.
 \end{gathered}
 \label{critic:cap}
\end{equation}
The canonical observable certificate is
\begin{equation}
 \mathsf{Safe}(U,S):\quad \|S\|<s/2
 \quad\text{and}\quad \|U\|<R_A-c_0s-r_{\rm prox}.
 \label{eq:safe}
\end{equation}
Let $\tau_a=\inf\{k\ge0:\mathsf{Safe}(U_k,S_k)\text{ fails}\}$,
with infimum $+\infty$ for the empty set. The check includes $k=T$.
The larger analytic neighborhood is
$\|S\|<s$, $\|U\|<R_A-c_0s-r_{\rm prox}/2$; it contains the prospective
step from a safe point when the increments are sufficiently small.
There is no upper-domain exit test. Lemma~\ref{critic:D-innovation}
controls innovations through that prospective step, including projected
critic updates.
\begin{theorem}[Maximum bound through a prospective first exit]
\label{critic:D-max}
Let $\tau_a$ be the first completed iteration at which an actor certificate fails. Suppose~\eqref{critic:D-rec}--\eqref{critic:D-innovation-bound} hold before and through that prospective step, and $0<\rho_D\eta_y\le1$. With probability at least $1-\delta$,
\begin{equation}
 \max_{0\le k\le T\wedge\tau_a}\mathcal D_k
 \le\max\left\{\mathcal D_0,\frac{D_\eps}{\rho_D}\right\}
 +B_D(H)\sqrt{\frac{2\eta_y}{\rho_D}\log\frac{2T}{\delta}}.
 \label{critic:D-max-bound}
\end{equation}
\end{theorem}
\begin{proof}
Put $a=1-\rho_D\eta_y$ and multiply each pre-exit innovation by the predictable indicator $I_j=\mathbf 1_{\{j<\tau_a\}}$, setting it to zero afterward. For a deterministic $k$, on the event $k\le\tau_a$, iteration of the conditional inequality gives
\[
 \mathcal D_k\le a^k\mathcal D_0+\frac{D_\eps}{\rho_D}(1-a^k)
             +\sum_{j=0}^{k-1}a^{k-1-j}I_jN_{j+1}.
\]
The last sum is a weighted martingale sum for this fixed $k$. Its squared increment bounds sum to at most
$\eta_y^2B_D(H)^2\sum_{i\ge0}a^{2i}\le\eta_y B_D(H)^2/\rho_D$.
The bounded-increment exponential martingale inequality bounds its upper tail by $\delta/(2T)$ at the deviation in~\eqref{critic:D-max-bound}. A union bound over $k\le T$ gives the result (with slack in the failure probability). The first two deterministic terms are a convex combination. The comparison is asserted only through the exit; contraction of a frozen post-exit value is neither used nor needed.
\end{proof}

\begin{theorem}[Constructive containment for the trained critic]
\label{critic:containment}
Under Assumptions~\ref{ass:regularity}, \ref{ass:gap}, \ref{ass:global-upper}, \ref{ass:critic-features}, and~\ref{ass:critic-realizability}, condition on any realized accepted metric sample from Definition~\ref{klfin:construction:metric}. Use the finite actor solvers and specified critic and sampling algorithm, initialize $U_0=S_0=0$ and any deterministic $z_0\in\CritSet$. There are finite fixed choices of regularization, radii, weights, positive step ratios, and positive sampling/tuning constants, uniform over accepted metric realizations, such that, for all sufficiently small $\eps$,
\begin{equation}
 \beta=k_\beta\eps,\qquad \eta_y=r_Q\beta,\qquad \eta_x=q\eta_y,
 \quad c^2=k_c\eps,\qquad T=\lceil k_T\eps^{-2}\rceil,
 \label{critic:schedule}
\end{equation}
with $m_Q\ge\max\{m_{Q,0},C_m\eps^{-1/2}\}$ and logarithmic $H$, satisfies
\begin{equation}
 \Prb(\tau_a\le T\mid s_1,\ldots,s_n)\le\delta_a,
 \qquad \delta_a=\eps/2.
 \label{critic:containment-bound}
\end{equation}
The critic remains in $\CritSet$ by construction. No upper compactness, accurate critic initialization, or explicit augmented containment margin is assumed.
\end{theorem}
\begin{proof}
\textbf{1. Fix actor geometry before critic constants.}
The actor radius and tube are selected from the actor certificate. Increase the finite regularization $\tau$ until the actor curvature conditions hold and
\begin{equation}
 J_1/\mu_\tau<R_A/4,\quad G_f/\mu_\tau<s/4,
 \qquad \mathcal T_0<T_{\rm cap}/4.
 \label{critic:init-A}
\end{equation}
This is possible by~\eqref{critic:actor-cert}, since $\mathcal T_0=O(\tau^{-1})$ and $T_{\rm cap}=\Theta(\tau)$. Now keep $\tau$ fixed. C2 concerns the lower criterion at this selected value. Construct $R_Q,m_{Q,0},R_u,R_v,M$ as in Step~\ref{app:main-step-td}, reducing $c_0$ within the preliminary cap as needed. The resulting constants may depend on this $\tau$; the proof never increases $\tau$ afterward.

\textbf{2. Fix weights and then the ratios.}
Let $C_{\max}=\operatorname{diam}_W(\CritSet)^2$, which bounds $\mathcal C_0$ uniformly in $x,m_Q,c$. Fix the weight $\lambda$ from Theorem~\ref{critic:fast}. Also fix the stationarity weights $a_\Psi,b_\Psi$ from Proposition~\ref{critic:master} in Step~\ref{app:main-step-stationarity} before choosing ratios. Those weights depend only on the already fixed recursion constants, so this forward reference introduces no circular choice. Choose $r_Q>0$ below all critic and combined-drift thresholds, with
\begin{equation}
 \lambda r_Q C_{\max}<T_{\rm cap}/4.
 \label{critic:init-C}
\end{equation}
It follows that $\mathcal D_0<T_{\rm cap}/2$. This controls any bounded initial critic error without a preliminary training phase. Choose $q>0$ below both tracking and stationarity thresholds and so small that
\begin{equation}
 C_DqG_\Phi^2/\rho_D<T_{\rm cap}/4.
 \label{critic:q-choice}
\end{equation}
If $G_\Phi=0$, this last restriction is vacuous. The constants $C_D,\rho_D$ were fixed uniformly over smaller ratios, so these choices are noncircular.

\textbf{3. Make the remaining floor and fluctuations small.}
After the deterministic prospective-step justification in Step 4, the maximum theorem applies with the following choices. Choose any fixed positive $k_\beta,k_c,k_T$, and take $\eps$ sufficiently small that all absolute step restrictions and $c\le c_0$ hold. Proposition~\ref{critic:paired-floor} gives $\AQ=O(\eps)$ with the stated $m_Q$. Take
$H=O((1-\gamma)^{-1}\log(1/\eps))$ large enough that
$b_{\rm ev}^2+b_{\rm TD}^2=o(\eps)$, and use the normalized batches from Lemma~\ref{critic:oracle}. Hence $\omega+\beta=O(\eps)$ and
$D_\eps/\rho_D<T_{\rm cap}/2$ for sufficiently small $\eps$. With $\delta=\delta_a=\eps/2$ and the schedule~\eqref{critic:schedule}, the fluctuation term in~\eqref{critic:D-max-bound} tends to zero: it is $\sqrt\eps$ times a polynomial in $\log(1/\eps)$. It is eventually less than $T_{\rm cap}/4$.

\textbf{4. Justify the prospective step before using its inequality.}
At a pre-exit point,~\eqref{eq:safe} leaves room for $U+c\overline B_s$ and the local solves. The samplewise normalized-pair bounds imply
$\norm{\Delta U}<r_{\rm prox}/2$ and $\norm{\Delta S}<s/4$ when $\eps$ is sufficiently small, because each is at most $\eps\operatorname{poly}(H)$. Thus the prospective endpoint and every segment to it lie in the larger actor tube with $\norm S<s$ and with the endpoint policies in $B_{R_A}$. Choose the preliminary neighborhoods with strict extra slack if a closed ball is used in a derivative bound. All local-solver and target derivative estimates therefore apply to this step before it is known to remain in the smaller certified set. The critic update stays in $\CritSet$ by projection, and its target at the prospective endpoint lies in that same set by Proposition~\ref{critic:pair-global}. Upper segments require no buffer restriction because the estimates are global in $x$.
In detail, each numerical iterate remains in its buffered local ball by
projection. The virtual reference update has the stated deterministic
increment bound; the implemented update differs by at most
$\sqrt5\eta_y^2$ in $(U,S)$ by Lemma~\ref{klfin:lem:paired-error}.
This proves the prospective claim for both updates, and the segment
between them, without first assuming containment
(Lemma~\ref{klfin:lem:history}).

\textbf{5. Contradict the first exit.}
On the event from Theorem~\ref{critic:D-max}, the previous choices give $\mathcal D_k<T_{\rm cap}$ for every $k\le T\wedge\tau_a$. Since $\mathcal T_k\le\mathcal D_k$,~\eqref{critic:actor-cert} makes each actor error strictly smaller than $b_{\rm cap}$. At a candidate first exit this implies
\[
 \norm{U_k}<J_1/\mu_\tau+b_{\rm cap}<R_A/2,
 \qquad \norm{S_k}<G_f/\mu_\tau+b_{\rm cap}<3s/8<s/2.
\]
Also $c_0s+r_{\rm prox}\le R_A/4$, so the distance of $U_k$ from $\partial B_{R_A}$ is strictly greater than the required buffer. Both certificates hold strictly, a contradiction. Therefore this event excludes $\tau_a\le T$ and has conditional probability at least $1-\delta_a$. Averaging this uniform conditional bound and adding the independent metric rejection budget $\delta_{\rm met}=\eps/2$ gives total failure probability at most $\eps$.
\end{proof}

\subsection{Stationarity and complete data accounting}
\label{app:main-step-stationarity}
The contracting potential $\mathcal D$ proves containment. For stationarity
we use positive weights on the actor and critic errors and include the
lower-bounded selected objective.
\begin{proposition}[Augmented objective descent]
\label{critic:master}
Choose $a_\Psi,b_\Psi>0$ with $a_\Psi u>b+1$ and $b_\Psi k>d+a_\Psi w+1$, using the constants in~\eqref{critic:Phi-rec}--\eqref{critic:A-rec}. For sufficiently small fixed $r_Q,q$,
\begin{equation}
 \Psi_t=\Phi(x_t)-\underline\Phi+a_\Psi q \mathcal T_t+b_\Psi q r_Q \mathcal C_t
 \label{critic:Psi}
\end{equation}
satisfies
\begin{equation}
 \E_t\Delta\Psi_t\le-a_*\eta_x(X_t+\mathcal T_t+\mathcal C_t)
                                +C_*\eta_x(\omega+\beta+\eta_y+\eta_x)
 \label{critic:master-bound}
\end{equation}
through the prospective actor exit, with fixed positive constants.
\end{proposition}
\begin{proof}
Add~\eqref{critic:Phi-rec}, $a_\Psi \eta_x/\eta_y$ times~\eqref{critic:A-rec}, and $b_\Psi \eta_x/\beta$ times~\eqref{critic:C-rec}. After division by $\eta_x$, the three coefficients are bounded above by
\[
 -a+a_\Psi vq+b_\Psi Cq^2r_Q^2,\qquad
 b-a_\Psi u+b_\Psi C r_Q^2,\qquad d+a_\Psi w-b_\Psi k.
\]
Choose the weights first, then $r_Q$ to make $b_\Psi C r_Q^2<1/2$, and $q$ to make $a_\Psi vq+b_\Psi Cq^2r_Q^2<a/2$. These can be imposed simultaneously with the containment choices. Weighted remainder terms have the bound in~\eqref{critic:master-bound}.
\end{proof}

\begin{proof}[Proof of Theorem~\ref{thm:main-headline}]
Fix the actor initialization and retained chart, and take
$\delta_{\rm met}=\delta_a=\eps/2$. Proposition~\ref{klfin:prop:acceptance}
bounds metric rejection by $\delta_{\rm met}$. Conditional on any accepted
metric sample, Proposition~\ref{klfin:prop:Fisher} supplies the same
Fisher constants uniformly on the actor ball. All following constants
are independent of the accepted realization. The finite-solver errors
are already included in the tracking recursions by
Proposition~\ref{klfin:prop:transfer}. In particular their weighted
contribution to $\Psi$ is at most
$Cq\eta_y^2=C\eta_x\eta_y$, which is absorbed by
\eqref{critic:master-bound}.

For each such metric realization, stop at $\tau_a$ and freeze the process
afterward. Multiply~\eqref{critic:master-bound} by the predictable
pre-exit indicator and telescope, including the prospective exit step:
\[
 a_*\eta_x\sum_{t<T}\E[\mathbf1_{\{t<\tau_a\}}X_t\mid s_1,\ldots,s_n]
 \le\Psi_0+C_*\eta_x T(\omega+\beta+\eta_y+\eta_x).
\]
The terminal potential is nonnegative. For $t<T$,
$\mathbf1_{\{\tau_a>T\}}\le\mathbf1_{\{t<\tau_a\}}$;
Theorem~\ref{critic:containment} gives the uniform conditional bound
$\Prb(\tau_a\le T\mid s_1,\ldots,s_n)\le\delta_a$.
Draw the independent uniform output index on success and return $x_0$
on either failure. Averaging over the metric data and using
$\|\nabla\Phi(x_0)\|\le G_\Phi$ gives
\begin{equation}
 \E\norm{\nabla\Phi(x_{\rm out})}^2
 \le C\left[\frac{\Psi_0}{\eta_x T}+c^2+\AQ
               +b_{\rm ev}^2+b_{\rm TD}^2+\beta+\eta_y+\eta_x\right]
       +(\delta_{\rm met}+\delta_a)G_\Phi^2.
 \label{critic:rate}
\end{equation}
Here the initial potential is bounded uniformly over metric realizations:
$\Phi(x_0)-\underline\Phi\le2B_L$ (or $B_L$ for Bradley--Terry), and
$\mathcal T_0,\mathcal C_0$ have the fixed bounds proved above.
This argument conditions only on initial metric data and uses predictable
stopping; it never conditions a descent inequality on future successful
training. Metric acceptance reproduces the existing geometric constants,
so no enlargement of the C2 neighborhood or increase in fixed $\tau$
is needed. The union of metric rejection and actor exit has probability
at most $\delta_{\rm met}+\delta_a=\eps$.

Under schedule~\eqref{eq:main-schedule},
$m_Q\ge\max\{m_{Q,0},C_m\eps^{-1/2}\}$ gives $\AQ=O(\eps)$,
and logarithmic $H$ makes $b_{\rm ev}^2+b_{\rm TD}^2=o(\eps)$.
Also $(\eta_xT)^{-1}=O(\eps)$ and every remaining term has order
$O(\eps)$. This proves the claimed stationarity bound.

Let $B_C,B_H,B_V$ count TD, lower evaluation and outer discounted
sample draws per iteration, and $B_{f,a},B_{f,x}$ the tuple-batch sizes.
A TD or discounted evaluation draw uses at most $H+1$ transitions;
a tuple uses at most $IK$ transitions. The complete conservative ledger is
\begin{align}
 N_{\rm seg}&\le T[B_C+B_H+B_V+I(B_{f,a}+B_{f,x})]+n+N_{\rm pilot,seg},\notag\\
 N_{\rm trans}&\le T[(H+1)(B_C+B_H+B_V)
                   +IK(B_{f,a}+B_{f,x})]+n+N_{\rm pilot,trans}.
 \label{critic:cost}
\end{align}
The $n$ charges the one-time metric reset sample from
\eqref{klfin:eq:metric-size}. Each reset is conservatively charged as
one segment and at most one transition; when no transition follows the
reset its actual transition contribution is zero. Optional, separately
collected initialization-pilot data are added once; they are unnecessary
when A2 is assumed. The fixed-gap pilot in
Proposition~\ref{red:prop:pilot-certification}, if used, needs
$O(\log(1/\eps))$ draws. For fixed $\gamma,r,G_Z,\lambda_0,I,K$,
$n=O(\log(1/\eps))$. Discounted training batches are fixed or at most polynomial in $H$,
preference batches are fixed, and labels are charged for both preference
blocks. Thus $N_{\rm seg},N_{\rm trans}=\widetilde O(\eps^{-2})$
and $N_{\rm labels}=O(\eps^{-2})$; every persistent TD update is counted.
Reusing metric states during numerical optimization incurs no additional
environment interactions or preference labels.

There are at most two finite local solves per completed training iteration.
Proposition~\ref{klfin:prop:solver} therefore gives
\begin{equation}
 N_{\rm KL\text{-}grad}\le2T(J_{\max}+1)
 =O\!\left(\kappa_K\eps^{-2}\log(1/\eps)\right),
 \qquad \kappa_K=M_{K,L}/m_{K,L}.
 \label{klfin:eq:KL-gradient-count}
\end{equation}
Indeed $G_{\rm prox}(H)=\operatorname{poly}(H)$,
$H=O(\log(1/\eps))$ and $c\eta_y=\Theta(\eps^{3/2})$
in \eqref{klfin:eq:Jmax}, while $\kappa_K$ is independent of $n$ and
accuracy. A finite-sum gradient has at most $n|\mathcal A|$
state--action terms. If $C_{\rm act}(\mathbf m,r)$ bounds the arithmetic
for one actor term, including its first derivative, the numerical work is
\[
 O\!\left(T(J_{\max}+1)
       [n|\mathcal A|C_{\rm act}(\mathbf m,r)+r]\right).
\]
Constructing the metric additionally needs $n|\mathcal A|$ actor-feature
evaluations, covariance accumulation and one $r\times r$ spectral
certificate. These finite computations are separate from environment
interactions and from the other network arithmetic. Although endpoint
parameter accuracy is $c\eta_y^2=\Theta(\eps^{5/2})$, geometric convergence
of the local solver makes its additional iteration cost logarithmic;
it does not introduce an $\eps^{-5/2}$ sample exponent.
\end{proof}

\FloatBarrier
\section{Supporting Results}
\label{app:supporting-results}
These construction details and auxiliary results support the six-step
proof in Appendix~\ref{app:main-results}. Assumptions and the selected
target are those in the main text; no additional upper localization or
oracle-accuracy assumption is introduced for the main theorem. The optional
global-response results in Section~\ref{app:gpl-global-response} explicitly
state their additional conditions.
\subsection{Actor construction and initialization certificates}
\label{app:support-actor}\label{app:actor-construction}
The following normalized architecture supplies A1(iii). A2 is the separate
initialization covariance condition; hidden width alone does not imply it.
All construction constants are fixed before choosing $\tau$.
Fix the hidden depth $L$, input width $m_0=d_\psi$, and hidden widths
\[
  \mathbf m=(m_1,\ldots,m_L).
\]
For a branch parameter
\[
  \vartheta=(W^1,\ldots,W^L),
  \qquad W^\ell\in\R^{m_\ell\times m_{\ell-1}},
\]
define
\begin{align}
  h^0_\vartheta(s,a)&=\psi(s,a),
  \label{eq:h0}\\
  z^\ell_\vartheta(s,a)
  &=\frac{1}{\sqrt{m_{\ell-1}}}
    W^\ell h^{\ell-1}_\vartheta(s,a),
  \label{eq:preactivation}\\
  h^\ell_\vartheta(s,a)
  &=\sigma_\ell\bigl(z^\ell_\vartheta(s,a)\bigr),
  \qquad \ell=1,\ldots,L,
  \label{eq:activation}\\
  G_{\mathbf m,\vartheta}(s,a)
  &=\frac{1}{\sqrt{m_L}}b^\top h^L_\vartheta(s,a),
  \qquad b_j\in\{-1,+1\}.
  \label{eq:branch-energy}
\end{align}
The activations are applied coordinatewise and the readout $b$ is fixed.

Equip the branch parameter space with the product Frobenius norm
\begin{equation}
  \norm{\Delta\vartheta}_{\mathrm{par}}^2
  :=\sum_{\ell=1}^L\norm{\Delta W^\ell}_F^2.
  \label{eq:product-norm}
\end{equation}
Let $B_{\mathbf m}:\R^r\to\prod_{\ell=1}^L
\R^{m_\ell\times m_{\ell-1}}$ be an isometry:
\begin{equation}
  B_{\mathbf m}^\top B_{\mathbf m}=\Id_r.
  \label{eq:chart-isometry}
\end{equation}
For an initialization $\vartheta_{\mathbf m,0}$, define the two antithetic branches for every $\alpha\in\R^r$ by
\begin{equation}
  \vartheta_{\mathbf m}^{\pm}(\alpha)
  =\vartheta_{\mathbf m,0}
   \pm\frac{1}{\sqrt2}B_{\mathbf m}\alpha,
  \qquad \alpha\in\R^r.
  \label{eq:antithetic-branches}
\end{equation}
Fix an ambient radius $A>0$ and let $\overline B_A=\overline B(0,A)$. The geometry construction later selects $0<R_{\rm geom}\le A/2$; write
\begin{equation}
  \Theta_R:=\overline B(0,R_{\rm geom}).
  \label{eq:geometry-proof-tube}
\end{equation}
This closed ball is used only as a local geometry-certification tube.  It is not the actor parameter space.  The exact energy is
\begin{equation}
  F_{\mathbf m,\alpha}(s,a)
  =\frac{1}{\sqrt2}
   \left[
     G_{\mathbf m,\vartheta^+_{\mathbf m}(\alpha)}(s,a)
     -G_{\mathbf m,\vartheta^-_{\mathbf m}(\alpha)}(s,a)
   \right].
  \label{eq:deep-energy}
\end{equation}
This is an exact nonlinear network.  It can be implemented as one wider block-diagonal network with a fixed antisymmetric readout.  At initialization,
\begin{equation}
  F_{\mathbf m,0}(s,a)=0,\qquad D_\alpha^2F_{\mathbf m,0}(s,a)=0.
  \label{eq:zero-energy-init}
\end{equation}

The reference $\piRef(a\mid s)>0$ is fixed and independent of $x$. The policy is its finite exponential tilt:
\begin{equation}
  \frac{d\pi_{\mathbf m,\alpha}(\cdot\mid s)}
       {d\piRef(\cdot\mid s)}(a)
  =\exp\{F_{\mathbf m,\alpha}(s,a)-A_{\mathbf m,s}(\alpha)\},
  \label{eq:deep-policy}
\end{equation}
where
\begin{equation}
  A_{\mathbf m,s}(\alpha)
  =\log\sum_{a'\in\cA}e^{F_{\mathbf m,\alpha}(s,a')}
  \piRef(a'\mid s).
  \label{eq:log-partition}
\end{equation}
Define the effective energy feature and policy score
\begin{align}
  Z_{\mathbf m,\alpha}(s,a)
  &:=\grad_\alpha F_{\mathbf m,\alpha}(s,a),
  \label{eq:effective-feature}\\
  \overline Z_{\mathbf m,\alpha}(s,a)
  &:=Z_{\mathbf m,\alpha}(s,a)
  -\E_{a'\sim\pi_{\mathbf m,\alpha}(\cdot\mid s)}
   [Z_{\mathbf m,\alpha}(s,a')]
  =\grad_\alpha\log\pi_{\mathbf m,\alpha}(a\mid s).
  \label{eq:centered-feature}
\end{align}
For a state law $\nu$, define
\begin{equation}
  \Fisher_{\mathbf m,\nu}(\alpha)
  :=\E_{s\sim\nu,\,a\sim\pi_{\mathbf m,\alpha}(\cdot\mid s)}
  [\overline Z_{\mathbf m,\alpha}(s,a)
   \overline Z_{\mathbf m,\alpha}(s,a)^\top].
  \label{eq:fisher}
\end{equation}
Write $\Fisher_{x,\mathbf m}(\alpha)$ when $\nu=d_\alpha^\gamma$.

\paragraph{Construction requirements.}\label{ass:architecture}
The depth $L$, effective dimension $r$, and ambient radius $A$ are independent of the hidden widths.  The actor domain itself is $\R^r$.  The input satisfies $\norm{\psi(s,a)}\le B_\psi$.  Each activation is bounded and belongs to $C^4(\R)$, with
\begin{equation}
  \norm{\sigma_\ell^{(k)}}_\infty\le M_{\ell,k},
  \qquad k=0,1,2,3,4.
  \label{eq:activation-envelope}
\end{equation}
The initialized normalized layer operators satisfy
\begin{equation}
  \frac{\norm{W^\ell_{\mathbf m,0}}_\op}
       {\sqrt{m_{\ell-1}}}\le\kappa_\ell,
  \qquad \ell=1,\ldots,L,
  \label{eq:init-spectral}
\end{equation}
with width-independent constants.  The chart map is an isometry as in \eqref{eq:chart-isometry}.  If it is selected from data, use data independent of the sample used to certify its covariance gap. Conditional analysis for a fixed accepted chart is also possible.

\paragraph{Chart spectral control and antithetic identities}
For a branch point $\vartheta_{\mathbf m}^{\pm}(\alpha)$, set
\begin{equation}
  A^\ell_{\pm,\alpha}
  :=\frac{1}{\sqrt{m_{\ell-1}}}
    W^\ell_{\pm,\alpha}.
  \label{eq:normalized-layer}
\end{equation}

\begin{lemma}[Uniform normalized spectral bound on the chart]
\label{lem:chart-spectral}
Under \cref{ass:architecture}, on the ambient ball $\overline B_A$,
\begin{equation}
  \norm{A^\ell_{\pm,\alpha}}_\op
  \le\kappa_\ell+\frac{A}{\sqrt{2m_{\ell-1}}}
  \le\bar\kappa_\ell
  :=\kappa_\ell+\frac{A}{\sqrt2}
  \label{eq:chart-spectral}
\end{equation}
for every branch, chart point, and admissible width.
\end{lemma}

\begin{proof}
The layer-$\ell$ component of $B_{\mathbf m}\alpha$ has Frobenius norm at most
$\norm{B_{\mathbf m}\alpha}_{\mathrm{par}}=\norm{\alpha}\le A$.  Therefore its operator norm is at most $A$.  Divide the triangle inequality for
$W^\ell_{\mathbf m,0}\pm2^{-1/2}(B_{\mathbf m}\alpha)^\ell$
by $\sqrt{m_{\ell-1}}$ and use \eqref{eq:init-spectral}.
\end{proof}

\begin{lemma}[Antithetic initialization identities]
\label{lem:antithetic-identities}
The full two-branch chart is an isometry.  Moreover,
\begin{align}
  F_{\mathbf m,0}(s,a)&=0,
  \label{eq:antithetic-zero}\\
  Z_{\mathbf m,0}(s,a)
  &=B_{\mathbf m}^\top
    \grad_\vartheta G_{\mathbf m,\vartheta_{\mathbf m,0}}(s,a).
  \label{eq:antithetic-feature}
\end{align}
More generally, for directions $u_1,\ldots,u_k\in\R^r$,
\begin{align}
  D_\alpha^kF_{\mathbf m,\alpha}[u_1,\ldots,u_k]
  =2^{-(k+1)/2}\Bigl(&D_\vartheta^kG_{\vartheta^+(\alpha)}
       [B_{\mathbf m}u_1,\ldots,B_{\mathbf m}u_k]\notag\\
  &-(-1)^kD_\vartheta^kG_{\vartheta^-(\alpha)}
       [B_{\mathbf m}u_1,\ldots,B_{\mathbf m}u_k]\Bigr).
  \label{eq:antithetic-kth}
\end{align}
\end{lemma}

\begin{proof}
The map
\[
  u\longmapsto
  \bigl(2^{-1/2}B_{\mathbf m}u,
        -2^{-1/2}B_{\mathbf m}u\bigr)
\]
has squared product norm
\[
  \frac12\norm{B_{\mathbf m}u}_{\mathrm{par}}^2
  +\frac12\norm{B_{\mathbf m}u}_{\mathrm{par}}^2
  =\norm{u}^2.
\]
The two branch values agree at $\alpha=0$, proving \eqref{eq:antithetic-zero}.  Differentiating once gives \eqref{eq:antithetic-feature}; repeated differentiation and the chain rule give \eqref{eq:antithetic-kth}.
\end{proof}

\paragraph{A width-independent mixed-derivative recursion}
For $\ell=0,\ldots,L$, define
\begin{equation}
  q_0:=B_\psi/\sqrt{m_0},
  \qquad
  q_\ell:=M_{\ell,0}\quad(\ell\ge1).
  \label{eq:q-base}
\end{equation}
Set $H_{0,k}=0$ for $k=1,2,3,4$.  Recursively, for
$\ell=1,\ldots,L$, let
\begin{equation}
  \widehat H_{\ell-1,0}:=q_{\ell-1},
  \qquad
  \widehat H_{\ell-1,j}:=H_{\ell-1,j}
  \quad(j=1,2,3),
  \label{eq:Hhat}
\end{equation}
and
\begin{equation}
  Z_{\ell,k}
  :=\bar\kappa_\ell H_{\ell-1,k}
    +k\widehat H_{\ell-1,k-1},
  \qquad k=1,2,3,4.
  \label{eq:Z-recursion}
\end{equation}
Define
\begin{align}
  H_{\ell,1}
  &:=M_{\ell,1}Z_{\ell,1},
  \label{eq:H1}\\
  H_{\ell,2}
  &:=M_{\ell,1}Z_{\ell,2}
    +M_{\ell,2}Z_{\ell,1}^2,
  \label{eq:H2}\\
  H_{\ell,3}
  &:=M_{\ell,1}Z_{\ell,3}
    +3M_{\ell,2}Z_{\ell,1}Z_{\ell,2}
    +M_{\ell,3}Z_{\ell,1}^3,
  \label{eq:H3}\\
  H_{\ell,4}
  &:=M_{\ell,1}Z_{\ell,4}
    +4M_{\ell,2}Z_{\ell,1}Z_{\ell,3}
    +3M_{\ell,2}Z_{\ell,2}^2\notag\\
  &\quad
    +6M_{\ell,3}Z_{\ell,1}^2Z_{\ell,2}
    +M_{\ell,4}Z_{\ell,1}^4.
  \label{eq:H4}
\end{align}
All these numbers depend on fixed depth, the activation envelopes, the fixed ambient radius $A$, the input bound, and the normalized spectral envelopes, but not on the hidden widths.

\begin{lemma}[Exact mixed layer-derivative recursion]
\label{lem:mixed-recursion}
Fix a branch point and parameter directions
$\Delta_1\vartheta,\ldots,\Delta_k\vartheta$ with
$\norm{\Delta_i\vartheta}_{\mathrm{par}}\le1$.
For a nonempty subset $I\subseteq\{1,\ldots,k\}$, write
\begin{align*}
  h_I^\ell
  &=D_\vartheta^{|I|}h^\ell
    [\Delta_i\vartheta:i\in I],\\
  z_I^\ell
  &=D_\vartheta^{|I|}z^\ell
    [\Delta_i\vartheta:i\in I].
\end{align*}
Then
\begin{equation}
  z_I^\ell
  =A^\ell h_I^{\ell-1}
  +\frac{1}{\sqrt{m_{\ell-1}}}
   \sum_{i\in I}\Delta_iW^\ell h_{I\setminus\{i\}}^{\ell-1},
  \label{eq:exact-z-mixed}
\end{equation}
where $h_\varnothing^{\ell-1}=h^{\ell-1}$.  For $|I|=k\le4$,
\begin{equation}
  \norm{z_I^\ell}\le Z_{\ell,k},
  \qquad
  \norm{h_I^\ell}\le H_{\ell,k}.
  \label{eq:mixed-bounds}
\end{equation}
\end{lemma}

\begin{proof}
Because $W^\ell$ is affine in the parameter, at most one derivative can fall on it.  Applying the product rule to
$m_{\ell-1}^{-1/2}W^\ell h^{\ell-1}$ gives
\eqref{eq:exact-z-mixed} exactly.

For the norm bound, first note that
$\norm{\Delta_iW^\ell}_\op\le
 \norm{\Delta_iW^\ell}_F\le1$.
If $I\setminus\{i\}=\varnothing$, then
\[
  \frac{\norm{h^{\ell-1}}}{\sqrt{m_{\ell-1}}}
  \le q_{\ell-1};
\]
this follows from the input bound when $\ell=1$ and from bounded activation coordinates when $\ell\ge2$.
If $I\setminus\{i\}$ is nonempty, induction gives
$\norm{h_{I\setminus\{i\}}^{\ell-1}}\le
H_{\ell-1,k-1}$, and division by
$\sqrt{m_{\ell-1}}\ge1$ can only reduce the bound.  Together with
\cref{lem:chart-spectral}, this proves
$\norm{z_I^\ell}\le Z_{\ell,k}$.

The coordinatewise Fa\`a di Bruno formula is
\begin{equation}
  h_I^\ell
  =\sum_{\mathcal P\in\mathfrak P(I)}
  \sigma_\ell^{(|\mathcal P|)}(z^\ell)
  \odot\prod_{B\in\mathcal P}z_B^\ell,
  \label{eq:faa-partitions}
\end{equation}
where $\mathfrak P(I)$ is the set of partitions of $I$ and products are coordinatewise.  For vectors $v_1,\ldots,v_j$,
\begin{equation}
  \norm{v_1\odot\cdots\odot v_j}_2
  \le\prod_{i=1}^j\norm{v_i}_2.
  \label{eq:hadamard-bound}
\end{equation}
The partitions of sets of sizes one through four have block-size multiplicities
$1$;
$1,1$;
$1,3,1$;
and $1,4,3,6,1$, respectively.  Substituting the $Z$ bounds into
\eqref{eq:faa-partitions} gives exactly
\eqref{eq:H1}--\eqref{eq:H4}.  Induction over $\ell$ proves
\eqref{eq:mixed-bounds}.
\end{proof}

\begin{theorem}[Exact deep derivative bounds]
\label{thm:deep-derivatives}
Under \cref{ass:architecture}, define
\begin{equation}
  C_{k,L}:=2^{(1-k)/2}H_{L,k},
  \qquad k=1,2,3,4.
  \label{eq:Ck-definition}
\end{equation}
Then, uniformly over all admissible hidden widths, points of $\overline B_A$, and state-action pairs,
\begin{equation}
  \norm{D_\alpha^kF_{\mathbf m,\alpha}(s,a)}_\op
  \le C_{k,L},
  \qquad k=1,2,3,4.
  \label{eq:deep-derivative-bound}
\end{equation}
Consequently,
\begin{align}
  \norm{Z_{\mathbf m,\alpha}(s,a)}
  &\le C_{1,L},
  \label{eq:Z-bound}\\
  \norm{Z_{\mathbf m,\alpha}(s,a)
        -Z_{\mathbf m,\beta}(s,a)}
  &\le C_{2,L}\norm{\alpha-\beta},
  \label{eq:Z-drift}\\
  \abs{F_{\mathbf m,\alpha}(s,a)
       -F_{\mathbf m,\beta}(s,a)}
  &\le C_{1,L}\norm{\alpha-\beta}.
  \label{eq:energy-Lipschitz}
\end{align}
\end{theorem}

\begin{proof}
Apply \cref{lem:mixed-recursion} at either branch to unit product-norm directions.  Since $\norm{b}/\sqrt{m_L}=1$,
\[
  \abs{D_\vartheta^kG[\Delta_1\vartheta,\ldots,
  \Delta_k\vartheta]}
  \le H_{L,k}.
\]
For unit $u_i\in\R^r$, the isometry gives
$\norm{B_{\mathbf m}u_i}_{\mathrm{par}}=1$.
Now use \eqref{eq:antithetic-kth}; the sum of the two branch bounds is
$2\cdot2^{-(k+1)/2}H_{L,k}=C_{k,L}$.
The first consequence is the case $k=1$.  The remaining two follow by integrating the second and first derivatives, respectively, along the segment from $\beta$ to $\alpha$.
\end{proof}

\begin{corollary}[Uniform energy and density envelope]
\label{cor:energy-density}
For every point of the smaller geometry chart $\Theta_R=\overline B_R$,
\begin{align}
  \abs{F_{\mathbf m,\alpha}(s,a)}
  &\le C_{1,L}R,
  \label{eq:energy-envelope}\\
  e^{-2C_{1,L}R}
  \le\frac{d\pi_{\mathbf m,\alpha}(\cdot\mid s)}
  {d\piRef(\cdot\mid s)}(a)
  &\le e^{2C_{1,L}R},
  \label{eq:density-comparison}\\
  \left\|\log\frac{d\pi_{\mathbf m,\alpha}(\cdot\mid s)}
  {d\piRef(\cdot\mid s)}\right\|_{\infty,a}
  &\le2C_{1,L}R.
  \label{eq:log-density-envelope}
\end{align}
\end{corollary}

\begin{proof}
Use $F_{\mathbf m,0}=0$ and \eqref{eq:energy-Lipschitz}.  The log normalizer lies between the minimum and maximum of the energy, so both the numerator and normalizer comparisons contribute at most $C_{1,L}R$.
\end{proof}

\paragraph{Log-partition and log-policy derivatives}

For nonnegative numbers $c_1,\ldots,c_k$, define the complete exponential Bell polynomial
\begin{equation}
  \mathsf B_k(c_1,\ldots,c_k)
  :=\sum_{\substack{n_1,\ldots,n_k\ge0\\
          \sum_{j=1}^kjn_j=k}}
  \frac{k!}{\prod_{j=1}^kn_j!(j!)^{n_j}}
  \prod_{j=1}^kc_j^{n_j}.
  \label{eq:Bell-polynomial}
\end{equation}
Set $\mathsf B_0:=1$.

\begin{lemma}[Log-partition derivatives with explicit envelopes]
\label{lem:partition-derivatives}
Fix a state and a $C^4$ scalar path $t\mapsto F_t(a)$.  Suppose
$\norm{\partial_t^jF_t}_\infty\le c_j$ for $j=1,2,3,4$ on an interval.  Let
\[
 Z(t)=\int e^{F_t(a)}\piRef(da\mid s),
 \qquad A(t)=\log Z(t).
\]
Define $b_j=\mathsf B_j(c_1,\ldots,c_j)$ and recursively
\begin{equation}
  a_1:=b_1,
  \qquad
  a_k:=b_k+\sum_{j=1}^{k-1}
  \binom{k-1}{j-1}a_jb_{k-j},
  \quad k=2,3,4.
  \label{eq:log-partition-recursion}
\end{equation}
Then $|A^{(k)}(t)|\le a_k$ for $k=1,2,3,4$.
\end{lemma}

\begin{proof}
Fa\`a di Bruno's formula for $e^{F_t(a)}$ gives
\[
  \partial_t^ke^{F_t(a)}
  =e^{F_t(a)}\mathsf B_k
  (\partial_tF_t(a),\ldots,\partial_t^kF_t(a)).
\]
After integration and division by $Z(t)$,
$|Z^{(k)}(t)/Z(t)|\le b_k$.  Differentiating the identity
$Z'=A'Z$ $k-1$ additional times gives
\begin{equation}
  \frac{Z^{(k)}}{Z}
  =\sum_{j=1}^k\binom{k-1}{j-1}
  A^{(j)}\frac{Z^{(k-j)}}{Z}.
  \label{eq:Z-A-recurrence}
\end{equation}
Isolate the $j=k$ term and apply induction, the triangle inequality, and the bounds on $Z^{(q)}/Z$.  This yields exactly \eqref{eq:log-partition-recursion}.
\end{proof}

\begin{lemma}[Polarization bound]
\label{lem:polarization}
Let $T$ be a continuous symmetric $k$-linear form on a real Hilbert space and let
$p_T(h)=T[h,\ldots,h]$.  Then
\begin{equation}
  \norm{T}_\op
  \le\frac{k^k}{k!}
  \sup_{\norm h\le1}|p_T(h)|.
  \label{eq:polarization-bound}
\end{equation}
\end{lemma}
\begin{proof}
The real polarization identity is
\[
 T[h_1,\ldots,h_k]
 =\frac{1}{2^kk!}\sum_{\epsilon\in\{-1,1\}^k}
 \left(\prod_{i=1}^k\epsilon_i\right)
 p_T\!\left(\sum_{i=1}^k\epsilon_ih_i\right).
\]
For $\norm{h_i}\le1$, the argument of $p_T$ has norm at most $k$.  Homogeneity gives
$|p_T(\sum_i\epsilon_ih_i)|\le k^k\sup_{\norm h\le1}|p_T(h)|$.  Summing $2^k$ terms proves the claim.
\end{proof}

\begin{lemma}[Width-uniform deep log-policy derivatives]
\label{lem:log-policy}
For $k=1,2,3,4$, let $b_{k,L}$ and $a_{k,L}$ be obtained from
\eqref{eq:Bell-polynomial}--\eqref{eq:log-partition-recursion} with
$c_j=C_{j,L}$, and define
\begin{equation}
  M_{\log,k,L}
  :=\frac{k^k}{k!}\bigl(C_{k,L}+a_{k,L}\bigr).
  \label{eq:Mlog-rigorous}
\end{equation}
Then, uniformly over widths, states, actions, and chart points,
\begin{equation}
  \norm{D_\alpha^k
  \log\pi_{\mathbf m,\alpha}(a\mid s)}_\op
  \le M_{\log,k,L},
  \qquad k=1,2,3,4.
  \label{eq:log-derivative-bound}
\end{equation}
For every fixed state law $\nu$, the Fisher matrix is Lipschitz:
\begin{equation}
  \norm{\Fisher_{\mathbf m,\nu}(\alpha)
        -\Fisher_{\mathbf m,\nu}(\beta)}_\op
  \le L_{F,L}\norm{\alpha-\beta},
  \label{eq:fisher-lipschitz}
\end{equation}
where
\begin{equation}
  L_{F,L}:=2M_{\log,1,L}M_{\log,2,L}
  +M_{\log,1,L}^3.
  \label{eq:LF}
\end{equation}
Consequently, on the Fisher-conditioned region,
\begin{equation}
  \norm{\Fisher_{\mathbf m,\nu}(\alpha)^{-1}
        -\Fisher_{\mathbf m,\nu}(\beta)^{-1}}_\op
  \le L_{P,L}\norm{\alpha-\beta},
  \qquad
  L_{P,L}:=\frac{L_{F,L}}{\underline\lambda_{F,L}^2}.
  \label{eq:inverse-fisher-lipschitz}
\end{equation}
\end{lemma}

\begin{proof}
Fix a unit direction $h$ and apply \cref{lem:partition-derivatives} to
$F_t(a)=F_{\mathbf m,\alpha+th}(s,a)$.  The deep derivative theorem gives
$\norm{\partial_t^jF_t}_\infty\le C_{j,L}$.  Since
$\log\pi_{\alpha}=F_\alpha-A_s(\alpha)$, the diagonal $k$th directional derivative has magnitude at most $C_{k,L}+a_{k,L}$.  The derivative is a symmetric $k$-linear form, so \cref{lem:polarization} gives \eqref{eq:log-derivative-bound}.

Let $S_\alpha=\grad_\alpha\log\pi_{\mathbf m,\alpha}$.  For a unit direction $h$, differentiation of a policy expectation gives
\begin{align*}
  D_\alpha\E_{\nu,\pi_\alpha}[S_\alpha S_\alpha^\top][h]
  =\E_{\nu,\pi_\alpha}\bigl[&
  (DS_\alpha[h])S_\alpha^\top
  +S_\alpha(DS_\alpha[h])^\top\\
  &+S_\alpha S_\alpha^\top\ip{S_\alpha}{h}\bigr].
\end{align*}
The operator norm is at most $L_{F,L}$.  Integrating this derivative along the segment from $\beta$ to $\alpha$ proves \eqref{eq:fisher-lipschitz}.  Finally use \cref{lem:inverse-perturbation} and the Fisher lower bound.
\end{proof}

\paragraph{Reference-KL envelopes.}
Use the fixed ambient actor ball. Put $\chi_\gamma=\gamma/(1-\gamma)$ and $v_0=2C_1A$. Since $F_0=0$, $|F_\alpha|\le C_1A$ and $|\log(\pi_\alpha/\pi_{\mathrm{ref}})|\le v_0$.

Let $S_1,S_2,S_3$ be uniform bounds on the first three derivatives of $\log\pi_\alpha$. They follow from differentiating the finite softmax and the energy bounds. One conservative choice is
\[
S_1=2C_1,\qquad S_2=2C_2+C_1^2,\qquad
S_3=2C_3+6C_1C_2+8C_1^3.
\]
Define
\begin{align}
P_1&=S_1,&P_2&=S_2+S_1^2,&P_3&=S_3+3S_1S_2+S_1^3,\\
D_1&=\chi_\gamma P_1,&D_2&=\chi_\gamma P_2+2\chi_\gamma^2P_1^2,&
D_3&=\chi_\gamma P_3+6\chi_\gamma^2P_1P_2+6\chi_\gamma^3P_1^3.
\label{eq:occupancy-bounds}
\end{align}
Here the norm for signed measures is the dual of the supremum norm, without a factor $1/2$.

\begin{lemma}[Occupancy and KL envelopes]\label{lem:KL-envelope}
The total-variation norms of the first three policy derivatives are at most $P_i$, and those of the normalized state occupancy derivatives are at most $D_i$. Let
\begin{align*}
q_0&=v_0,&q_1&=S_1+P_1v_0,\\
q_2&=S_2+2P_1S_1+P_2v_0,&
q_3&=S_3+3P_1S_2+3P_2S_1+P_3v_0.
\end{align*}
Then
\begin{equation}
\norm{D_\alpha^3\mathcal K}\le L_K:=q_3+3D_1q_2+3D_2q_1+D_3q_0.
\label{eq:LK}
\end{equation}
In particular the KL Hessian is $L_K$-Lipschitz on actor segments in the ambient ball. All these constants are independent of $\tau$, width, and target accuracy.
\end{lemma}
\begin{proof}
Differentiate $\pi=e^{\log\pi}$ and sum absolute values over actions to obtain $P_i$. The actor-induced Markov kernel $T_\alpha$ is a contraction in the supremum norm, and $\norm{D^iT_\alpha}\le P_i$. The normalized occupancy is
\[
d_\alpha=(1-\gamma)\rho(I-\gamma T_\alpha)^{-1}.
\]
Differentiate its resolvent identity. The resolvent norm is at most $(1-\gamma)^{-1}$, while $\norm{d_\alpha}=1$. Product differentiation gives exactly the bounds $D_i$ in \eqref{eq:occupancy-bounds}; the coefficients count the ordered product-rule terms. The statewise KL is $k_\alpha(s)=\sum_a\pi_\alpha(a\mid s)\log(\pi_\alpha(a\mid s)/\pi_{\mathrm{ref}}(a\mid s))$. Product differentiation bounds its derivatives by $q_i$. Finally differentiate $\mathcal K=d_\alpha k_\alpha$ three times. The four groups of terms have bounds $q_3,3D_1q_2,3D_2q_1,D_3q_0$. All differentiations are justified by the finite action sums, derivative envelopes, and discounted resolvent. The argument does not require transition likelihood ratios for derivatives in $\alpha$.
\end{proof}

\begin{lemma}[Exact KL Hessian at initialization]\label{lem:reference}
With $a_0=(1-\gamma)\lambda_{\mathrm{eff},L}$, on the fixed actor chart,
\begin{equation}
\mathcal K(0)=0,\quad \nabla_\alpha \mathcal K(0)=0,\quad
\nabla_{\alpha\alpha}^2\mathcal K(0)=\Fisher_{x,\mathbf m}(0)\succeq a_0\Id,
\label{eq:reference}
\end{equation}
where $\Fisher_{x,\mathbf m}(0)$ is the occupancy-weighted policy Fisher matrix.
\end{lemma}
\begin{proof}
For every state $s$, $k_0(s)=0$ and $D k_0(s)=0$. At zero the score is
$Z_0(s,a)-\E_{\pi_{\mathrm{ref}}}Z_0(s,\cdot)$, so the information identity yields
$D^2k_0(s)=\operatorname{Cov}_{\pi_{\mathrm{ref}}}(Z_0(s,\cdot))$.
Differentiating $\mathcal K=d_\alpha k_\alpha$ twice leaves only $d_0D^2k_0$ because the terms multiplying $k_0$ and $Dk_0$ vanish statewise. Finally $d_0^\gamma\succeq(1-\gamma)\rho$ and \cref{ass:gap} give the lower bound. Normalization is essential: the bound contains $(1-\gamma)$, not its reciprocal.
\end{proof}

\begin{lemma}[Persistence of the restricted initialization covariance]
\label{lem:covariance-persistence}
Under \cref{ass:gap,thm:deep-derivatives}, for every $\alpha\in\Theta_R$,
\begin{equation}
  \E_{s\sim\rho}
  \Cov_{a\sim\piRef(\cdot\mid s)}
  \bigl(Z_{\mathbf m,\alpha}(s,a)\bigr)
  \succeq
  \bigl(\sqrt{\lambda_{\mathrm{eff},L}}
        -C_{2,L}\norm{\alpha}\bigr)_+^2\Id_r.
  \label{eq:ref-cov-persistence}
\end{equation}
\end{lemma}

\begin{proof}
Fix a unit vector $v\in\R^r$ and let the joint reference law be
$s\sim\rho$, $a\sim\piRef(\cdot\mid s)$.  Set
$X=v^\top Z_{\mathbf m,0}$ and
$Y=v^\top Z_{\mathbf m,\alpha}$.  By
\eqref{eq:Z-drift},
\[
  \norm{X-Y}_{L_2}\le C_{2,L}\norm{\alpha}.
\]
Conditional centering is a contraction, so
\begin{align*}
  \left\|Y-\E[Y\mid s]\right\|_{L_2}
  &\ge
  \left\|X-\E[X\mid s]\right\|_{L_2}
  -\norm{X-Y}_{L_2}\\
  &\ge\sqrt{\lambda_{\mathrm{eff},L}}
  -C_{2,L}\norm{\alpha},
\end{align*}
where the final inequality is \eqref{eq:main-population-gap}.  Since the left-hand side is nonnegative, take the positive part of the right-hand side before squaring, and then take the infimum over unit $v$.  This proves the claim.
\end{proof}

\begin{lemma}[Persistent discounted and common-state Fisher excitation]
\label{lem:fisher-persistence}
For the radius in \eqref{eq:R}, for every
$\alpha\in\Theta_R$ and every state law
$\nu\succeq(1-\gamma)\rho$,
\begin{equation}
  \underline\lambda_{F,L}\Id_r
  \preceq\Fisher_{\mathbf m,\nu}(\alpha)
  \preceq\overline\lambda_{F,L}\Id_r,
  \label{eq:fisher-bounds}
\end{equation}
where
\begin{equation}
 \underline\lambda_{F,L}=(1-\gamma)e^{-2C_{1,L}R}\lambda_{\rm eff,L}/4,
 \qquad \overline\lambda_{F,L}=4C_{1,L}^2.
 \label{eq:fisher-lower-summary}
\end{equation}  In particular, this holds for each discounted occupancy and for every common mixture law \eqref{eq:common-state-law}.
\end{lemma}

\begin{proof}
The upper bound follows from
$\norm{\overline Z_{\mathbf m,\alpha}}\le2C_{1,L}$.
For the lower bound, \cref{cor:energy-density} gives, conditionally on each state,
\[
  \pi_{\mathbf m,\alpha}(\cdot\mid s)
  \succeq e^{-2C_{1,L}R}\piRef(\cdot\mid s).
\]
Apply \cref{lem:cov-transfer} conditionally in the action variable to
$Z_{\mathbf m,\alpha}$, integrate with respect to
$\nu\succeq(1-\gamma)\rho$, and then use
\cref{lem:covariance-persistence}.  For every unit $v$,
\begin{align*}
  v^\top\Fisher_{\mathbf m,\nu}(\alpha)v
  &\ge(1-\gamma)e^{-2C_{1,L}R}
  \E_{s\sim\rho}
  \Var_{a\sim\piRef}
  \bigl(v^\top Z_{\mathbf m,\alpha}(s,a)\bigr)\\
  &\ge\underline\lambda_{F,L}.
\end{align*}
The common mixture dominates $(1-\gamma)\rho$ because each of its two components does.
\end{proof}

\paragraph{Optional one-time initialization certificate.}
\begin{proposition}[Independent pilot certification]
\label{red:prop:pilot-certification}
Fix a realized network and chart with $\|Z_{\mathbf m,0}\|\le G_Z:=C_{1,L}$. Draw independent pilot triples
\[
 s_i\sim\rho,
 \qquad a_i,a_i'\stackrel{\mathrm{i.i.d.}}{\sim}\piRef(\cdot\mid s_i),
 \qquad i=1,\ldots,N,
\]
and define
\begin{equation}
 \widehat{\cG}_{N}
 =\frac{1}{2N}\sum_{i=1}^{N}
 (Z_i-Z_i')(Z_i-Z_i')^\top,
 \label{red:eq:pilot-covariance}
\end{equation}
where $Z_i=Z_{\mathbf m,0}(s_i,a_i)$ and $Z_i'=Z_{\mathbf m,0}(s_i,a_i')$. Then
\begin{equation}
 \E[\widehat{\cG}_N\mid\mathbf m,\text{initialization}]
 =\cG_{\mathbf m,0}.
 \label{red:eq:pilot-unbiased}
\end{equation}
Moreover, there is a universal constant $C_{\mathrm B}$ such that, with probability at least $1-\delta$,
\begin{equation}
 \norm{\widehat{\cG}_N-\cG_{\mathbf m,0}}_\op
 \le
 C_{\mathrm B}G_Z^2
 \left[
  \sqrt{\frac{\log(2r/\delta)}{N}}
  +\frac{\log(2r/\delta)}{N}
 \right]
 =:\epsilon_N.
 \label{red:eq:pilot-concentration}
\end{equation}
Consequently,
\begin{equation}
 \lambda_{\min}(\widehat{\cG}_N)
 \ge\lambda_{\mathrm{cert}}+\epsilon_N
 \label{red:eq:pilot-acceptance}
\end{equation}
certifies $\cG_{\mathbf m,0}\succeq\lambda_{\mathrm{cert}}\Id_r$ with probability at least $1-\delta$.
\end{proposition}

\begin{proof}
Conditionally on $s_i$, the two actions are independent and identically distributed, so
\[
 \frac12\E[(Z_i-Z_i')(Z_i-Z_i')^\top\mid s_i]
 =\Cov_{a\sim\piRef(\cdot\mid s_i)}(Z_{\mathbf m,0}(s_i,a)).
\]
Averaging over $s_i$ proves \eqref{red:eq:pilot-unbiased}. Each summand in \eqref{red:eq:pilot-covariance} is positive semidefinite and has operator norm at most $2G_Z^2$. Self-adjoint matrix Bernstein gives \eqref{red:eq:pilot-concentration} after changing a universal numerical constant. Weyl's inequality then proves the certification statement.
\end{proof}

The chart must be fixed independently of the certification sample. Chart
selection from data therefore requires a separate certificate split.
The proposition bounds false acceptance by $\delta$; it does not assert
that every chart is accepted. For a fixed positive gap and slack, the
sample size grows only logarithmically in $1/\delta$. Charge any collected
reset/pilot data in \eqref{critic:cost}. A2 can instead be assumed directly.
The actor Fisher certificate is distinct from C1's critic feature Gram gap.

\subsection{Preference losses and global mixed-derivative bounds}
\label{app:support-preference}
\paragraph{Bradley--Terry losses satisfy explicit global envelopes}
For two trajectories and a realized binary label $Y$, put
\begin{equation}
 \Delta_x=\sum_{t=0}^{K-1}r_x(s_t^1,a_t^1)
          -\sum_{t=0}^{K-1}r_x(s_t^2,a_t^2),\qquad
 L=\log(1+e^{\Delta_x})-Y\Delta_x.
 \label{glob:BT-loss}
\end{equation}
The conditional law of $Y$ given the trajectories is fixed when
differentiating. In particular, it is not a reward-model label distribution
whose parameter dependence has been omitted.

\begin{proposition}[Samplewise preference-loss envelopes]
\label{glob:BT}
If $|r_x|\le B_r$, $\norm{\nabla_xr_x}\le G_r$, and
$\norm{\nabla_x^2r_x}_\op\le L_r$ uniformly, then
\eqref{glob:BT-loss} obeys
\begin{equation}
 \begin{gathered}
 0\le L\le B_L:=\log2+2KB_r,\qquad
 \norm{\nabla_xL}\le G_L:=2KG_r,\\
 \norm{\nabla_x^2L}_\op\le L_L:=K^2G_r^2+2KL_r.
 \end{gathered}
 \label{glob:BT-envelopes}
\end{equation}
These bounds hold for each realized label, and $\Phi_\tau\ge0$.
\end{proposition}
\begin{proof}
The corresponding bounds for $\Delta_x$, its gradient, and its Hessian
are $2KB_r$, $2KG_r$, and $2KL_r$. For
$\sigma(z)=(1+e^{-z})^{-1}$,
\begin{align*}
 \nabla_xL&=(\sigma(\Delta_x)-Y)\nabla_x\Delta_x,\\
 \nabla_x^2L&=\sigma'(\Delta_x)\nabla_x\Delta_x
                  \nabla_x\Delta_x^\top
       +(\sigma(\Delta_x)-Y)\nabla_x^2\Delta_x.
\end{align*}
Use $|\sigma-Y|\le1$, $0\le\sigma'\le1/4$, and
$0\le L\le\log2+|\Delta_x|$. Taking the policy expectation preserves
nonnegativity.
\end{proof}

For example, $r_x(s,a)=B\tanh(x^\top\varphi(s,a))$ with
$\sup_{s,a}\norm{\varphi(s,a)}\le C_\varphi$ satisfies the assumptions
on unrestricted $x$, with $B_r=B$, $G_r=BC_\varphi$, and
$L_r=2BC_\varphi^2$. Its higher derivatives also exist and are locally
uniformly bounded. Bounded output alone does not give the derivative
envelopes for a generic reward network with every weight trainable.
For a fixed offline comparison dataset independent of $\alpha$, this
preference term instead has $f_\alpha=0$; the policy-generated objective
in the theorem retains the response-chain contribution.

\paragraph{The mixed derivatives used by the proof}
Let $M_q$ be the width-independent actor log-policy derivative bounds from
Lemma~\ref{lem:log-policy}, for $q=1,\ldots,4$, on the ambient actor
neighborhood. Define the likelihood-derivative polynomials
\begin{align*}
 b_0(n)&=1,\qquad b_1(n)=nM_1,\\
 b_2(n)&=nM_2+n^2M_1^2,\\
 b_3(n)&=nM_3+3n^2M_1M_2+n^3M_1^3,\\
 b_4(n)&=nM_4+4n^2M_1M_3+3n^2M_2^2
              +6n^3M_1^2M_2+n^4M_1^4.
\end{align*}
Set $A_q(\gamma)=(1-\gamma)\sum_{t\ge0}\gamma^t b_q(t+1)$,
$(R_0,R_1,R_2)=(B_r,G_r,L_r)$, and
$(F_0,F_1,F_2)=(B_L,G_L,L_L)$.

\begin{lemma}[Uniform mixed-derivative bounds on unrestricted $x$]
\label{glob:mixed}
For $0\le p\le2$ and $0\le q\le4$, the separated mixed derivatives
exist and satisfy
\begin{align}
 \norm{D_x^pD_\alpha^q j(x,\alpha)}_\op
 &\le R_p A_q(\gamma),\label{glob:j-derivatives}\\
 \norm{D_x^pD_\alpha^q f(x,\alpha)}_\op
 &\le F_p b_q(IK),\label{glob:f-derivatives}
\end{align}
uniformly in $x$ and width. For $p\ge1$, the corresponding derivatives
of $g_\tau=-j+\tau\mathcal K$ equal those of $-j$; for $p=0$, add
$\tau$ times the pure actor KL envelope. In particular, with
$M_0=2C_{1,L}A$ and $M_i=M_{\log,i,L}$ for $i\ge1$,
\begin{equation}
 \|D_\alpha^q\mathcal K(\alpha)\|_{\rm op}
 \le (1-\gamma)\sum_{t\ge0}\gamma^t
       \sum_{j=0}^q\binom qj b_j(t+1)M_{q-j}<\infty,
 \quad 0\le q\le4.
 \label{glob:KL-derivatives}
\end{equation}
No global bounds on pure
third or fourth $x$ derivatives are claimed.
\end{lemma}
\begin{proof}
The trajectory likelihood has log likelihood equal to a sum of $n$
log-policy terms plus a parameter-independent term. Differentiating its
exponential $q$ times gives sums over set partitions; taking operator
norms yields exactly $b_q(n)$ as a bound on the likelihood derivative
divided by the likelihood. Fixed dynamics imply that $x$ derivatives act
only on $r_x$ or $L$, while actor derivatives act only on likelihoods.
The feedback kernel has no derivative. For the reward at time $t$ there
are $t+1$ action factors, and the upper tuple has $IK$ action factors.
The global bounds give domination, and each $A_q(\gamma)$ is finite
because a geometric sequence times a fixed-degree polynomial is
summable. This justifies differentiation under the expectations and the
discounted sum. The same separated calculation gives mixed derivatives
with $p+q>4$ when needed, using only $p$ primitive $x$ derivatives and
$q$ actor derivatives; it does not assume joint global $C^6$ bounds.
Finally, $\mathcal K$ is independent of $x$. For its time-$t$ contribution,
apply the product rule to the prefix likelihood and the terminal
log-policy-ratio criterion. Assigning $j$ derivatives to the likelihood
gives $b_j(t+1)$; the remaining derivatives of that criterion are bounded
by $M_{q-j}$. Summing the $\binom qj$ assignments and the discounted
series proves \eqref{glob:KL-derivatives}, including the fourth actor
derivative used by the tracking Hessian.
\end{proof}

For clarity, the largest derivative blocks used in the argument are
listed below; their lower-order counterparts are also used.
\begin{center}
\begin{tabularx}{\linewidth}{@{}p{0.29\linewidth}X@{}}
\toprule
Proof component & Derivative blocks\\
\midrule
Curvature and responses &
$g_{\alpha\alpha},g_{\alpha\alpha\alpha},g_{\alpha x},
 f_\alpha,f_{\alpha\alpha}$\\
Target smoothness &
$g_{\alpha xx},g_{\alpha\alpha x},g_{\alpha\alpha\alpha}$;
 all second derivatives of $f$\\
Response secant &
$g_{x\alpha},g_{x\alpha\alpha},g_{\alpha\alpha\alpha},
 f_{x\alpha},f_{\alpha\alpha},f_{\alpha\alpha\alpha}$\\
Full tracking Hessian &
$g_{\alpha\alpha xx},g_{\alpha\alpha\alpha x},
 g_{\alpha\alpha\alpha\alpha},f_{\alpha xx},
 f_{\alpha\alpha x},f_{\alpha\alpha\alpha}$\\
Common-KL update & Actor log-policy and KL derivatives through order four\\
\bottomrule
\end{tabularx}
\end{center}
A1's compact-set $C^4$ statement is not invoked as a global bound. The needed
discounted derivative tails are bounded by
$(1-\gamma)\sum_{t\ge H}\gamma^t R_p b_q(t+1)$.
They therefore admit the logarithmic truncation schedule. The length
$K$ of the preference objective is fixed by its definition and introduces
no discounted truncation bias.

\subsection{Response and tracking-potential calculus}
\label{app:support-response}
This subsection supplies the local-root and gap calculations used in
Step~\ref{app:main-step-actor}. Its upper-domain bounds use only the
separated derivative blocks in Lemma~\ref{glob:mixed}.

\begin{lemma}[Integral secants are uniform in the perturbation]
\label{lem:integral-secant}
On any segment in the analytic actor neighborhood,
\begin{equation}
 \frac{G(x,U+cS)-G(x,U)}c
 =\int_0^1D_\alpha G(x,U+tcS)[S]\,dt.
 \label{eq:secant-integral}
\end{equation}
If $D_\alpha G$ is bounded and Lipschitz in the variables being compared,
then the secant is Lipschitz in those variables and in bounded $S$,
uniformly for $0<c\le c_0$. Its continuous value at zero is
$D_\alpha G(x,U)[S]$. A Lipschitz constant $L_G$ for $D_\alpha G$
also gives error at most $L_Gc\|S\|^2/2$ from this zero-perturbation value.
\end{lemma}
\begin{proof}
Apply the fundamental theorem of calculus to $t\mapsto G(x,U+tcS)$.
For two triples, subtract the integral expressions and add and subtract
the derivative at one common point. Its bounded norm controls $S-S'$;
its Lipschitz bound controls the remaining difference, using
$\|U+tcS-U'-tcS'\|\le\|U-U'\|+c_0\|S-S'\|$.
Subtracting $D_\alpha G(x,U)[S]$ and integrating $t$ proves the last bound.
\end{proof}

\begin{lemma}[Uniformly Lipschitz population residual]
\label{lem:R-lipschitz}
On the prospective actor neighborhood, $R_c$ is globally Lipschitz in
$x$ and locally uniformly Lipschitz in $(U,S)$, with constants independent
of $c$ and width. At the selected response pair,
\begin{equation}
 R_c(x,\alpha_0^*(x),S_c^*(x))=0.
 \label{eq:R-zero}
\end{equation}
\end{lemma}
\begin{proof}
Use Lemma~\ref{lem:integral-secant} for $G=g_\alpha$ and add
$f_\alpha(x,U+cS)$. The derivative blocks $g_{\alpha\alpha x}$,
$g_{\alpha\alpha\alpha}$, $f_{\alpha x}$ and
$f_{\alpha\alpha}$ are uniformly bounded by Lemma~\ref{glob:mixed}.
Subtracting the base and perturbed response equations gives
\eqref{eq:R-zero}. The segment between any two admissible $U$ values
preserves the proof-ball inclusion by convexity.
\end{proof}

The scaled potential from \eqref{eq:scaled-objective} has the exact integral form
\begin{align}
 \mathcal H_c(x,U,S)
 &=\int_0^1(1-t)S^\top g_{\alpha\alpha}(x,U+tcS)S\,dt\notag\\
 &\quad+\int_0^1\langle f_\alpha(x,U+tcS),S\rangle\,dt.
 \label{glob:H-integral}
\end{align}
This formula defines its continuous extension at $c=0$ without a
divergent additive constant.

\begin{proposition}[Interior root of the scaled residual]
\label{red:prop:root-existence}
If $U+c\overline B_s\subset B_R$ and $G_f/\mu_\tau<s$, then
$\mathcal H_c(x,U,\cdot)$ has a unique interior minimizer
$\bar S_c(x,U)\in B_s$, satisfying
\begin{equation}
 R_c(x,U,\bar S_c)=0,\qquad \|\bar S_c(x,U)\|\le G_f/\mu_\tau,
 \qquad \bar S_c(x,\alpha_0^*(x))=S_c^*(x).
 \label{eq:Sbar-target}
\end{equation}
These statements hold uniformly in $x$ and extend to $c=0$.
\end{proposition}
\begin{proof}
Direct differentiation gives
\begin{equation}
 \nabla_S\mathcal H_c=R_c,\qquad
 D_SR_c=h_{c,\alpha\alpha}(x,U+cS)\succeq\mu_\tau I.
 \label{eq:scaled-hessian}
\end{equation}
At $S=0$, $R_c=f_\alpha(x,U)$. On the boundary $\|S\|=s$,
$\langle R_c,S\rangle\ge\mu_\tau s^2-G_fs>0$.
Thus a minimizer on the compact ball cannot lie on its boundary.
Strong convexity gives uniqueness; comparing its zero gradient with
$R_c(x,U,0)$ gives the norm bound. Equation~\eqref{eq:R-zero}
and uniqueness identify the response-pair value. All arguments follow
from the continuous integral formula at $c=0$ as well.
\end{proof}

\begin{lemma}[Global gap regularity and self-bounds]
\label{glob:tracking-regularity}\label{lem:tracking-smooth}
Fix sufficiently large finite $\tau$ and any fixed $\lambda_S>0$.
On the prospective actor neighborhood, $\bar S_c$ is continuously
differentiable, and $\mathcal T$ is twice continuously differentiable.
The following bounds are uniform in $x,c$ and width:
\begin{gather}
 \tfrac12\mu_\tau\|U-\alpha_0^*\|^2\le\Delta_0,
 \qquad \tfrac12\mu_\tau\|S-\bar S_c\|^2\le\Gamma_c,
 \label{eq:gap-distance}\\
 \|a\|^2\ge2\mu_\tau\Delta_0,
 \qquad \|R_c\|^2\ge2\mu_\tau\Gamma_c,
 \qquad \|a\|^2+\|R_c\|^2\le C\mathcal T,
 \label{eq:gap-gradient}\\
 \|\nabla_{(x,U,S)}\mathcal T\|^2\le C\mathcal T,
 \qquad \|D^2_{(x,U,S)}\mathcal T\|_{\rm op}\le C.
 \label{eq:T-full-Hessian-bound}
\end{gather}
The gap values and their first derivatives are bounded on this
neighborhood. These constants may depend on the fixed $\tau$.
\end{lemma}
\begin{proof}
Actor Hessian upper bounds and Lemma~\ref{lem:strong-smooth-gradient}
give the distance and gradient inequalities for both gaps.
The implicit-function theorem applied to $R_c(x,U,\bar S_c)=0$ gives
\[
 D_y\bar S_c=-(D_SR_c)^{-1}D_yR_c,\qquad y=(x,U).
\]
The inverse norm is at most $\mu_\tau^{-1}$. Two derivatives of
\eqref{glob:H-integral} use only
$g_{\alpha\alpha xx},g_{\alpha\alpha\alpha x},
g_{\alpha\alpha\alpha\alpha}$ and
$f_{\alpha xx},f_{\alpha\alpha x},f_{\alpha\alpha\alpha}$,
together with lower-order blocks. Lemma~\ref{glob:mixed} supplies
uniform bounds for exactly these derivatives. Hence $D_y\bar S_c$
and the full Hessian of $\mathcal H_c$ are bounded.

For $y=x,U$, the envelope identities are
\[
 D_y\Gamma_c=D_y\mathcal H_c(x,U,S)
             -D_y\mathcal H_c(x,U,\bar S_c),\qquad
 D_x\Delta_0=g_x(x,U)-g_x(x,\alpha_0^*).
\]
The first difference has norm at most $C\|S-\bar S_c\|$;
the second at most $C\|U-\alpha_0^*\|$. Together with
$D_U\Delta_0=a$, $D_S\Gamma_c=R_c$ and the gap inequalities,
these prove the gradient self-bound. Differentiating the envelope
identities once uses only $D^2\mathcal H_c$, $D\bar S_c$ and
$D_x\alpha_0^*$, whose bound is in Corollary~\ref{glob:Phi}.
This proves the full Hessian bound without differentiating the minimizing
separation twice. Upper Hessian bounds times squared actor distances
bound the gap values. No pure third or fourth global $x$ derivative is used.
\end{proof}

\begin{lemma}[Regularization-uniform weight, distance control and initialization]
\label{lem:tau-scaling}\label{lem:tracking-controls}
Fix the actor and separation neighborhoods before $\tau$, and let
$p_-=\overline\lambda_{F,L}^{-1}$,
$p_+=\underline\lambda_{F,L}^{-1}$ and $K_{\max}=L_{P,L}s$.
There are $d_*,E_*>0$ and a choice $\lambda_S>0$, independent of
sufficiently large $\tau$, such that
\begin{equation}
 \|U-\alpha_0^*(x)\|^2+\|S-S_c^*(x)\|^2
 \le\frac{E_*}{\mu_\tau}\mathcal T(x,U,S),\qquad
 \mathcal T(x,0,0)\le\frac{J_1^2+\lambda_SG_f^2}{2\mu_\tau}.
 \label{eq:Ebound}
\end{equation}
One can also require $\|D_U\bar S_c\|\le d_*$ and
$\|\nabla_U\Gamma_c\|\le d_*\|R_c\|$ on this neighborhood.
\end{lemma}
\begin{proof}
Choose a fixed $\tau_0\ge1$ above the curvature threshold in
\eqref{glob:tau-choice}. Since $D_\alpha^3\mathcal K$ is bounded by
$L_K$, differentiating the residual integral gives
\[
 \|D_UR_c\|\le(J_3+\tau L_K)s+L_f\le D_*\tau,
 \qquad D_*=(J_3/\tau_0+L_K)s+L_f/\tau_0.
\]
Set $d_*=D_*/\kappa$. Implicit differentiation bounds $D_U\bar S_c$
by $d_*$. The envelope identity and strong monotonicity imply
$\|\nabla_U\Gamma_c\|\le D_*\tau\|S-\bar S_c\|
\le d_*\|R_c\|$. Choose
\begin{equation}
 0<\lambda_S\le\min\{1,p_-^2/[4(p_+d_*+K_{\max})^2]\}.
 \label{eq:lambdaS}
\end{equation}
Omit the second restriction if its denominator is zero. This choice is
independent of $\tau$ and makes the population direction descend as
proved in Lemma~\ref{lem:population-coupled-descent}.

Integrating $D_U\bar S_c$ on the segment from $U$ to $\alpha_0^*$ gives
$\|\bar S_c(x,U)-S_c^*(x)\|\le d_*\|U-\alpha_0^*\|$.
The proof-ball inclusion holds along that segment by convexity.
The triangle inequality and \eqref{eq:gap-distance} yield the first
bound with $E_* =\max\{2(1+2d_*^2),4/\lambda_S\}$.
At $U=S=0$, the KL gradient vanishes, so $\|a\|\le J_1$ and
$\|R_c\|=\|f_\alpha(x,0)\|\le G_f$. Apply the residual-to-gap
inequalities in \eqref{eq:gap-gradient} to prove the initialization bound.
This argument holds for every $x$ and requires no warm start.
\end{proof}

\begin{lemma}[Tracked corrected-field error]
\label{lem:tracked-field}
Let $V_c^*(x)=V_c(x,\alpha_0^*(x),S_c^*(x))$. Then
\begin{equation}
 \|V_c(p)-\nabla\Phi(x)\|^2\le C(\mathcal T(p)+c^2),
 \qquad
 \|V_c(p)\|^2\le C(\|\nabla\Phi(x)\|^2+\mathcal T(p)+c^2).
 \label{eq:tracked-field-potential}
\end{equation}
\end{lemma}
\begin{proof}
Lemma~\ref{glob:secant} gives the Lipschitz bound for $V_c$ in the actor
pair and $\|V_c^* -\nabla\Phi\|\le C_{\rm sec}c$. Combining it with
\eqref{eq:Ebound} gives the first inequality. Adding $\nabla\Phi$
and using the squared triangle inequality gives the second.
\end{proof}

\subsection{Common-KL geometry and finite local updates}
\label{app:support-prox}
Both actors use the same stored empirical state measure
$\dbar_t=\widehat\rho_n$ from
Definition~\ref{klfin:construction:metric}. Its observable covariance
certificate provides the same Fisher constants as the population law
\begin{equation}
 \nu_t^{\rm pop}=\tfrac12d_{U_t}^\gamma+\tfrac12d_{U_t+cS_t}^\gamma.
 \label{eq:common-state-law}
\end{equation}
The latter explains the original constants but is not evaluated by the
algorithm. The metric data are drawn once, independently of the actor
initialization and all subsequent training blocks, and their cost is
included in \eqref{critic:cost}. A rejected covariance certificate returns
$x_0$. Conditional on acceptance, the following analytic reference maps
apply to the actual empirical metric. Their finite implementation and
its error propagation are proved below.

For a probability state law $\nu$ and actor points $\alpha,b$, define
\begin{equation}
 \mathcal K_{\mathbf m,\nu}(\alpha\mid b)
 =\E_{s\sim\nu}\KL(\pi_{\mathbf m,b}(\cdot\mid s)\|\pi_{\mathbf m,\alpha}(\cdot\mid s)),
 \label{eq:common-kl}
\end{equation}
and the exact local map
\begin{equation}
 \mathsf M_{\mathbf m,\nu,\eta}(b,v)
 =\argmin_{\alpha\in\overline B(b,r_{\rm prox})}
 \{\langle v,\alpha-b\rangle+\eta^{-1}\mathcal K_{\mathbf m,\nu}(\alpha\mid b)\}.
 \label{eq:prox-map}
\end{equation}
Choose
\begin{equation}
 0<r_{\rm prox}\le\min\{R/8,\underline\lambda_{F,L}/(2M_{\log,3,L})\},
 \label{eq:prox-radius}
\end{equation}
omitting a zero-denominator restriction. If $G_{\rm prox}(H)\ge1$ bounds both
sampled actor directions, impose the deterministic local-step restriction
\begin{equation}
 \eta_yG_{\rm prox}(H)\le m_{K,L}r_{\rm prox}/2,
 \qquad m_{K,L}=\underline\lambda_{F,L}/2.
 \label{eq:prox-step-condition}
\end{equation}
The sampled bounds in Appendix~\ref{app:support-sampling} make
$G_{\rm prox}(H)$ polynomial in $H$. This condition is satisfied by the
headline schedule for sufficiently small accuracy. The exact map is used
only as a mathematical reference. Algorithm~\ref{alg:main-trained} returns
the residual-controlled finite approximation in
Definition~\ref{klfin:construction:PGD}; its error is incorporated below.
For brevity, write
\begin{equation}
  k_{\mathbf m,\nu}(\alpha,b)
  :=\grad_\alpha\cK_{\mathbf m,\nu}(\alpha\mid b).
  \label{eq:k-definition}
\end{equation}

\begin{lemma}[Derivatives of the oriented common KL]
\label{lem:kl-derivatives}
For every state law $\nu$,
\begin{align}
  k_{\mathbf m,\nu}(\alpha,b)
  &=-\E_{\nu,\pi_{\mathbf m,b}}
   [\grad_\alpha\log\pi_{\mathbf m,\alpha}(a\mid s)],
  \label{eq:k-formula}\\
  k_{\mathbf m,\nu}(b,b)&=0,
  \label{eq:k-zero}\\
  D_1k_{\mathbf m,\nu}(b,b)
  &=\Fisher_{\mathbf m,\nu}(b),
  \label{eq:k-alpha-diagonal}\\
  D_2k_{\mathbf m,\nu}(b,b)
  &=-\Fisher_{\mathbf m,\nu}(b).
  \label{eq:k-beta-diagonal}
\end{align}
On the fixed ambient actor ball, all mixed derivatives of $k_{\mathbf m,\nu}$ of total order at most three are bounded uniformly in the widths and in the probability law $\nu$.
\end{lemma}

\begin{proof}
Differentiate the first argument of the oriented KL under the integral to obtain \eqref{eq:k-formula}.  The conditional score identity
$\E_{\pi_b}\grad_b\log\pi_b=0$ proves \eqref{eq:k-zero}.  Differentiating this zero-score identity in a direction $h$ gives
\[
  \E_{\pi_b}
  [D^2\log\pi_b[h]+\grad\log\pi_b
   \ip{\grad\log\pi_b}{h}]=0.
\]
Hence $-\E D^2\log\pi_b$ equals the Fisher matrix, proving \eqref{eq:k-alpha-diagonal}.  Differentiating the expectation in \eqref{eq:k-formula} with respect to the sampling parameter $b$ gives the negative Fisher matrix at the diagonal, proving \eqref{eq:k-beta-diagonal}.  Each further derivative is a finite sum of products of log-policy derivatives and score factors.  \Cref{lem:log-policy} bounds all terms through the required order.
\end{proof}

\begin{lemma}[Uniform local KL conditioning]
\label{lem:local-kl}
Let $\nu$ be a probability law such that $\underline\lambda_{F,L}I_r\preceq\Fisher_{\mathbf m,\nu}(b)\preceq\overline\lambda_{F,L}I_r$, let $b\in\Theta_R$, and suppose the segment from $b$ to $\alpha$ lies in the local geometry tube with
$\norm{\alpha-b}\le r_{\mathrm{prox}}$.  Then
\begin{equation}
  m_{K,L}\Id_r
  \preceq\grad_{\alpha\alpha}^2
  \cK_{\mathbf m,\nu}(\alpha\mid b)
  \preceq M_{K,L}\Id_r,
  \label{eq:KL-hessian-bounds}
\end{equation}
where
\begin{equation}
  m_{K,L}:=\frac12\underline\lambda_{F,L},
  \qquad
  M_{K,L}:=\overline\lambda_{F,L}
  +M_{\log,3,L}r_{\mathrm{prox}}.
  \label{eq:KL-constants}
\end{equation}
Consequently,
\begin{equation}
  \frac{m_{K,L}}{2}\norm{\alpha-b}^2
  \le\cK_{\mathbf m,\nu}(\alpha\mid b)
  \le\frac{M_{K,L}}{2}\norm{\alpha-b}^2.
  \label{eq:KL-quadratic}
\end{equation}
Moreover,
\begin{equation}
  \left|\cK_{\mathbf m,\nu}(b+\Delta\mid b)
  -\frac12\ip{\Delta}
  {\Fisher_{\mathbf m,\nu}(b)\Delta}\right|
  \le\frac{M_{\log,3,L}}{6}\norm{\Delta}^3.
  \label{eq:KL-cubic}
\end{equation}
\end{lemma}

\begin{proof}
By \cref{lem:kl-derivatives}, the first-argument Hessian at the diagonal is the Fisher matrix.  Since
\[
 \grad_{\alpha\alpha}^2\cK(\alpha\mid b)
 =-\E_{\nu,\pi_b}D_\alpha^2\log\pi_\alpha,
\]
its derivative with respect to $\alpha$ has norm at most
$M_{\log,3,L}$.  Therefore
\[
  \norm{\grad_{\alpha\alpha}^2\cK(\alpha\mid b)
  -\Fisher_{\mathbf m,\nu}(b)}_\op
  \le M_{\log,3,L}\norm{\alpha-b}.
\]
Use the stated Fisher bounds and \eqref{eq:prox-radius} to obtain \eqref{eq:KL-hessian-bounds}. The implemented empirical metric satisfies these bounds by Proposition~\ref{klfin:prop:Fisher}; its proof does not assume domination by a population state law. Since the KL and its first derivative vanish at the diagonal, integrating the Hessian twice along the segment gives \eqref{eq:KL-quadratic}. The integral third-order Taylor remainder gives \eqref{eq:KL-cubic}.
\end{proof}

\begin{lemma}[Exact local proximal map and quantitative remainder]
\label{lem:prox-expansion}
Let $\nu$ satisfy the Fisher premise of \cref{lem:local-kl}, let $b$ lie at distance at least $r_{\mathrm{prox}}$ from $\partial\Theta_R$, and let $\norm v\le G_{\mathrm{prox}}(H)$.  Under \eqref{eq:prox-step-condition}, the problem \eqref{eq:prox-map} has a unique interior minimizer
$M=\mathsf M_{\mathbf m,\nu,\eta}(b,v)$ satisfying
\begin{equation}
  k_{\mathbf m,\nu}(M,b)+\eta v=0,
  \qquad
  \norm{M-b}\le\frac{\eta}{m_{K,L}}\norm v
  \le\frac12r_{\mathrm{prox}}.
  \label{eq:prox-FOC-bound}
\end{equation}
Let
\begin{equation}
  P_{\mathbf m,\nu}(b)
  :=\Fisher_{\mathbf m,\nu}(b)^{-1}.
  \label{eq:P-definition}
\end{equation}
There is a $C^1$ remainder $Q_{\mathbf m,\nu}$ such that
\begin{equation}
  \mathsf M_{\mathbf m,\nu,\eta}(b,v)
  =b-\eta P_{\mathbf m,\nu}(b)v
  +\eta^2Q_{\mathbf m,\nu}(b,v,\eta),
  \label{eq:prox-expansion}
\end{equation}
and
\begin{align}
  \norm{Q_{\mathbf m,\nu}(b,v,\eta)}
  &\le C_{Q,L}\norm v^2,
  \label{eq:Q-bound}\\
  \norm{D_b Q_{\mathbf m,\nu}(b,v,\eta)}_\op
  &\le C_{Q,L}\norm v^2,
  \label{eq:Q-beta-bound}\\
  \norm{D_v Q_{\mathbf m,\nu}(b,v,\eta)}_\op
  &\le C_{Q,L}\norm v.
  \label{eq:Q-v-bound}
\end{align}
The constant is uniform in the widths, $\nu$, and sufficiently small $\eta$;
it is also independent of $H$ under \eqref{eq:prox-step-condition}.
\end{lemma}

\begin{proof}
Strong convexity of the local objective gives uniqueness.  For a constrained minimizer $M$, the variational inequality with comparison point $b$ gives
\[
  \ip{k(M,b)+\eta v}{b-M}\ge0.
\]
Since $k(b,b)=0$ and $k(\cdot,b)$ is $m_{K,L}$-strongly monotone,
\[
 m_{K,L}\norm{M-b}^2
 \le\ip{k(M,b)}{M-b}
 \le-\eta\ip v{M-b}
 \le\eta\norm v\norm{M-b}.
\]
This proves the displacement bound.  The tube and stepsize conditions make the solution strictly interior, so the first-order equation holds.

Write $M=b+\eta d$.  For $\eta\ge0$, define the continuously extended map
\begin{equation}
  \Psi(d,b,v,\eta)
  :=\int_0^1D_1k(b+t\eta d,b)[d]dt+v.
  \label{eq:Psi-prox}
\end{equation}
For $\eta>0$, the fundamental theorem of calculus gives the exact identity
\begin{equation}
  \Psi(d,b,v,\eta)
  =\eta^{-1}k(b+\eta d,b)+v,
  \label{eq:Psi-quotient-identity}
\end{equation}
while at $\eta=0$ it becomes
$\Fisher_{\mathbf m,\nu}(b)d+v=0$.  Differentiating \eqref{eq:Psi-quotient-identity} with respect to $d$ gives, for $\eta>0$,
\begin{equation}
  D_d\Psi(d,b,v,\eta)
  =D_1k(b+\eta d,b),
  \label{eq:Psi-d-Jacobian}
\end{equation}
and the same formula holds at $\eta=0$ by continuity with value $\Fisher_{\mathbf m,\nu}(b)$.  Hence \cref{lem:local-kl} gives
$D_d\Psi\succeq m_{K,L}\Id_r$ throughout the local tube.  Since the derivatives of $k$ through order three are uniformly bounded, the quantitative implicit-function theorem gives a uniformly $C^2$ solution
$d=d(b,v,\eta)$ with
$d(b,v,0)=-P_{\mathbf m,\nu}(b)v$ and
$\norm d\le m_{K,L}^{-1}\norm v$.

Differentiate \eqref{eq:Psi-prox} with respect to $\eta$.  Every term in $\partial_\eta\Psi$ contains two factors of $d$, so
$\norm{\partial_\eta\Psi}\le C_L\norm v^2$.  Uniform invertibility of $D_d\Psi$ gives
$\norm{\partial_\eta d}\le C_L\norm v^2$.  Differentiating this identity in $b$ and $v$ and using the bounded third derivatives of $k$ gives
\[
 \norm{D_b\partial_\eta d}\le C_L\norm v^2,
 \qquad
 \norm{D_v\partial_\eta d}\le C_L\norm v.
\]
Finally,
\[
 d(b,v,\eta)+P(b)v
 =\eta\int_0^1\partial_\eta d(b,v,t\eta)dt.
\]
Define $Q$ by the integral on the right.  Multiplying by $\eta$ in $M=b+\eta d$ gives \eqref{eq:prox-expansion}, and the three displayed derivative bounds follow from the preceding estimates.
Only the displacement $\eta\|v\|$ is restricted to the fixed local ball;
the displayed powers of $\|v\|$ account explicitly for the magnitude of
the direction. Thus increasing the polynomial sample envelope with $H$
does not enter the remainder constant.
\end{proof}

\begin{lemma}[$c$-uniform coupled expansion for the exact deep KL updates]
\label{lem:pair-prox}
Fix a common state law $\nu$ and let
\begin{align*}
  U^+&=\mathsf M_{\mathbf m,\nu,\eta}(U,\widehat A),\\
  U^++cS^+&=\mathsf M_{\mathbf m,\nu,\eta}
  (U+cS,\widehat A+c\widehat R).
\end{align*}
Set
\begin{equation}
  P^0=P_{\mathbf m,\nu}(U),
  \qquad P^c=P_{\mathbf m,\nu}(U+cS),
  \qquad K_c(U,S)=\frac{P^0-P^c}{c}.
  \label{eq:Kc-definition}
\end{equation}
Then
\begin{align}
  U^+-U&=-\eta P^0\widehat A+\rho^U,
  \label{eq:base-pair-expansion}\\
  S^+-S&=-\eta P^c\widehat R
  +\eta K_c(U,S)\widehat A+\rho^S,
  \label{eq:scaled-pair-expansion}\\
  \norm{K_c(U,S)}_\op
  &\le L_{P,L}\norm S,
  \label{eq:Kc-bound}
\end{align}
where
\begin{align}
  \norm{\rho^U}
  &\le C_{Q,L}\eta^2\norm{\widehat A}^2,
  \label{eq:rhoU}\\
  \norm{\rho^S}
  &\le C_{Q,L}\eta^2\Bigl[
  \norm S(\norm{\widehat A}+c\norm{\widehat R})^2
  +(\norm{\widehat A}+c\norm{\widehat R})
   \norm{\widehat R}\Bigr].
  \label{eq:rhoS}
\end{align}
Every constant is uniform as $c\downarrow0$ and over the hidden widths.
\end{lemma}

\begin{proof}
Apply \eqref{eq:prox-expansion} to both updates and subtract before dividing by $c$:
\begin{align*}
  S^+-S
  &=-\eta\frac{P^c(\widehat A+c\widehat R)-P^0\widehat A}{c}\\
  &\quad+\eta^2\frac{
  Q(U+cS,\widehat A+c\widehat R,\eta)
  -Q(U,\widehat A,\eta)}{c}.
\end{align*}
The inverse-Fisher Lipschitz estimate gives \eqref{eq:Kc-bound}.  For the nonlinear remainder, apply the fundamental theorem of calculus along
$\theta\mapsto(U+\theta cS,\widehat A+\theta c\widehat R)$ and use
\eqref{eq:Q-beta-bound}--\eqref{eq:Q-v-bound}.  This proves \eqref{eq:rhoS}; the base estimate is immediate from \eqref{eq:Q-bound}.
\end{proof}

\begin{lemma}[Deterministic prospective actor increments]
\label{lem:actor-step-pointwise}
On the safe actor set, for sufficiently small absolute steps,
\[
 \|U^+-U\|+\|S^+-S\|\le C\eta_y\operatorname{poly}(H).
\]
The constants are uniform in $c$, width and critic scale.
\end{lemma}
\begin{proof}
Lemma~\ref{lem:pair-prox} bounds the virtual exact increments by linear
sampled directions plus quadratic local-map remainders. The directions
are polynomial in $H$ by Lemma~\ref{lem:paired-evaluation}, and
$\|S\|\le s$, $c\le c_0$, with uniformly bounded inverse metrics.
Lemma~\ref{klfin:lem:paired-error} adds at most $c\eta_y^2$ to the base
increment and $2\eta_y^2$ to the normalized increment. The displayed bound
therefore holds for the finite implementation as well when $c,\eta_y\le1$.
Selecting accuracy small
enough gives $\|\Delta U\|<r_{\rm prox}/2$ and $\|\Delta S\|<s/4$.
For all interpolation points used in the paired remainder proof, the
base point is $U+tcS$ and its direction is
$\widehat a+tc\widehat R$, $0\le t\le1$. They obey the same buffered
actor inclusion and direction bound, so the local map exists along the
entire interpolation before containment is invoked.
\end{proof}

\paragraph{The population direction.}
Let $p_-=\overline\lambda_{F,L}^{-1}$,
$p_+=\underline\lambda_{F,L}^{-1}$ and $K_{\max}=L_{P,L}s$.
The constant $d_*$ and the weight $\lambda_S$ are fixed in
Lemma~\ref{lem:tau-scaling}.
\begin{lemma}[Population coupled direction is descending]
\label{lem:population-coupled-descent}
Let $A=a(p)=g_\alpha(x,U)$ and $R=R_c(p)$. Let
\begin{align*}
 d_U&=-P^0A,\\
 d_S&=-P^cR+K_c(U,S)A,
\end{align*}
where $P^0,P^c$ are inverse common-state Fisher matrices.  Then
\begin{equation}
  \ip{\grad_U\mathcal T}{d_U}
  +\ip{\grad_S\mathcal T}{d_S}
  \le-\rho_{0,L}\mathcal T,
  \qquad
  \rho_{0,L}:=p_-\min\{\mu_\tau,\mu_\tau\}.
  \label{eq:population-T-descent}
\end{equation}
\end{lemma}
\begin{proof}
Since $p_-\Id\preceq P^0,P^c\preceq p_+\Id$ and
$\norm{K_c}\le K_{\max}$, the left-hand side is at most
\begin{align*}
  -p_-\norm A^2-\lambda_Sp_-\norm R^2
  +\lambda_S(p_+d_*+K_{\max})\norm A\norm R.
\end{align*}
Let $D=p_+d_*+K_{\max}$.  Young's inequality gives
\[
 \lambda_SD\norm A\norm R
 \le\frac{p_-}{4}\norm A^2
 +\frac{\lambda_S^2D^2}{p_-}\norm R^2.
\]
The choice \eqref{eq:lambdaS} makes the last coefficient at most
$\lambda_Sp_-/4$.  Hence the directional derivative is at most
$-(p_-/2)\norm A^2-(\lambda_Sp_-/2)\norm R^2$.  The PL inequalities for the base gap and $\Gamma$ give
$\norm A^2\ge2\mu_\tau\Delta_0$ and
$\norm R^2\ge2\mu_\tau\Gamma$, proving \eqref{eq:population-T-descent}.
\end{proof}

Throughout this subsection, suppress width subscripts and write
$Z_\alpha=\nabla_\alpha F_{\mathbf m,\alpha}$,
$G_Z=C_{1,L}$, $\lambda_0=\lambda_{\rm eff,L}$ and
$L_3=M_{\log,3,L}$. The actor bounds in
Appendix~\ref{app:support-actor} and the radius~\eqref{eq:R} give
\begin{align}
 \|Z_\alpha(s,a)\|&\le G_Z, &
 \|Z_\alpha(s,a)-Z_b(s,a)\|&\le C_{2,L}\|\alpha-b\|,
 \label{klfin:eq:feature-bounds}\\
 \mathcal G_0:=\E_\rho\Cov_{\piRef}(Z_0)&\succeq\lambda_0I_r,
 & C_{2,L}R&\le\sqrt{\lambda_0}/2,
 \label{klfin:eq:radius}\\
 \pi_\alpha(a\mid s)&\ge e^{-2G_ZR}\piRef(a\mid s),&
 \|D_\alpha^3\log\pi_\alpha(a\mid s)\|_\op&\le L_3.
 \label{klfin:eq:density}
\end{align}
All these inequalities hold on $\overline B_R$. In particular, the
population constants already used in the exact reference-map analysis are
\begin{equation}
 \underline\lambda_{F,L}=(1-\gamma)e^{-2G_ZR}\lambda_0/4,
 \qquad \overline\lambda_{F,L}=4G_Z^2.
 \label{klfin:eq:Fisher-constants}
\end{equation}
The finite metric below preserves these constants, so it does not change
the fixed regularization, the actor buffers or the domain required by C2.
It is used only for update geometry: the population objectives,
selected response and all trajectory laws for evaluation and TD retain
their original definitions. Actor probabilities and derivatives are
computed on finitely many states with full finite-action sums, in the
same real-arithmetic model as the other neural calculations.

\subsubsection{An empirical metric with the existing Fisher constants}
\label{klfin:sec:metric}

\begin{definition}[One initial metric sample]
\label{klfin:construction:metric}
Fix the realized actor initialization and retained chart before drawing
metric data. For a chosen \(\delta_{\mathrm{met}}\in(0,1)\), independently
draw \(s_1,\ldots,s_n\) from \(\rho\), with
\begin{equation}
 n\ge
 \left\lceil
 \frac{8G_Z^2}{\gamma^2\lambda_0}
 \log\frac{r}{\delta_{\mathrm{met}}}
 \right\rceil.
 \label{klfin:eq:metric-size}
\end{equation}
Define
\begin{align}
 \widehat\rho_n&=\frac1n\sum_{i=1}^n\delta_{s_i}, \label{klfin:eq:empirical-law}\\
 B(s)&=\Cov_{a\sim\piRef(\cdot\mid s)}(Z_0(s,a)),
 \qquad
 \widehat{\mathcal G}_n=\frac1n\sum_{i=1}^n B(s_i).
 \label{klfin:eq:empirical-cov}
\end{align}
Compute each covariance by summing over all actions. Accept the metric
if
\begin{equation}
 \widehat{\mathcal G}_n\succeq(1-\gamma/2)\lambda_0I_r .
 \label{klfin:eq:accept}
\end{equation}
If the certificate fails, return the prescribed fallback \(x_0\).
Upon acceptance, use \(\bar d_t=\widehat\rho_n\) for both endpoint updates at
every iteration. All subsequent training blocks are generated using
the specified sampling protocol, independently of the metric data
conditional on the old parameters and history.
\end{definition}

This is the actual initialization covariance under \(\widehat\rho_n\).
The specified optional pilot based on two sampled actions estimates
a population covariance and is a different matrix. The full action sums
in \eqref{klfin:eq:empirical-cov} directly certify the empirical KL
measure used here.

\begin{proposition}[Certificate acceptance probability]
\label{klfin:prop:acceptance}
Conditional on any fixed initialization and chart satisfying A1--A2,
Definition~\ref{klfin:construction:metric} rejects with probability at
most \(\delta_{\mathrm{met}}\).
\end{proposition}
\begin{proof}
For every unit vector \(v\),
\[
 0\le v^\top B(s)v
   =\Var_{\piRef}(v^\top Z_0(s,\cdot))
   \le\E_{\piRef}(v^\top Z_0(s,\cdot))^2
   \le G_Z^2.
\]
Thus \(0\preceq B(s_i)\preceq G_Z^2I_r\), and
\(\E B(s_i)=\mathcal G_0\succeq\lambda_0I_r\).
Let \(\mu_{\min}=\lambda_{\min}(\sum_i\E B(s_i))\ge n\lambda_0\).
The matrix Chernoff lower-tail inequality
\cite[Corollary 5.2 and Remark 5.3]{tropp2012} gives
\[
 \Prb\!\left(
 \lambda_{\min}\Bigl(\sum_i B(s_i)\Bigr)
 \le(1-t)\mu_{\min}\right)
 \le r\exp\!\left(-\frac{t^2\mu_{\min}}{2G_Z^2}\right),
 \qquad 0<t<1.
\]
The rejection event in \eqref{klfin:eq:accept} is contained in this
event with \(t=\gamma/2\). Hence
\begin{equation}
 \Prb(\text{metric rejection})
 \le r\exp\!\left(-\frac{n\gamma^2\lambda_0}{8G_Z^2}\right)
 \le\delta_{\mathrm{met}}.
 \label{klfin:eq:reject-prob}
\end{equation}
\end{proof}

\begin{proposition}[Uniform empirical Fisher certificate]
\label{klfin:prop:Fisher}
On the observable acceptance event, simultaneously for every
\(\alpha\in\overline B_R\),
\begin{equation}
 \underline\lambda_{F,L}I_r
 \preceq\widehat{\Fisher}_n(\alpha):=\Fisher_{\mathbf m,\widehat\rho_n}(\alpha)
 \preceq\overline\lambda_{F,L}I_r .
 \label{klfin:eq:empirical-Fisher}
\end{equation}
These are the constants in \eqref{klfin:eq:Fisher-constants}.
The inverse \(P_n(\alpha)=\widehat{\Fisher}_n(\alpha)^{-1}\) is uniformly bounded
and Lipschitz, with bounds independent of \(n\).
\end{proposition}
\begin{proof}
Fix a unit \(v\), and use the joint law
\(s\sim\widehat\rho_n,\ a\sim\piRef(\cdot\mid s)\).
Conditional centering is an orthogonal projection in \(L^2\).
Applying the triangle inequality to centered
\(v^\top Z_\alpha\) and \(v^\top Z_0\), followed by
\eqref{klfin:eq:feature-bounds}, gives
\[
 \left(\E_{\widehat\rho_n}\Var_{\piRef}(v^\top Z_\alpha)\right)^{1/2}
 \ge
 \left(\E_{\widehat\rho_n}\Var_{\piRef}(v^\top Z_0)\right)^{1/2}
       -C_{2,L}\norm{\alpha}.
\]
On acceptance, the first term on the right is at least
\(\sqrt{(1-\gamma/2)\lambda_0}\). The right-hand side is positive
on \(\overline B_R\) by \eqref{klfin:eq:radius}.

For any real-valued action function \(Y\), the density lower bound
in \eqref{klfin:eq:density} implies
\[
 \Var_{\pi_\alpha}(Y)
 =\inf_{a_0\in\R}\E_{\pi_\alpha}(Y-a_0)^2
 \ge e^{-2G_ZR}\inf_{a_0\in\R}\E_{\piRef}(Y-a_0)^2.
\]
Therefore
\begin{equation}
 v^\top\widehat{\Fisher}_n(\alpha)v
 \ge e^{-2G_ZR}
 \left(\sqrt{(1-\gamma/2)\lambda_0}-C_{2,L}R\right)^2.
 \label{klfin:eq:intermediate-Fisher}
\end{equation}
Set \(a=\sqrt{1-\gamma}\). The root-mean-square inequality gives
\[
 \sqrt{1-\gamma/2}
 =\sqrt{\frac{1+a^2}{2}}\ge\frac{1+a}{2}.
\]
Combining this with \eqref{klfin:eq:radius} proves the lower bound
in \eqref{klfin:eq:empirical-Fisher}.
The centered feature has norm at most \(2G_Z\), proving the upper bound.

The specified log-policy derivative bounds give, for every fixed
probability law \(\nu\),
\[
 \norm{\Fisher_{\mathbf m,\nu}(\alpha)-\Fisher_{\mathbf m,\nu}(\beta)}_{\op}
 \le L_F\norm{\alpha-\beta},
\]
where \(L_F\) is independent of \(\nu\). Apply this to \(\widehat\rho_n\), and use
the identity \(A^{-1}-B^{-1}=A^{-1}(B-A)B^{-1}\), to obtain
\begin{equation}
 \norm{P_n(\alpha)}_{\op}\le\underline\lambda_{F,L}^{-1},
 \qquad
 \norm{P_n(\alpha)-P_n(\beta)}_{\op}
 \le\frac{L_F}{\underline\lambda_{F,L}^2}\norm{\alpha-\beta}.
 \label{klfin:eq:inverse-bounds}
\end{equation}
\end{proof}

\begin{remark}[Why one sample works throughout training]
\label{klfin:rem:metric}
The accepted covariance certificate implies
\eqref{klfin:eq:empirical-Fisher} for all points of the actor ball,
including later data-dependent iterates. No union bound over iterations
or covering argument over actor parameters is needed.
An atomic empirical measure need not dominate \((1-\gamma)\rho\);
the proof above establishes its Fisher bound directly.
Nor is an approximation of the current population occupancy required:
the existing update analysis uses the measure as a predictable,
well-conditioned common metric.

The threshold \((1-\gamma/2)\lambda_0\) is chosen to reproduce the
existing constants. A half-gap threshold also gives positive curvature
with suitable constants, but is not used for that purpose here.
The \(\gamma^{-2}\) factor in \eqref{klfin:eq:metric-size} is an explicit
sufficient cost of this choice; no optimal dependence on \(\gamma\)
is claimed.
\end{remark}

\subsubsection{An explicit finite local solver}
\label{klfin:sec:solver}

Fix accepted metric data, a buffered old actor \(b\), and a frozen
sampled direction \(h\), with \(\norm h\le G_{\mathrm{prox}}(H)\).
Set
\begin{align}
 \widehat{\mathcal K}_n(\alpha\mid b)
 &=\frac1n\sum_{i=1}^n
   \KL\bigl(\pi_b(\cdot\mid s_i)\,\|\,\pi_\alpha(\cdot\mid s_i)\bigr),
 \label{klfin:eq:KL}\\
 \ell_{b,h}(\alpha)
 &=\widehat{\mathcal K}_n(\alpha\mid b)+\eta_y\ip h{\alpha-b}.
 \label{klfin:eq:scaled-problem}
\end{align}
This scaled objective has the same minimizer as the specified
linear cost plus \(\eta_y^{-1}\) times KL.
Every derivative in the numerical solver is with respect to \(\alpha\),
holding \(b,h\), and the metric sample fixed.

The gradient is the finite sum
\begin{equation}
 \nabla\ell_{b,h}(\alpha)
 =\frac1n\sum_{i=1}^n\sum_{a\in\mathcal A}
       \bigl(\pi_\alpha(a\mid s_i)-\pi_b(a\mid s_i)\bigr)
       Z_\alpha(s_i,a)
       +\eta_y h.
 \label{klfin:eq:finite-gradient}
\end{equation}
Indeed, the derivative of KL is
\(-\sum_a\pi_b(a\mid s)\nabla_\alpha\log\pi_\alpha(a\mid s)\);
substitute the centered-feature identity from the softmax actor definition.
The calculation uses actor first derivatives and finite action sums.
It requires neither transition probabilities nor a population-state
expectation.

Retain the local radius and step restrictions
\begin{equation}
 0<r_{\mathrm{prox}}\le
 \min\left\{\frac R8,\frac{\underline\lambda_{F,L}}{2L_3}\right\},
 \qquad
 \eta_y G_{\mathrm{prox}}(H)\le\frac{m_{K,L}r_{\mathrm{prox}}}{2},
 \label{klfin:eq:prox-restrictions}
\end{equation}
omitting the second radius restriction if \(L_3=0\), where
\begin{equation}
 m_{K,L}=\frac{\underline\lambda_{F,L}}{2},
 \qquad
 M_{K,L}=\overline\lambda_{F,L}+L_3r_{\mathrm{prox}},
 \qquad
 \kappa_K=\frac{M_{K,L}}{m_{K,L}}.
 \label{klfin:eq:KL-constants}
\end{equation}
Assume the local ball
\(\mathcal B_b=\overline B(b,r_{\mathrm{prox}})\) lies in \(\overline B_R\).
The specified observable safe-set test provides this buffer for
both endpoint bases and their interpolation.

By Lemmas~\ref{lem:local-kl} and~\ref{lem:prox-expansion},
$\ell_{b,h}$ is $m_{K,L}$-strongly convex and $M_{K,L}$-smooth on
$\mathcal B_b$, and its reference minimizer
$M(b,h)=\mathsf M_{\mathbf m,\widehat\rho_n,\eta_y}(b,h)$ is interior,
with $\nabla\ell_{b,h}(M)=0$ and
$\|M-b\|\le\eta_y\|h\|/m_{K,L}\le r_{\rm prox}/2$.
These are analysis points; the algorithm computes the finite routine below.

\begin{definition}[Residual-controlled projected gradient descent]
\label{klfin:construction:PGD}
Take \(0<c\le\min\{c_0,1\}\) and set
\begin{equation}
 d_{\mathrm{sol}}=c\eta_y^2,
 \qquad
 t_{\mathrm{res}}=m_{K,L}c\eta_y^2.
 \label{klfin:eq:solver-tolerances}
\end{equation}
Initialize \(\alpha^{(0)}=b\). At numerical iteration \(j\), compute
\(g^{(j)}=\nabla\ell_{b,h}(\alpha^{(j)})\) from
\eqref{klfin:eq:finite-gradient}. If
\begin{equation}
 \norm{g^{(j)}}\le t_{\mathrm{res}},
 \label{klfin:eq:stop}
\end{equation}
return \(\widetilde{\mathsf M}(b,h)=\alpha^{(j)}\). Otherwise set
\begin{equation}
 \alpha^{(j+1)}
 =\Pi_{\mathcal B_b}
      \left(\alpha^{(j)}-\frac{g^{(j)}}{2M_{K,L}}\right)
 \label{klfin:eq:PGD}
\end{equation}
and continue. The Euclidean projection is explicit:
\[
 \Pi_{\mathcal B_b}(y)
 =b+(y-b)\min\{1,r_{\mathrm{prox}}/\norm{y-b}\},
\]
with value \(b\) when \(y=b\).
The same frozen metric sample is reused at every numerical iteration.
\end{definition}

\begin{proposition}[Finite termination and certified endpoint accuracy]
\label{klfin:prop:solver}
Definition~\ref{klfin:construction:PGD} returns an endpoint satisfying
\begin{equation}
 \norm{\widetilde{\mathsf M}(b,h)-M(b,h)}\le c\eta_y^2.
 \label{klfin:eq:endpoint-accuracy}
\end{equation}
Define \(\log_+(u)=\max\{0,\log u\}\) for \(u>0\). The number of
projected updates is at most
\begin{equation}
 J_{\max}
 =
 \left\lceil
 2\kappa_K
 \log_+\left(\frac{M_{K,L}G_{\mathrm{prox}}(H)}{m_{K,L}^2c\eta_y}\right)
 \right\rceil.
 \label{klfin:eq:Jmax}
\end{equation}
At most \(J_{\max}+1\) finite-sum gradients are evaluated.
\end{proposition}
\begin{proof}
Let \(M=M(b,h)\). For \(\alpha\in\mathcal B_b\), the integral Hessian
along the segment from \(M\) to \(\alpha\) has eigenvalues in
\([m_{K,L},M_{K,L}]\). Since \(\nabla\ell_{b,h}(M)=0\), the unprojected
gradient map with step \(1/(2M_{K,L})\) contracts distance to \(M\) by
at most
\[
 q_K=1-\frac{m_{K,L}}{2M_{K,L}}.
\]
Projection onto the convex ball is nonexpansive and fixes \(M\).
Consequently,
\begin{equation}
 \norm{\alpha^{(j)}-M}
 \le q_K^j\norm{b-M}
 \le e^{-j/(2\kappa_K)}\frac{\eta_y G_{\mathrm{prox}}(H)}{m_{K,L}}.
 \label{klfin:eq:distance-contraction}
\end{equation}
Local smoothness gives
\[
 \norm{\nabla\ell_{b,h}(\alpha^{(j)})}
 \le M_{K,L}\norm{\alpha^{(j)}-M}
 \le\frac{M_{K,L}\eta_y G_{\mathrm{prox}}(H)}{m_{K,L}}e^{-j/(2\kappa_K)}.
\]
This is at most \(m_{K,L}c\eta_y^2\) by \(j=J_{\max}\).
Finally, strong monotonicity and the zero gradient at \(M\) imply
\[
 m_{K,L}\norm{\alpha-M}^2
 \le\ip{\nabla\ell_{b,h}(\alpha)}{\alpha-M}
 \le\norm{\nabla\ell_{b,h}(\alpha)}\norm{\alpha-M}.
\]
At the stopping point, this yields \eqref{klfin:eq:endpoint-accuracy}.
All iterates stay in the local ball by projection, so the proof does
not presuppose actor containment beyond the current buffered step.
\end{proof}

\begin{remark}[Normalization and arithmetic]
The stopping test is for the scaled objective
\eqref{klfin:eq:scaled-problem}. If implemented using
\(\ip h{\alpha-b}+\eta_y^{-1}\widehat{\mathcal K}_n(\alpha\mid b)\), the equivalent
gradient-residual threshold is \(m_{K,L}c\eta_y\).
The parameter tolerance \(c\eta_y^2\) is not an objective-value tolerance.
With \(c=\Theta(\sqrt\epsilon)\) and \(\eta_y=\Theta(\epsilon)\), it is
\(O(\epsilon^{5/2})\); geometric convergence gives a logarithmic
iteration dependence on this accuracy.
Neither the stopping rule nor the count above assumes an accurate
optimization oracle. Gradient evaluations are the explicit finite
neural computations in \eqref{klfin:eq:finite-gradient}.
\end{remark}

\subsubsection{Paired numerical errors and the sampling history}
\label{klfin:sec:pair}

At an old iterate \(p=(x,U,S)\), use the specified frozen lower
estimates \((\widehat a,\widehat R)\), and form
\(\widehat C=\widehat a+c\widehat R\).
The two bases and directions are
\[
 b_0=U,\quad h_0=\widehat a,\qquad
 b_c=U+cS,\quad h_c=\widehat a+c\widehat R.
\]
Let \(M_0=M(b_0,h_0)\), \(M_c=M(b_c,h_c)\). These are mathematical
reference points used only in the proof.
The implemented updates are
\begin{equation}
 U^+=\widetilde{\mathsf M}(b_0,h_0),\qquad
 \alpha_c^+=\widetilde{\mathsf M}(b_c,h_c),\qquad
 S^+=\frac{\alpha_c^+-U^+}{c}.
 \label{klfin:eq:implemented-pair}
\end{equation}

\begin{lemma}[Samplewise paired error]
\label{klfin:lem:paired-error}
At the same old iterate and with the same sampled directions, define
the virtual exact update
\(\bar U^+=M_0,\ \bar S^+=(M_c-M_0)/c\).
Then, samplewise,
\begin{align}
 e_U&:=U^+-\bar U^+,
 &\norm{e_U}&\le c\eta_y^2, \label{klfin:eq:eU}\\
 e_S&:=S^+-\bar S^+,
 &\norm{e_S}&\le2\eta_y^2. \label{klfin:eq:eS}
\end{align}
In particular, for \(c\le1\),
\begin{equation}
 \norm{(e_U,e_S)}\le\sqrt5\,\eta_y^2.
 \label{klfin:eq:joint-error}
\end{equation}
No differentiability of the stopping rule or coupling of numerical
stopping times is required.
\end{lemma}
\begin{proof}
Set \(e_0=\widetilde{\mathsf M}(b_0,h_0)-M_0\) and
\(e_c=\widetilde{\mathsf M}(b_c,h_c)-M_c\).
Proposition~\ref{klfin:prop:solver} gives
\(\norm{e_0},\norm{e_c}\le c\eta_y^2\). The identities
\(e_U=e_0\) and \(e_S=(e_c-e_0)/c\) prove the first two bounds.
Their squared sum is at most \((c^2+4)\eta_y^4\).
\end{proof}

For completeness, the exact reference-map expansion from Lemma~\ref{lem:pair-prox}
therefore yields the implemented expansion
\begin{align}
 U^+-U&=-\eta_y P^0\widehat a+\rho_U+e_U, \label{klfin:eq:U-expansion}\\
 S^+-S&=-\eta_y P^c\widehat R
          +\eta_y K_c\widehat a+\rho_S+e_S, \label{klfin:eq:S-expansion}\\
 P^0&=P_n(U),\qquad
 P^c=P_n(U+cS),\qquad K_c=\frac{P^0-P^c}{c}.
 \notag
\end{align}
The original remainders \(\rho_U,\rho_S\) have conditional first moments
\(O(\eta_y^2)\), second moments \(O(\eta_y^4)\), and fourth moments
\(O(\eta_y^8)\). Adding \((e_U,e_S)\) preserves these orders:
its corresponding bounds follow immediately from
\eqref{klfin:eq:joint-error}. The normalized inverse-metric difference
is still bounded by
\(\norm{K_c}_{\op}\le
L_F\underline\lambda_{F,L}^{-2}\norm S\).

\begin{lemma}[Filtration and prospective-step validity]
\label{klfin:lem:history}
Include the entire initial metric sample in the pre-draw history
\(\mathcal F_t\). Use old parameters for all three training blocks,
and run both local solvers solely with the frozen lower block and
metric data. Then the simultaneous \((x,U,S)\) increment remains
conditionally independent of the current TD block. The conditional
movement estimates of Appendix~\ref{app:support-tracking} retain their orders.
For sufficiently small accuracy in the headline schedule, the
implemented prospective step lies in the existing larger analytic
actor neighborhood, before containment has been established.
\end{lemma}
\begin{proof}
Each numerical iterate and its stopping index are measurable functions
of the old history and lower block. The outer update uses its independent
old-parameter block. The TD block is independent of these two blocks
conditional on \(\mathcal F_t\), as stipulated in the existing protocol.
Hence the conditional centering argument for moving-target TD applies.
The mean and centered numerical perturbations have orders
\(O(\eta_y^2)\), with second and fourth moment bounds supplied by
\eqref{klfin:eq:joint-error}; these preserve the movement estimates.

At a safe old point, both local balls and the corresponding balls along
the endpoint interpolation lie in \(\overline B_R\).
The projected inner iterations stay in their respective balls.
The virtual exact coupled increments have deterministic bound
\(C\eta_y\operatorname{poly}(H)\) by the specified paired expansion.
The implemented increments add at most \(\sqrt5\eta_y^2\).
With logarithmic \(H\) and \(\eta_y=\Theta(\epsilon)\), both tend to zero.
In particular, the existing prospective restrictions
\(\norm{\Delta U}<r_{\mathrm{prox}}/2\) and \(\norm{\Delta S}<s/4\) hold for
sufficiently small \(\epsilon\). The virtual and implemented points,
and the segment between them, lie in the same convex buffered analytic
neighborhood. This argument uses only the current safe point and
therefore introduces no circular appeal to future containment.
\end{proof}

\subsubsection{Transfer to tracking, containment, and stationarity}
\label{klfin:sec:transfer}

Use the specified actor potential
\(\mathcal T(p)=\Delta_0(x,U)+\lambda_S\Gamma_c(x,U,S)\).
To distinguish the moving critic target from the actor feature \(Z_\alpha\),
write that target as \(\mathcal Z(p)\), and set
\[
 \mathcal C(p,z)=\norm{z-\mathcal Z(p)}_W^2,\qquad
 X(p)=\norm{\nabla\Phi(x)}^2,\qquad
 \omega=c^2+\AQ+b_{\mathrm{ev}}^2+b_{\mathrm{TD}}^2.
\]
The established bounds on the prospective neighborhood include
\begin{equation}
 \norm{\nabla_p\mathcal T}\le L_{\mathcal T},\qquad
 \norm{\mathcal Z(p)-\mathcal Z(\bar p)}_W
       \le L_{\mathcal Z}\norm{p-\bar p},\qquad
 \mathcal Z(p),z\in\CritSet.
 \label{klfin:eq:transfer-inputs}
\end{equation}
These constants are global in the upper parameter and uniform in the
critic scale and \(c\). Put \(D_Q=\operatorname{diam}_W(\CritSet)<\infty\).

\begin{proposition}[One-step transfer without new accuracy assumptions]
\label{klfin:prop:transfer}
At a safe old iterate, compare the implemented step with the virtual
exact-proximal step at the same old iterate, using identical lower,
outer, and TD blocks. The outer and critic updates themselves agree,
and the next actor/upper tuples \(p^+,\bar p^+\) satisfy
\[
 x^+=\bar x^+,\qquad z^+=\bar z^+,\qquad
 \norm{p^+-\bar p^+}\le\sqrt5\eta_y^2.
\]
Consequently,
\begin{align}
 |\mathcal T(p^+)-\mathcal T(\bar p^+)|
       &\le\sqrt5 L_{\mathcal T}\eta_y^2, \label{klfin:eq:T-transfer}\\
 |\mathcal C(p^+,z^+)-\mathcal C(\bar p^+,z^+)|
       &\le2\sqrt5 D_Q L_{\mathcal Z}\eta_y^2, \label{klfin:eq:C-transfer}\\
 \Phi(x^+)-\Phi(\bar x^+)&=0. \label{klfin:eq:Phi-transfer}
\end{align}
The existing conditional actor/critic tracking recursions and outer
descent inequality thus retain their form, with enlarged second-order
remainder constants.
\end{proposition}
\begin{proof}
Both the outer and TD routines use the old parameters, so using the
same sampled blocks gives identical \(x^+\) and \(z^+\).
Lemma~\ref{klfin:lem:paired-error} bounds the remaining difference.
Integrating the actor-potential gradient along the segment supplied by
Lemma~\ref{klfin:lem:history} gives \eqref{klfin:eq:T-transfer}.
For the critic error, both
\(z^+-\mathcal Z(p^+)\) and \(z^+-\mathcal Z(\bar p^+)\)
have \(W\)-norm at most \(D_Q\). The difference of their squared
norms is at most
\[
 2D_Q\norm{\mathcal Z(p^+)-\mathcal Z(\bar p^+)}_W,
\]
which proves \eqref{klfin:eq:C-transfer}.
The last identity is immediate.

The specified exact-map inequalities are one-step inequalities
uniform over safe old iterates. They can therefore be applied to the
virtual step at each implemented iterate; a separate exact-solver
trajectory is unnecessary. Take conditional expectations in the
displayed samplewise bounds. Each additional actor or critic term is
\(O(\eta_y^2)\), already present in the corresponding recurrence.
The outer descent inequality is unchanged.
\end{proof}

In particular, the transferred recursions have the existing orders
\begin{align}
 \E_t\Delta\mathcal C_t
 &\le-k\beta\mathcal C_t
  +C\frac{\eta_y^2}{\beta}\mathcal T_t
  +C\frac{\eta_x^2}{\beta}X_t
  +C\beta\omega+C(\beta^2+\eta_y^2+\eta_x^2),
 \label{klfin:eq:C-rec}\\
 \E_t\Delta\mathcal T_t
 &\le-u\eta_y\mathcal T_t+v\eta_x X_t+w\eta_y\mathcal C_t
       +C\eta_y\omega+C(\eta_y^2+\eta_x^2),
 \label{klfin:eq:T-rec}\\
 \E_t\Delta\Phi_t
 &\le-a\eta_xX_t+b\eta_x\mathcal T_t+d\eta_x\mathcal C_t
       +C\eta_x\omega+C\eta_x^2 .
 \label{klfin:eq:Phi-rec}
\end{align}
Here the positive drift constants and sufficiently small fixed
step-ratio choices are those furnished by the specified proof;
the numerical perturbation enters the indicated remainder terms.
In particular, it does not create a new \(c^{-1}\) error floor.

\begin{corollary}[Containment and augmented descent]
\label{klfin:cor:potentials}
Let \(\eta_y=r_Q\beta\), \(\eta_x=q\eta_y\), with the sufficiently small
fixed positive ratios selected in the analysis.
The contracting potential
\(\mathcal D=\mathcal T+\lambda_Dr_Q\mathcal C\)
and stationarity potential
\[
 \Psi=\Phi-\underline\Phi
       +a_\Psi q\mathcal T+b_\Psi qr_Q\mathcal C
\]
retain their required drift and innovation bounds.
In particular, the new contribution to the one-step drift of \(\Psi\)
is \(O(\eta_x\eta_y)\), and the added contribution to the drift of
\(\mathcal D\) is \(O(\eta_y^2)\).
\end{corollary}
\begin{proof}
For \(\Psi\), multiply \eqref{klfin:eq:T-transfer} by \(a_\Psi q\)
and \eqref{klfin:eq:C-transfer} by \(b_\Psi qr_Q\).
Since \(r_Q\le1\) and \(q\eta_y=\eta_x\), the extra term is bounded by
\(Cq\eta_y^2=C\eta_x\eta_y\). It is absorbed in the existing
\(C\eta_x(\omega+\beta+\eta_y+\eta_x)\) remainder.
For \(\mathcal D\), the additional term is \(C\eta_y^2\).
After division by \(\eta_y\), it is \(C\eta_y\le C\beta\), again
within the existing contracting floor.

The difference between the implemented next \(\mathcal D\) and the
virtual exact next \(\mathcal D\) is samplewise \(O(\eta_y^2)\).
Centering at most doubles this bound. The original innovation bound
\(|N_{t+1}|\le\eta_y B_D(H)\) therefore remains valid after enlarging
its polynomial envelope. Together with
Lemma~\ref{klfin:lem:history}, the same stopped martingale maximum
argument and first-exit contradiction apply.
\end{proof}

\subsection{Critic calculus and auxiliary TD estimates}
\label{app:support-critic}
The critic in \eqref{eq:main-trained-critic} trains a fixed-dimensional
chart $\theta$. Increasing $m_Q$ changes its nonlinear scale, not its
number of independent trainable coordinates. For integer $m_Q$, tied
tanh units give this realization exactly.
\begin{lemma}[Nonlinear critic derivatives and linearization]
\label{lem:critic-calculus}
On $\|\theta\|\le R_Q$, the following bounds hold uniformly over inputs:
\begin{align}
 |Q_{m_Q,\theta}|&\le F_Q R_Q,&
 \norm{\nabla Q_{m_Q,\theta}}&\le F_Q,\notag\\
 \norm{Q_{m_Q,\theta}-\varphi_Q^\top\theta}_\infty
 &\le F_Q^3R_Q^3/(3m_Q),&
 \norm{\nabla Q_{m_Q,\theta}-\varphi_Q}_\infty
 &\le F_Q^3R_Q^2/m_Q,\notag\\
 \norm{\nabla^2Q_{m_Q,\theta}}_\infty
 &\le 2F_Q^3R_Q/m_Q,&
 \norm{\nabla^3Q_{m_Q,\theta}}_\infty&\le2F_Q^3/m_Q.
 \label{critic:neural-bounds}
\end{align}
Fourth derivatives are also uniformly bounded for $m_Q\ge1$.
\end{lemma}
\begin{proof}
Put $t=\varphi_Q^\top\theta/\sqrt{m_Q}$. Direct differentiation gives
$\nabla Q=\operatorname{sech}^2(t)\varphi_Q$ and
$\nabla^2Q=-2\operatorname{sech}^2(t)\tanh(t)
\varphi_Q^{\otimes2}/\sqrt{m_Q}$. Use
$|\tanh t|\le|t|$, $|\tanh t-t|\le|t|^3/3$, and
$1-\operatorname{sech}^2t\le t^2$.
The third derivative is
$(4\operatorname{sech}^2t-6\operatorname{sech}^4t)
\varphi_Q^{\otimes3}/m_Q$, whose scalar coefficient has magnitude at
most two. The fourth derivative is a bounded polynomial in
$\tanh t$ and $\operatorname{sech}t$, multiplied by
$\varphi_Q^{\otimes4}/m_Q^{3/2}$. These formulas prove all claims.
\end{proof}
\begin{lemma}[Anisotropic derivative bounds on unrestricted $x$]
\label{critic:anisotropic}
Regard $(U,\alpha,\theta)$ as independent variables and denote this group by $y$. On the indicated neighborhoods, every derivative block
\begin{equation}
 D_x^aD_y^b \mathsf H(x,U,\alpha,\theta),
 \qquad 0\le a\le2,\quad a+b\le3,
 \label{critic:anisotropic-bound}
\end{equation}
is uniformly bounded. Derivatives of $\widehat{\mathsf H}$ at a fixed transition, holding its sampling law fixed, have uniform envelopes of the corresponding orders. Derivatives of $\nu_U$ are treated separately by likelihood differentiation and geometric moments; their prefix likelihood factors have polynomial length envelopes. Pure $D_x^3\mathsf H$ is not needed or claimed uniformly bounded.
\end{lemma}
\begin{proof}
The base occupancy is independent of $x$ and $\alpha$. Its $U$ derivatives through order three are controlled by differentiating the prefix likelihood under $\pi_U$. Bounded derivatives of the log-policy give degree-at-most-three polynomials in prefix length. The sums $\sum_{k\ge0}(1-\gamma)\gamma^k(k+1)^j$ are finite. This proves differentiation under the discounted sum in total variation and supplies uniform bounds for those derivatives.

For the TD integrand, an $x$ derivative acts only on $q=-r_x+\tau\log(\pi_\alpha/\piRef)$. Thus $D_x^aq=-D_x^ar_x$ for $a=1,2$, and $D_x^aD_\alpha^bq=0$ when $a,b\ge1$. Actor derivatives act on $K_\alpha$, the entropy criterion, or the base likelihood. Three derivatives in $\theta$ of the TD integrand use derivatives of $Q$ through order four, uniformly bounded by~\eqref{critic:neural-bounds} and its fourth-order analogue. Product differentiation and~\eqref{eq:main-global-envelopes} give every block in~\eqref{critic:anisotropic-bound}.

For example, at independent $(U,\alpha,\theta)$,
$\mathsf H_{xx}=\E_{\nu_U}[r_{xx}\nabla Q]$,
$\mathsf H_{xx\alpha}=0$, and
$\mathsf H_{xx\theta}=\E_{\nu_U}[r_{xx}\nabla^2Q]$;
$\mathsf H_{xxU}$ differentiates only the base law in the first expression. These are the potentially delicate blocks needed after taking the normalized endpoint difference. No pure third $x$ derivative enters. The same domination proves continuity of these derivatives, not merely a norm bound.
\end{proof}

\subsection{Sampling routines and paired oracle estimates}
\label{app:support-sampling}
All blocks use the pre-draw history and old parameters specified at the
start of Appendix~\ref{app:main-results}, including the stored accepted
metric sample. The reset states used for that metric are independent of
these training blocks and are reused only for the finite KL computations.
Write
$M_1=M_{\log,1,L}$. Define the sample TD residual at a common transition by
\begin{align}
 \widehat{\mathsf H}(x,U,\alpha,\theta)
 &=\Bigl[Q_{m_Q,\theta}(s,a)-q_{x,\alpha}(s,a)\notag\\
 &\qquad-\gamma\sum_{a'}\pi_\alpha(a'\mid s')Q_{m_Q,\theta}(s',a')\Bigr]
 \nabla_\theta Q_{m_Q,\theta}(s,a),\notag\\
 F^{\rm samp}(p,z;s,a,s')
 &=\begin{pmatrix}
 \widehat{\mathsf H}(x,U,U,u)\\[2pt]
 [\widehat{\mathsf H}(x,U,U+cS,u+cv)-\widehat{\mathsf H}(x,U,U,u)]/c
 \end{pmatrix}.
 \label{eq:algorithm-TD}
\end{align}
Here $q_{x,\alpha}$ is the regularized lower criterion in \eqref{eq:regularized-value}.
Both endpoints use the same base-law transition; the next-action expectations are exact finite sums.
With $\widehat F_{t,H}$ the truncated sample average, update
\begin{equation}
 z_{t+1}=\Pi_{\mathcal K_Q}^{W}(z_t-\beta\widehat F_{t,H}),\qquad
 \mathcal K_Q=\overline B_{R_u}\times\overline B_{R_v},\qquad
 \|(u,v)\|_W^2=M\|u\|^2+\|v\|^2.
 \label{eq:algorithm-critic-update}
\end{equation}
The critic construction in Step~\ref{app:main-step-td} supplies fixed $R_u,R_v,M$ and
an admissible critic scale. Projection is a specified critic update; critic accuracy is proved.
Initialize any deterministic $z_0\in\mathcal K_Q$, without an accuracy or burn-in requirement.
Both algorithms use the single actor test \eqref{eq:safe}, initially and after every completed iteration.

\paragraph{Explicit evaluation samplers.}
Write $\alpha_0=U$, $\alpha_c=U+cS$, $\theta_0=u$, $\theta_c=u+cv$, and
$Q_j=Q_{m_Q,\theta_j}$. The critic represents the unnormalized discounted action value,
whereas $g_\tau$ is normalized by $1-\gamma$. Consequently the following score-times-value
actor estimator has no additional $(1-\gamma)^{-1}$ factor.
For a common-start mixture prefix $\xi=(s_0,a_0,\ldots,s_n,a_n)$ with $s_0\sim\rho$, set
\begin{equation}
 \ell_j(\xi)=\prod_{k=0}^n\pi_{\alpha_j}(a_k\mid s_k),\qquad
 w_j(\xi)=\frac{2\ell_j(\xi)}{\ell_0(\xi)+\ell_c(\xi)},\qquad j\in\{0,c\}.
 \label{eq:trained-prefix-weights}
\end{equation}
The terminal action is included. Shared reset and transition factors cancel in this ratio.
At the same terminal state-action pair define
\begin{align}
 G_j(\xi)&=\nabla_\alpha\log\pi_{\alpha_j}(a_n\mid s_n)Q_j(s_n,a_n)
             +\partial_\alpha q_{x,\alpha_j}(s_n,a_n),\notag\\
 a(\xi)&=w_0G_0,\qquad
 d(\xi)=\frac{w_cG_c-w_0G_0}c,\qquad
 e(\xi)=-\frac{w_c-w_0}c\nabla_xr_x(s_n,a_n).
 \label{eq:trained-evaluation-samples}
\end{align}
The direct criterion derivative $\partial_\alpha q$ differentiates the policy-ratio criterion,
not the trained critic coordinates. For a fresh preference tuple and label at the old perturbed actor,
\begin{equation}
 P_\alpha(\boldsymbol\xi,Y)=L(x;\boldsymbol\xi,Y)
 \sum_{i=1}^I\sum_{k=0}^{K-1}\nabla_\alpha\log\pi_{\alpha_c}(a_k^i\mid s_k^i),
 \qquad P_x(\boldsymbol\xi,Y)=\nabla_xL(x;\boldsymbol\xi,Y).
 \label{eq:main-preference-oracles}
\end{equation}
These are unbiased for $f_\alpha,f_x$ and satisfy
$\|P_\alpha\|\le B_LIKM_1$, $\|P_x\|\le G_L$. Neither is divided by $c$.

In \cref{alg:deep}, a capped geometric draw means drawing $N$ independently with
$\Pr(N=n)=(1-\gamma)\gamma^n$ and using a zero sample if $N\ge H$.
For $N<H$, an equal-mixture prefix chooses $J\sim\operatorname{Unif}\{0,c\}$,
resets from $\rho$, and follows $\pi_{\alpha_J}$ for $N$ transitions and a terminal action.
This convention does not condition on $N<H$; such conditioning would change the field normalization.
The three blocks below define $\mathsf{LowerPair}$, $\mathsf{OuterPair}$, and $\mathsf{TDPair}$
in \cref{alg:main-trained}.

\begin{algorithm}[!htbp]
\caption{One iteration: explicit sampling and persistent paired TD}
\label{alg:deep}
\small
\begin{algorithmic}[1]
\REQUIRE Frozen $p=(x,U,S)$, $z=(u,v)$, stored certified $\dbar=\widehat\rho_n$;
 $H$, $B_H,B_V,B_C,B_{f,a},B_{f,x}$ and the main algorithm's steps and radii.
\ENSURE New $(x^+,U^+,S^+,z^+)$ or failure; all oracle arguments are old.
\STATE Set $\alpha_0=U$, $\alpha_c=U+cS$, $\theta_0=u$, $\theta_c=u+cv$.
\FOR{$b=1,\ldots,B_H$}
 \STATE Independently draw geometric $N_b$, with $\Pr(N_b=n)=(1-\gamma)\gamma^n$.
 \IF{$N_b\ge H$}
  \STATE Set $(a_b,d_b)=(0,0)$.
 \ELSE
  \STATE Draw an equal-mixture reset prefix $\xi_b$ with $N_b$ transitions and terminal action.
  \STATE Compute $(a_b,d_b)=(a(\xi_b),d(\xi_b))$ using both old critics.
 \ENDIF
\ENDFOR
\STATE Independently draw $B_{f,a}$ preference tuples and labels at old $\alpha_c$;
 let $\widehat P_\alpha$ average $P_\alpha$.
\STATE $\widehat a=B_H^{-1}\sum_ba_b$, $\widehat R=B_H^{-1}\sum_bd_b+\widehat P_\alpha$;
 $\widehat C=\widehat a+c\widehat R$.
\STATE Use Definition~\ref{klfin:construction:PGD} to compute
 $U^+=\widetilde{\mathsf M}(U,\widehat a)$;
 $\alpha_c^+=\widetilde{\mathsf M}(\alpha_c,\widehat C)$;
 $S^+=(\alpha_c^+-U^+)/c$.
\FOR{$b=1,\ldots,B_V$}
 \STATE Independently draw a fresh capped geometric equal-mixture prefix at the old actor pair.
 \STATE Set $e_b=0$ for a capped draw and $e_b=e(\xi_b)$ otherwise.
\ENDFOR
\STATE Independently draw $B_{f,x}$ preference tuples and labels at old $\alpha_c$;
 let $\widehat P_x$ average $P_x$.
\STATE $\widehat V=B_V^{-1}\sum_be_b+\widehat P_x$;
 $x^+=x-\eta_x\widehat V$.
\FOR{$b=1,\ldots,B_C$}
 \STATE Independently of both evaluation blocks, draw geometric $N_b$.
 \IF{$N_b\ge H$}
  \STATE Set $F_b=0$.
 \ELSE
  \STATE Reset from $\rho$ and follow $\pi_U$ for $N_b$ transitions to $s$;
 draw $a\sim\pi_U(\cdot\mid s)$, $s'\sim P(\cdot\mid s,a)$.
  \STATE Evaluate $F_b=F^{\rm samp}(p,z;s,a,s')$ from \eqref{eq:algorithm-TD} using old $p,z$.
 \ENDIF
\ENDFOR
\STATE $\widehat F=B_C^{-1}\sum_bF_b$;
 $z^+=\Pi_{\mathcal K_Q}^{W}(z-\beta\widehat F)$.
\IF{$\mathsf{Safe}(U^+,S^+)$ fails}
 \STATE \textbf{return} failure; \cref{alg:main-trained} returns the fixed $x_0$.
\ENDIF
\STATE \textbf{return} $(x^+,U^+,S^+,z^+)$.
\end{algorithmic}
\end{algorithm}

\FloatBarrier
\paragraph{Mean identities and truncation.}
If a scalar test function $v$ is integrable, geometric prefix sampling gives
$\E[v(s_N,a_N)]=\E_{\nu_\alpha}v$. Differentiating the discounted
Bellman equation for $Q^{\rm true}$ and multiplying by the normalized
reset resolvent yields
\begin{equation}
 \begin{aligned}
 g_\alpha(x,\alpha)&=\E_{\nu_\alpha}
 [\nabla_\alpha\log\pi_\alpha\,Q^{\rm true}_{x,\alpha}
       +\partial_\alpha q_{x,\alpha}],\\
 g_x(x,\alpha)&=\E_{\nu_\alpha}[-\nabla_xr_x].
 \end{aligned}
 \label{critic:PG}
\end{equation}
Indeed, differentiating $Q=q+\gamma K_\alpha Q$ first accounts for
the change in the current action law, the criterion and all future action
laws. The resolvent collects these score terms over discounted occupancy;
the lower normalization cancels the reset-resolvent factor. Fixed dynamics
make the second identity immediate. No derivative is taken through learned
critic coordinates. With zero output for $N\ge H$, the TD tail is
\begin{equation}
 \|\E_t\widehat F_{t,H}-F(p_t,z_t)\|_W\le C_{\rm TD}\gamma^H=b_{\rm TD}.
 \label{critic:td-bias}
\end{equation}
This is truncation, not conditioning on $N<H$.

\paragraph{Common-prefix balance weights.}
For fixed prefix length $n$ (including $n+1$ action factors), let
$p^0,p^c$ have reset law $\rho$, common dynamics and endpoint policies;
let $q=(p^0+p^c)/2$ and $w_j=dp^j/dq$. Selecting the endpoint once
per trajectory implements precisely this mixture.
\begin{lemma}[Balance-weight identities]
\label{lem:balance}
\begin{equation}
  0\le w_0,w_c\le2,
  \qquad w_0+w_c=2\quad q\text{-a.s.},
  \qquad
  \E_q[w_jZ]=\E_{p^j}[Z],\quad j\in\{0,c\}.
  \label{eq:balance-identities}
\end{equation}
\end{lemma}
\begin{proof}
Since $q\ge p^j/2$, $w_j\le2$.  The identity $w_0+w_c=2$ follows from $p^0+p^c=2q$, and the expectation identity is the Radon--Nikodym change-of-measure formula.
\end{proof}

\begin{lemma}[Divided balance-weight difference is uniform in $c$]
\label{lem:weight-difference}
Let $r(\xi)=\log(dp^c/dp^0)(\xi)$.  Then
\begin{equation}
  w_c-w_0=2\tanh(r/2)
  \label{eq:tanh-weight}
\end{equation}
and
\begin{equation}
  \left|\frac{w_c(\xi)-w_0(\xi)}{c}\right|
  \le (n+1)M_{\log,1,L}\norm S.
  \label{eq:divided-weight-bound}
\end{equation}
\end{lemma}
\begin{proof}
Writing $z=dp^c/dp^0=e^r$ gives
$w_c=2z/(1+z)$ and $w_0=2/(1+z)$, hence \eqref{eq:tanh-weight}.  The common initial law and common environment dynamics cancel from the path likelihood ratio, leaving
\[
  r(\xi)=\sum_{h=0}^{n}
  \left[\log\pi_{U+cS}(a_h\mid s_h)
       -\log\pi_U(a_h\mid s_h)\right].
\]
The integral mean-value formula and \cref{lem:log-policy} give
$|r|\le c(n+1)M_{\log,1,L}\norm S$.  Since
$|\tanh z|\le|z|$, \eqref{eq:divided-weight-bound} follows.
\end{proof}

\begin{lemma}[Paired evaluation errors, tails and moments]
\label{lem:paired-evaluation}
For $z^*=Z(p)$ and $e_j=Q_{m_Q,\theta_j}-Q^{\rm true}_{x,\alpha_j}$,
\begin{equation}
 \|e_0\|_\infty^2+\|(e_c-e_0)/c\|_\infty^2
 \le C\bigl(\|z-z^*\|_W^2+\mathcal A_Q\bigr).
 \label{eq:paired-evaluation-error}
\end{equation}
The base and normalized lower evaluation biases are therefore bounded
in squared norm by $C(\|z-z^*\|_W^2+\mathcal A_Q+b_{\rm ev}^2)$.
The outer bias is bounded by $b_{\rm ev}$ without critic error.
For some fixed integer $\ell$,
$b_{\rm ev}\le C(H+1)^\ell\gamma^H$.
Every fixed-order conditional moment of the sampled directions is
bounded independently of $H,c,m_Q,x$; their samplewise envelopes are
polynomial in $H$.
\end{lemma}
\begin{proof}
Use the integral
\[
 \frac{Q_{m_Q,u+cv}-Q_{m_Q,u}}c
 =\int_0^1\nabla Q_{m_Q,u+tcv}^{\top}v\,dt.
\]
The first two critic derivative bounds make this map Lipschitz in
$(u,v)$, uniformly in $c$ on $\mathcal K_Q$. The base value is also
Lipschitz. Subtract their values at $z^*$ and use \eqref{critic:AQ}
to obtain \eqref{eq:paired-evaluation-error}.

For lower score-times-value errors, set
$T_j=\nabla\log\pi_{\alpha_j}\,e_j$. Both the scores and their
normalized endpoint differences are bounded. Decompose before dividing:
\begin{align*}
 \frac{w_cT_c-w_0T_0}c
 &=w_c\frac{T_c-T_0}c+\frac{w_c-w_0}cT_0,\\
 \frac{T_c-T_0}c
 &=\nabla\log\pi_{\alpha_c}\frac{e_c-e_0}c
  +\frac{\nabla\log\pi_{\alpha_c}-\nabla\log\pi_{\alpha_0}}c e_0.
\end{align*}
Lemmas~\ref{lem:balance}--\ref{lem:weight-difference} bound this by
$C(n+1)(\|z-z^*\|_W+\sqrt{\mathcal A_Q})$.
The direct criterion derivatives introduce no critic error. Taking
geometric expectations proves the untruncated bias bound. The outer
sample in \eqref{eq:trained-evaluation-samples} uses only $r_x$ derivatives.

On the projection set, base learned values and normalized differences
are bounded even when inaccurate. Thus each untruncated direction is
bounded by $C(n+1)^\ell$ for a fixed degree. Every moment
$\sum_{n\ge0}(1-\gamma)\gamma^n(n+1)^j$ is finite.
The removed tail is at most $C(H+1)^\ell\gamma^H$, by writing
$n=H+k$ and summing the geometric polynomial. Capping supplies the
samplewise polynomial envelope; it does not worsen any fixed-order
moment. Preference terms are bounded by Proposition~\ref{pref:oracle-means}.
For TD, the integral identity \eqref{critic:pair-integral} bounds the
sample uniformly, and its omitted tail proves \eqref{critic:td-bias}.
\end{proof}

\paragraph{Direct preference samples.}
\begin{proposition}[Unbiased bounded tuple gradients]
\label{pref:oracle-means}
For a tuple and label sampled at any frozen actor $\alpha$, use $P_x,P_\alpha$ from \eqref{eq:main-preference-oracles}, with $\alpha_c$ replaced by $\alpha$.
Conditionally on every pre-sampling history that fixes $(x,\alpha)$,
\begin{equation}
 \E[P_x\mid x,\alpha]=f_x(x,\alpha),\qquad
 \E[P_\alpha\mid x,\alpha]=f_\alpha(x,\alpha).
 \label{pref:unbiased}
\end{equation}
For every realized sample,
\begin{equation}
 \|P_x\|\le G_L,
 \qquad \|P_\alpha\|\le B_LIKM_1.
 \label{pref:oracle-bounds}
\end{equation}
Averages $\widehat F_x,\widehat F_\alpha$ of respectively
$B_{f,x},B_{f,a}$ independent tuple--label samples have conditional
variance at most $G_L^2/B_{f,x}$ and
$(B_LIKM_1)^2/B_{f,a}$.
\end{proposition}
\begin{proof}
The joint tuple--label law factors into the $I$ trajectory laws and
$\mathcal D$. Fixed initial law and dynamics and the parameter-independent
label kernel imply that its actor log derivative is the score sum
$\sum_{i,k}\nabla_\alpha\log\pi_\alpha(a_k^i\mid s_k^i)$; its $x$ derivative is zero.
Differentiating under the finite expectation is justified by the displayed
uniform envelopes. Since the sample loss has no explicit actor dependence,
this gives \eqref{pref:unbiased}. The triangle inequality gives
a score-sum bound of $IKM_1$, proving \eqref{pref:oracle-bounds}.
Conditional independence and centering divide the second-moment bounds by
the batch size. These statements concern each sampled label, rather than
only a label-averaged population loss.
\end{proof}

\paragraph{Batch sizes and timing.}
Fix positive preference batch sizes $B_{f,a},B_{f,x}$ independently of
accuracy. One may take any positive TD/evaluation batch sizes
$B_C,B_H,B_V$ bounded by a fixed polynomial in $H$, including fixed
sizes: the geometric moment bounds already give the uniform moments
through order eight used in the proof. Averaging independent samples
preserves those bounds and reduces centered second moments.
Conditional on $\mathcal F_t$, each evaluation stage has its own
independent tuple/label block, and TD is independent of both evaluation
stages. The exact identity $\widehat C=\widehat a+c\widehat R$ is
samplewise. After conditioning further on $\mathcal G_t$, the outer
block still has its old-parameter conditional mean. No lower labels are
reused for the outer estimator, and the critic update is first used at
iteration $t+1$. The full interaction ledger is \eqref{critic:cost}.

\FloatBarrier
\subsection{Conditional tracking estimates}
\label{app:support-tracking}
The results here use the derived trained-critic bias from
Lemma~\ref{critic:oracle}, the residual-controlled finite solves, and the pre-draw history
$\mathcal F_t$ defined in Appendix~\ref{app:main-results}. They do not
introduce an independent oracle-accuracy assumption.

\begin{lemma}[Conditional means and moments of the simultaneous increment]
\label{lem:conditional-movement}\label{lem:prox-remainder-moments}
Write $\Delta p_t=p_{t+1}-p_t=d_t+w_{t+1}$, where
$d_t=\E_t\Delta p_t$ and $\E_tw_{t+1}=0$. For the prospective step,
\begin{align}
 \|d_t\|^2&\le C\{(\eta_y^2+\eta_x^2)
  (\mathcal T_t+\mathcal C_t+\omega)+\eta_x^2X_t
  +(\eta_y^2+\eta_x^2)^2\},\notag\\
 \E_t\|w_{t+1}\|^2&\le C(\eta_y^2+\eta_x^2),\qquad
 \E_t\|\Delta p_t\|^4\le C(\eta_y+\eta_x)^4.
 \label{critic:movement}
\end{align}
The combined reference-map and finite-solver remainders have conditional
first, second and fourth moments $O(\eta_y^2)$, $O(\eta_y^4)$ and
$O(\eta_y^8)$, respectively.
The entire $\Delta p_t$ is conditionally independent of the current TD
block. All constants are uniform over subsequently smaller step ratios.
\end{lemma}
\begin{proof}
For each sampled direction decompose, for example,
$\widehat a=a+e_a+\zeta_a$, where
$e_a=\E_t\widehat a-a$ and $\E_t\zeta_a=0$; use the same decomposition
for $\widehat R,\widehat V$. The inverse metrics $P^0,P^c$ and correction
$K_c$ and the full metric sample are fixed under $\mathcal F_t$.
Lemma~\ref{lem:pair-prox} and the samplewise bounds in
Lemma~\ref{klfin:lem:paired-error} give
\begin{align*}
 \Delta U&=-\eta_yP^0\widehat a+\rho^U+e_U,\\
 \Delta S&=-\eta_yP^c\widehat R+\eta_yK_c\widehat a+\rho^S+e_S,\\
 \Delta x&=-\eta_x\widehat V.
\end{align*}
The reference-map remainders are $\eta_y^2$ times polynomials of degree at most two
in the directions, with bounded $S,c$. Direction moments through order
eight from Lemma~\ref{critic:oracle} bound their means, second moments
and fourth moments. Moreover, $\|(e_U,e_S)\|\le\sqrt5\eta_y^2$
samplewise, so their means and centered parts preserve these moment
orders. Centered first-order terms have zero conditional
mean. Thus the population field bounds \eqref{critic:actor-field-bounds}
and the squared bias bound \eqref{critic:oracle-bias} give the first
line of \eqref{critic:movement}. The bounded second moments of all
directions and the remainder estimates give its second line; the same
argument using fourth moments of the increment gives the last bound.
No independence between the base and scaled lower estimates is needed.

The two finite actor updates, including their stopping indices, are
measurable functions of the lower block and stored metric sample, while the upper
update is a function of its separate outer block. Both use old parameters.
The TD block is independent of this pair of blocks under $\mathcal F_t$.
Hence revealing TD first preserves the centering of $w_{t+1}$, exactly
as used in the moving-target critic proof. Conditional fourth moments
are finite without any bound on upper iterates, because the reward and
loss envelopes are global.
\end{proof}

\begin{lemma}[Conditional Taylor transfer for actor tracking]
\label{lem:tracking-taylor}\label{lem:moving-drift}
Let $d_U^{\rm pop}=-P^0a$ and
$d_S^{\rm pop}=-P^cR_c+K_ca$. On a prospective step,
\begin{align}
 \E_t\Delta\mathcal T_t
 &\le\eta_y\langle\nabla_{U,S}\mathcal T_t,
                 (d_U^{\rm pop},d_S^{\rm pop})\rangle\notag\\
 &\quad-\eta_x\langle\nabla_x\mathcal T_t,V_c(p_t)\rangle
 +C(\eta_y+\eta_x)\sqrt{\mathcal T_t}
                 \sqrt{\mathcal C_t+\omega}
 +C(\eta_y^2+\eta_x^2).
 \label{eq:trained-tracking-Taylor}
\end{align}
\end{lemma}
\begin{proof}
Apply the quadratic Taylor formula in all variables $(x,U,S)$ along the
segment from $p_t$ to $p_t+\Delta p_t$. Lemma~\ref{glob:tracking-regularity}
bounds the full Hessian there, including mixed actor--upper blocks.
Lemma~\ref{lem:conditional-movement} therefore bounds the expected
quadratic remainder by $C(\eta_y^2+\eta_x^2)$. Substitute the conditional
mean of each update from its paired expansion. Centered first-order
noise vanishes because the old gradient and metric are
$\mathcal F_t$-measurable. The conditional means of the combined
reference-map and numerical remainders cost $C\eta_y^2$, since the old
potential gradient is bounded.
The remaining bias products are bounded by
$C(\eta_y+\eta_x)\|\nabla\mathcal T_t\|
\sqrt{\mathcal C_t+\mathcal A_Q+b_{\rm ev}^2}$.
The gradient self-bound gives the displayed inequality. This calculation
also covers the lower-first implementation order: all fields remain
functions of the old parameters, and fresh outer data retain their
conditional mean after revealing $\mathcal G_t$. No conditional expectation
is taken with respect to a future non-exit event.
\end{proof}

\subsection{Analytic and probabilistic tools}
\label{app:support-tools}
All segmentwise statements below are used on the larger prospective
actor neighborhood. Short Young and quadratic Taylor inequalities are
applied directly in the main proofs.
\begin{lemma}[Covariance transfer under domination]
\label{lem:cov-transfer}
If $P\succeq qQ$ for probability measures $P,Q$ and $q\in(0,1]$, then for every square-integrable vector $X$,
\[
  \Cov_P(X)\succeq q\Cov_Q(X).
\]
\end{lemma}
\begin{proof}
For a vector $v$, write $Y=v^\top X$.  The variational identity
$\Var_P(Y)=\inf_{a\in\R}\E_P[(Y-a)^2]$ gives
\[
  \Var_P(Y)\ge q\inf_a\E_Q[(Y-a)^2]
  =q\Var_Q(Y).
\]
Apply the scalar conclusion to every $v$.
\end{proof}

\begin{lemma}[Conditional centering is an $L_2$ contraction]
\label{lem:centering}
On a joint state-action probability space, let
$\Pi X=X-\E[X\mid s]$.  Then
\[
  \norm{\Pi X}_{L_2}\le\norm{X}_{L_2},
  \qquad
  \norm{\Pi X-\Pi Y}_{L_2}\le\norm{X-Y}_{L_2}.
\]
\end{lemma}
\begin{proof}
Conditional expectation is the orthogonal projection onto the closed subspace of $s$-measurable random variables.  Hence $\Pi$ is the complementary orthogonal projection and has operator norm one.
\end{proof}

\begin{lemma}[Inverse perturbation]
\label{lem:inverse-perturbation}
If $A$ and $B$ are invertible matrices with
$\norm{A^{-1}}_\op,\norm{B^{-1}}_\op\le\mu^{-1}$, then
\[
  \norm{A^{-1}-B^{-1}}_\op
  \le\mu^{-2}\norm{A-B}_\op.
\]
\end{lemma}
\begin{proof}
Use $A^{-1}-B^{-1}=A^{-1}(B-A)B^{-1}$.
\end{proof}

\begin{lemma}[Gradient bounds from strong convexity and smoothness]
\label{lem:strong-smooth-gradient}
Let $H$ be differentiable, $\mu$-strongly convex, and have $L_H$-Lipschitz gradient on a convex set.  Suppose its minimizer $z^\star$ is interior, so $\grad H(z^\star)=0$.  Then
\begin{align}
  \frac{\mu}{2}\norm{z-z^\star}^2
  &\le H(z)-H(z^\star),
  \label{eq:generic-qg}\\
  2\mu\bigl(H(z)-H(z^\star)\bigr)
  &\le\norm{\grad H(z)}^2,
  \label{eq:generic-pl}\\
  \norm{\grad H(z)}^2
  &\le\frac{2L_H^2}{\mu}
  \bigl(H(z)-H(z^\star)\bigr).
  \label{eq:generic-upper-gradient}
\end{align}
\end{lemma}
\begin{proof}
Strong convexity at $z^\star$ gives \eqref{eq:generic-qg}.  Strong convexity at $z$ gives
\[
 H(z^\star)\ge H(z)+\ip{\grad H(z)}{z^\star-z}
 +\frac\mu2\norm{z-z^\star}^2.
\]
Maximizing the right-hand upper bound on $H(z)-H(z^\star)$ over the displacement yields
$H(z)-H(z^\star)\le\norm{\grad H(z)}^2/(2\mu)$, proving \eqref{eq:generic-pl}.  Finally,
$\norm{\grad H(z)}\le L_H\norm{z-z^\star}$ and \eqref{eq:generic-qg} give \eqref{eq:generic-upper-gradient}.
\end{proof}

\begin{lemma}[Bounded innovations despite critic projection]
\label{critic:D-innovation}
There is a polynomial envelope $B_D(H)$ such that, at pre-exit steps including the prospective first exit,
\begin{equation}
 |N_{t+1}|\le\eta_y B_D(H),\qquad
 N_{t+1}=\mathcal D_{t+1}-\E_t\mathcal D_{t+1}.
 \label{critic:D-innovation-bound}
\end{equation}
No differentiability or inactivity of the critic projection is needed.
\end{lemma}
\begin{proof}
Since $z_t\in\CritSet$, nonexpansiveness gives
$\norm{z_{t+1}-z_t}_W\le\beta\norm{\widehat F_{t,H}}_W\le C\beta$.
Uniform target derivatives and~\eqref{critic:actor-expansion} give
$\norm{Z(p_{t+1})-Z(p_t)}_W\le C\eta_y\operatorname{poly}(H)$.
Both old and new error vectors have bounded norm because the critic iterate and prospective target lie in $\CritSet$. Hence
\[
 |\mathcal C_{t+1}-\mathcal C_t|\le C[\beta+\eta_y\operatorname{poly}(H)].
\]
Multiplication by $\lambda r_Q$ gives $C\eta_y\operatorname{poly}(H)$, using $r_Q\beta=\eta_y$ and $r_Q\le1$. Uniform first derivatives of the actor potential similarly give $|\mathcal T_{t+1}-\mathcal T_t|\le C\eta_y\operatorname{poly}(H)$. Centering doubles the resulting deterministic bound. Relative to the
virtual exact step, Proposition~\ref{klfin:prop:transfer} changes the
next $\mathcal D$ by at most $C\eta_y^2$ samplewise, and centering changes
its innovation by at most $2C\eta_y^2$. This is absorbed by enlarging
$B_D(H)$, without a $c^{-1}$ factor. The argument is valid for all
sufficiently small absolute steps on the reserved prospective buffers.
\end{proof}

\subsection{Assumption verification and the PL counterexample}
\label{app:support-verification}
A1's actor derivative requirement is verified by
Theorem~\ref{thm:deep-derivatives}; the fixed bounded input, normalized
initialization and fixed chart are part of that construction. A2 is an
identifiability requirement with the optional independent pilot in
Proposition~\ref{red:prop:pilot-certification}. Its local persistence
does not imply global nondegeneracy for unrestricted neural parameters.
For G, Proposition~\ref{glob:BT} and the bounded tanh reward example in
Appendix~\ref{app:support-preference} give explicit global envelopes.
Neither requires a correctly specified fitted preference model.

For C1, a feature Gram gap under the base reset law $\rho_U$ is sufficient
by $\nu_U\succeq(1-\gamma)\rho_U$. This concerns fixed critic features,
not the actor tangent features. A sufficient structural certificate for
C2 is that the feature span contains $r_x$ and the log-policy-ratio
criterion separately and is invariant under every admitted $K_\alpha$.
The Neumann series for $(I-\gamma K_\alpha)^{-1}$ then stays in this
finite-dimensional closed span. Finite state--action indicator features
provide one such realization when their occupancy Gram matrix has a
uniform gap. For continuous states the same span invariance and Gram gap
are substantive restrictions. The theorem requires C2 at the selected
fixed $\tau$; increasing $\tau$ or $m_Q$ alone does not establish it.
Even under exact C2, the finite nonlinear critic retains the derived
$O(m_Q^{-2})$ squared paired approximation floor in \eqref{critic:AQ}.

\paragraph{What generic policy smoothness does not imply.}
\begin{proposition}[Smooth actors and exact critics do not imply uniform PL]\label{prop:PL-counterexample}\label{plan:prop-PL}
Consider a one-state self-loop MDP with actions $0,1$, rewards $r(1)=1$ and $r(0)=0$, and the actor
\begin{equation}
\pi_\theta(1)=\tfrac12+\tfrac14\sin\theta,
\qquad \pi_\theta(0)=1-\pi_\theta(1),
\qquad \theta\in[-\pi,\pi].
\label{plan:PLexample}
\end{equation}
The actor probabilities are uniformly positive and all required score derivatives are bounded. A tabular critic can represent the action-value function exactly. Nevertheless the normalized discounted return fails every uniform maximization PL inequality with a positive constant.
\end{proposition}
\begin{proof}
The one state is visited at every step, so the normalized return is
\[
J(\theta)=(1-\gamma)\sum_{t\ge0}\gamma^t\pi_\theta(1)
=\tfrac12+\tfrac14\sin\theta.
\]
Its maximum is $J^*=3/4$ at $\theta=\pi/2$, whereas
\[
J'(-\pi/2)=\tfrac14\cos(-\pi/2)=0,
\qquad J^*-J(-\pi/2)=\tfrac12.
\]
Thus $\tfrac12\norm{\nabla J(\theta)}^2\ge\mu(J^*-J(\theta))$ fails for every $\mu>0$. Equivalently, the minimization PL inequality fails for $g=-J$. Since both action probabilities lie in $[1/4,3/4]$, their log derivatives are smooth and bounded on the compact domain. The ordinary discounted action values are
$Q_\theta(a)=r(a)+\gamma\pi_\theta(1)/(1-\gamma)$, which an exact two-entry tabular critic represents.
\end{proof}

The example establishes that generic actor smoothness, positive probabilities, and critic realizability do not \emph{imply} uniform PL. It does not contradict a theorem that imposes PL separately. It is not a witness that our entire actor/assumption class strictly contains Gaur et al.'s class.

\subsection{KL-regularized counterexamples and degenerate neural policies}
\label{app:pl-strengthened}
The example in \cref{prop:PL-counterexample} concerns the unregularized return. We now give an additional smooth policy with a spurious strict local minimum for every positive KL-regularization level, and explain a related obstruction in multilayer neural parameterizations.

\subsubsection{Setting and the PL condition}

Consider a discounted MDP with finite action set $\mathcal A$, a strictly
positive reference policy $\piRef$, and a parametric policy family
$\{\pi_\theta:\theta\in\Theta\}$. For an upper parameter $x$ and $\tau\ge0$, the
normalized lower cost is
\begin{equation}
  g_\tau(x,\theta)
  =(1-\gamma)\,\E_{\pi_\theta}\sum_{t\ge0}\gamma^t
   \Bigl[-r_x(s_t,a_t)+\tau\log\frac{\pi_\theta(a_t\mid s_t)}{\piRef(a_t\mid s_t)}\Bigr],
  \label{plx:eq:g}
\end{equation}
as in \eqref{eq:main-lower}, so the regularizer is
$\KL(\pi_\theta\|\piRef)$ averaged over the discounted occupancy. Write
$g_\tau^\star(x)=\inf_{\theta\in\Theta}g_\tau(x,\theta)$.

A uniform lower PL condition with constant $\mu>0$ on $\Theta$ requires, for
every $x$ in the upper domain and every $\theta\in\Theta$,
\begin{equation}
  \tfrac12\,\|\nabla_\theta g_\tau(x,\theta)\|^2
  \;\ge\;\mu\,\bigl(g_\tau(x,\theta)-g_\tau^\star(x)\bigr).
  \label{plx:eq:PL}
\end{equation}
Consequently, a single pair $(x,\bar\theta)$ with
$\nabla_\theta g_\tau(x,\bar\theta)=0$ and
$g_\tau(x,\bar\theta)>g_\tau^\star(x)$ rules out \eqref{plx:eq:PL} for every
$\mu>0$.

\subsubsection{A spurious local minimum for every regularization level}

\paragraph{One-state reduction.}
Take one state with a self-loop, $\mathcal A=\{0,1\}$, $\piRef=(\tfrac12,\tfrac12)$,
and reward $r_x(1)=u(x)$, $r_x(0)=0$. Write $p=\pi_\theta(1)$ and
\[
  k(p)=\KL\bigl(\mathrm{Bern}(p)\,\|\,\mathrm{Bern}(\tfrac12)\bigr)
      =p\log(2p)+(1-p)\log(2(1-p)).
\]
The state is visited at every step, so \eqref{plx:eq:g} reduces to
\begin{equation}
  g_\tau(x,\theta)=h_{u,\tau}(p(\theta)),\qquad
  h_{u,\tau}(p)=-u\,p+\tau\,k(p).
  \label{plx:eq:reduction}
\end{equation}
For $\tau>0$, $h_{u,\tau}$ is strictly convex on $(0,1)$, with
$h_{u,\tau}'(p)=-u+\tau\log\frac{p}{1-p}$.

\paragraph{The policy family.}
For $\theta\in\R$ (equivalently, on the circle), let
\begin{equation}
  p(\theta)=\frac12+\frac{\cos\theta+\tfrac12\cos2\theta}{6}.
  \label{plx:eq:family}
\end{equation}

\begin{lemma}[Properties of the family]\label{plx:lem:family}
The map $p$ is $C^\infty$ and $2\pi$-periodic, and its range is exactly
$[\tfrac38,\tfrac34]$. Within each period it has a strict global maximum $p(0)=\tfrac34$, strict
global minima $p(\pm\tfrac{2\pi}3)=\tfrac38$, and a strict local maximum
$p(\pi)=\tfrac5{12}$. All log-policy derivatives of every order are bounded.
\end{lemma}

\begin{proof}
Put $\varphi(\theta)=\cos\theta+\tfrac12\cos2\theta$. Then
$\varphi'(\theta)=-\sin\theta\,(1+2\cos\theta)$, so the critical points in
$(-\pi,\pi]$ are $\theta\in\{0,\pi,\pm\tfrac{2\pi}3\}$, with values
$\varphi(0)=\tfrac32$, $\varphi(\pi)=-\tfrac12$ and
$\varphi(\pm\tfrac{2\pi}3)=-\tfrac34$. Since
$\varphi''(\theta)=-\cos\theta-2\cos2\theta$, we get $\varphi''(0)=-3$,
$\varphi''(\pi)=-1$ and $\varphi''(\pm\tfrac{2\pi}3)=\tfrac32$. This gives the
classification and the range $\varphi\in[-\tfrac34,\tfrac32]$, hence
$p\in[\tfrac38,\tfrac34]$. Both action probabilities lie in
$[\tfrac14,\tfrac58]\cup[\tfrac38,\tfrac34]\subset[\tfrac14,\tfrac34]$, so
$\log\pi_\theta$ is a smooth function of a trigonometric polynomial bounded away
from the singularities of $\log$, and all of its derivatives are bounded.
\end{proof}

\begin{proposition}[Spurious local minimum for every $\tau$]\label{plx:prop:spurious}
Let $u\ge0$ and $\tau>0$. In the setting \eqref{plx:eq:reduction}--\eqref{plx:eq:family}:
\begin{enumerate}[label=(\roman*),itemsep=2pt]
  \item $\bar\theta=\pi$ is a strict local minimizer of $g_\tau(x,\cdot)$ with
  $\nabla_\theta g_\tau(x,\pi)=0$;
  \item its suboptimality gap satisfies
  \begin{equation}
    g_\tau(x,\pi)-g_\tau^\star(x)
    \;\ge\;\frac{u}{12}+\tau\,k\!\left(\tfrac5{12}\right)\;>\;0;
    \label{plx:eq:gap}
  \end{equation}
  \item consequently, \eqref{plx:eq:PL} fails for every $\mu>0$ on any parameter set
  containing $\pi$ and a point $\theta$ with $p(\theta)=\tfrac12$.
\end{enumerate}
Here $k(\tfrac5{12})=\tfrac5{12}\log\tfrac56+\tfrac7{12}\log\tfrac76\approx0.01395$.
\end{proposition}

\begin{proof}
(i) By the chain rule,
$\nabla_\theta g_\tau(x,\theta)=h'_{u,\tau}(p(\theta))\,p'(\theta)$, and
$p'(\pi)=0$ by Lemma~\ref{plx:lem:family}. At $p=\tfrac5{12}$,
$h'_{u,\tau}(\tfrac5{12})=-u+\tau\log\tfrac57<0$, because $u\ge0$, $\tau>0$ and
$\log\tfrac57<0$. Differentiating once more at a critical point of $p$ gives
\[
  \partial_\theta^2 g_\tau(x,\pi)
  =h''_{u,\tau}(p(\pi))\,p'(\pi)^2+h'_{u,\tau}(p(\pi))\,p''(\pi)
  =h'_{u,\tau}(\tfrac5{12})\cdot\bigl(-\tfrac16\bigr)>0,
\]
so $\pi$ is a strict local minimizer.

(ii) The range of $p$ contains $\tfrac12$, so
$g_\tau^\star(x)\le h_{u,\tau}(\tfrac12)$. Since $k(\tfrac12)=0$,
\[
  h_{u,\tau}(\tfrac5{12})-h_{u,\tau}(\tfrac12)
  =u\bigl(\tfrac12-\tfrac5{12}\bigr)+\tau\bigl(k(\tfrac5{12})-0\bigr)
  =\frac u{12}+\tau\,k(\tfrac5{12}).
\]
This quantity is positive because $k(\tfrac5{12})>0$ and $\tau>0$.

(iii) Combine (i) and (ii) with the observation following \eqref{plx:eq:PL}.
\end{proof}

\begin{remark}[What the proposition adds]\label{plx:rem:adds}
\begin{enumerate}[label=(\alph*),itemsep=2pt]
  \item \emph{Regularization does not repair PL.} The lower bound in \eqref{plx:eq:gap}
  is positive for every $\tau>0$ and increases with $\tau$. For $u>0$ it is
  also bounded below by $u/12$ uniformly over $\tau>0$. The bad point
  is a genuine local minimizer, not a local maximizer, so gradient methods can
  converge to it from a neighborhood.
  \item \emph{Uniformity in the upper parameter.} The conclusion holds for
  every $x$ with $u(x)\ge0$, so it applies throughout any upper domain on which
  the reward scale is nonnegative, for example $u=\tanh$ on $[0,\infty)$.
  \item \emph{Exact value representation.} Write
  $q(a)=-r_x(a)+\tau\log(\pi_\theta(a)/\piRef(a))$ and
  $\bar q=\sum_a\pi_\theta(a)q(a)$. The regularized cost action values are
  $Q(a)=q(a)+\gamma\bar q/(1-\gamma)$, so a two-entry tabular critic
  represents them exactly, as in \cref{prop:PL-counterexample}.
  \item \emph{The original example for every $\tau$.} For the earlier
  family $p(\theta)=\tfrac12+\tfrac14\sin\theta$, the same reduction gives
  $\nabla_\theta g_\tau(x,\pm\tfrac\pi2)=0$ for every $\tau$. By symmetry of
  $k$ about $\tfrac12$, $g_\tau(x,-\tfrac\pi2)-g_\tau(x,\tfrac\pi2)=u/2$ for
  every $\tau$. That example therefore also extends to all regularization
  levels when $u>0$, but its bad point is a local maximizer.
\end{enumerate}
\end{remark}

\subsubsection{Degenerate policy Jacobians and multilayer networks}

\begin{lemma}[Vanishing policy Jacobian forces a critical point]\label{plx:lem:jacobian}
Let $\mathcal A$ be finite, the rewards bounded, and $\piRef$ strictly
positive. Suppose $R_\theta=\pi_\theta/\piRef$ is differentiable at
$\bar\theta$ as a map into bounded functions with the supremum norm,
and that $0<c\le R_\theta(s,a)\le C<\infty$ on a neighborhood of
$\bar\theta$. If $D_\theta\pi_{\bar\theta}(a\mid s)=0$ for every $(s,a)$,
then $\nabla_\theta g_\tau(x,\bar\theta)=0$ for every $x$ and finite
$\tau\ge0$. If this point is suboptimal for some $x$, uniform lower PL
fails on any parameter set containing it and a strictly better point.
The differentiability and ratio bounds hold for finite state spaces
with smooth positive policies and for smooth bounded-input neural
exponential tilts on bounded parameter neighborhoods.
\end{lemma}

\begin{proof}
The pointwise derivative condition implies that the derivative of
$R_\theta$ at $\bar\theta$ is the zero bounded function. As a function
of $R$, the state transition operator is affine and bounded in the
supremum norm. Its discounted resolvent $(I-\gamma T_\pi)^{-1}$ is
Fr\'echet differentiable, by the resolvent identity and
$\|(I-\gamma T_\pi)^{-1}\|\le(1-\gamma)^{-1}$.
The expected one-step cost is
\[
 \sum_a\piRef(a\mid s)R(s,a)
       [-r_x(s,a)+\tau\log R(s,a)],
\]
which is differentiable in the same norm because $R$ stays in
$[c,C]$. Thus the normalized discounted objective is a differentiable
function of $R$, and the chain rule gives a zero parameter gradient.
A strictly positive objective gap then contradicts \eqref{plx:eq:PL}.
\end{proof}

\begin{proposition}[Zero hidden weights are degenerate for depth at least two]\label{plx:prop:deep}
Let $\sigma$ be continuously differentiable with $\sigma(0)=0$ (e.g.\ $\tanh$). For depth
$L\ge2$, arbitrary widths, bounded input $\psi(s,a)$ and fixed readout $b$,
consider
\[
 \begin{gathered}
  h^0=\psi,\qquad
  h^\ell=\sigma\bigl(W_\ell h^{\ell-1}\bigr),\quad \ell=1,\dots,L,\\
  F_\theta(s,a)=b^{\mathsf T}h^L,\qquad
  \pi_\theta(a\mid s)\propto\piRef(a\mid s)\,e^{F_\theta(s,a)}.
 \end{gathered}
\]
with $\theta=(W_1,\dots,W_L)$, and let $\bar\theta$ have $W_\ell=0$ for all
$\ell$. Then $D_\theta F_{\bar\theta}(s,a)=0$ and
$D_\theta\pi_{\bar\theta}(a\mid s)=0$ for all $(s,a)$, and
$\pi_{\bar\theta}=\piRef$. The same holds for $L=1$ if the readout $b$ is also
trainable and set to zero. Any normalization constants such as
$m_{\ell-1}^{-1/2}$ do not change the conclusion.
\end{proposition}

\begin{proof}
At $\bar\theta$, $h^1=\sigma(0)=0$, and inductively $h^\ell=0$ for all
$\ell\ge1$. Backpropagation gives
$\partial F/\partial W_\ell=\delta_\ell\,(h^{\ell-1})^{\mathsf T}$, where
$\delta_L=b\odot\sigma'(0)$ and
$\delta_\ell=(W_{\ell+1}^{\mathsf T}\delta_{\ell+1})\odot\sigma'(0)$ for
$\ell<L$. For $\ell\ge2$, $h^{\ell-1}=0$. For $\ell=1$, $\delta_1$ contains the
factor $W_2^{\mathsf T}=0$, which uses $L\ge2$. Hence $D_\theta F_{\bar\theta}=0$.
Since $F_{\bar\theta}=0$, $\pi_{\bar\theta}=\piRef$, and
$D_\theta\pi_{\bar\theta}=D_F\pi\circ D_\theta F_{\bar\theta}=0$. For $L=1$
with trainable $b=0$, $\partial F/\partial b=h^1=0$ and
$\partial F/\partial W_1=(b\odot\sigma'(0))\psi^{\mathsf T}=0$.
\end{proof}

\begin{corollary}[Uniform PL fails on domains containing zero hidden weights]\label{plx:cor:deep}
In the setting of Proposition~\ref{plx:prop:deep}, suppose that for some upper
parameter $x$ and some $\theta\in\Theta$, $g_\tau(x,\theta)<g_\tau(x,\bar\theta)$.
Then uniform lower PL \eqref{plx:eq:PL} fails on $\Theta$ for every $\mu>0$ and every
$\tau\ge0$ for which this strict inequality holds.
\end{corollary}

\begin{proof}
Combine Proposition~\ref{plx:prop:deep} with Lemma~\ref{plx:lem:jacobian}.
\end{proof}

\paragraph{A concrete instance.}
Take a one-state bandit with $\mathcal A=\{-1,+1\}$, $\piRef=\tfrac12$,
$r_x(a)=u(x)\,a$ with $u(x)>0$, input $\psi(a)=a$, depth $L=2$, width one,
$\tanh$ activations and readout $b=1$. Then
$F_\theta(a)=\tanh(w_2\tanh(w_1a))$, which is odd in $a$, so
\[
  \pi_\theta(+1)=\sigma_{\mathrm{log}}\bigl(2\tanh(w_2\tanh w_1)\bigr),
  \qquad \sigma_{\mathrm{log}}(v)=(1+e^{-v})^{-1}.
\]
Writing $q=\pi_\theta(+1)$, the cost is
$\tilde h(q)=-u(2q-1)+\tau\,k(q)$, with $\tilde h'(\tfrac12)=-2u<0$. Taking
$w_1=w_2=w>0$ small gives $q\in(\tfrac12,\tfrac12+\delta)$ for small $\delta$,
hence $g_\tau(x,\theta)<\tilde h(\tfrac12)=g_\tau(x,\bar\theta)$. So
$\bar\theta=(0,0)$ is a critical point that is suboptimal for every $x$ with
$u(x)>0$ and every $\tau\ge0$. When $u<\tau$, the regularized optimum
$q^\star=\sigma_{\mathrm{log}}(2u/\tau)$ is representable, and the gap equals
$\tau\log\cosh(u/\tau)$.

\begin{remark}[Continuous actions]
Lemma~\ref{plx:lem:jacobian} carries over to Gaussian policies
$\mathcal N(\mu_\theta(s),\Sigma)$ with fixed covariance, provided the
occupancy and cost are differentiable in the mean map. If the mean network has
all hidden weights zero and depth at least two, then $D_\theta\mu_\theta=0$ by
the argument of Proposition~\ref{plx:prop:deep}, and the same conclusion follows.
This gives the analogous obstruction for fixed-covariance continuous-action policies under the stated differentiability condition.
\end{remark}

\subsubsection{Relation to the local assumptions}

\begin{remark}[Why the local analysis avoids these points]
\begin{enumerate}[label=(\alph*),itemsep=2pt]
  \item A vanishing policy Jacobian forces the centered score
  $Z-\E_{\pi}Z$ to vanish, so its conditional covariance is zero.
  For the zero-hidden-weight construction, $Z=D_\theta F$ itself also
  vanishes. Either case prevents an A2 certificate at that point.
  Assumption~\ref{ass:gap} requires A2 at
  the reference point and localizes the analysis within the radius on which
  the covariance gap persists (Appendix~\ref{app:support-actor}). Degenerate points are therefore
  excluded from the certified neighborhood by construction.
  \item This also explains why the two-hidden-layer construction in
  \cref{crex:rem:depth-two} uses a nonzero second-layer initialization:
  all-zero hidden weights would give the degeneracy of
  Proposition~\ref{plx:prop:deep}.
  \item In Proposition~\ref{plx:prop:spurious}, a neighborhood of a point with
  $p(\theta_0)=\tfrac12$ and $p'(\theta_0)\neq0$ is the analogue of the
  certified local region. For sufficiently large $\tau$ the unique local
  response lies near $p=\tfrac12$ and away from $\theta=\pi$.
\end{enumerate}
\end{remark}

\begin{remark}[Scope of the claim]
These results show that uniform lower PL is not implied by smoothness,
positivity, exact value representation, or reference-KL regularization at any
level. For the neural architectures in \cref{plx:prop:deep}, PL also
fails on domains containing zero hidden weights whenever that point is
suboptimal. A theorem that assumes PL separately, such as \citet{gaur2025sample},
is not contradicted by these examples. They do not establish containment
between that assumption class and ours; the obstruction concerns domains
that actually contain a suboptimal degenerate point.
\end{remark}

\providecommand{\jR}{\mathbb R}
\providecommand{\jE}{\mathbb E}
\providecommand{\jId}{\mathrm{Id}}
\providecommand{\jKL}{\operatorname{KL}}
\providecommand{\jCov}{\operatorname{Cov}}
\providecommand{\jUnif}{\operatorname{Unif}}
\providecommand{\jSpan}{\operatorname{span}}
\providecommand{\jsech}{\operatorname{sech}}
\providecommand{\jatanh}{\operatorname{arctanh}}
\providecommand{\jref}{\mathrm{ref}}
\subsection{A concrete continuous-state instance with a nontrivial response correction}
\label{crex:example}
We give a joint satisfiability example for A1, A2, G, and C1--C2 in a
class with finitely many policy contexts. It has continuous state
variation in rewards and transition probabilities, a nonlinear tied
actor, and five fixed critic features. An attainable preference teacher
produces a finite interior stationary point of the response-induced
upper objective, at which the response-chain contribution is nonzero.
The population illustration below evaluates this objective independently.
Appendix~\ref{app:continuous-validation} additionally implements sampled
single-loop training on the same instance, using its two-hidden-layer
actor specialization and the persistent nonlinear paired critic.

\subsubsection{A class with finite policy contexts and Bellman closure}
\label{crex:sec:class}

\begin{proposition}[A sufficient structural certificate for C1--C2]
\label{crex:prop:class}
Let $\mathcal S$ be compact, $\mathcal A$ finite, and
$\{\mathcal S_j\}_{j=1}^{M}$ a measurable partition. Write $j(s)$ for
the cell index. Suppose the following conditions hold.
\begin{enumerate}[label=(\roman*),itemsep=2pt,topsep=2pt]
\item The fixed transition kernel has a finite-rank mixture
$P(ds'\mid s,a)=\sum_{\ell=1}^{M'}w_\ell(s,a)\mu_\ell(ds')$,
where the $\mu_\ell$ are probability measures and the measurable
weights are nonnegative and sum to one.
\item The tied actor uses the bounded, fixed input
$\psi(s,a)=\psi_0(j(s),a)$, with fixed hidden depth and the normalized
initialization, activation, and chart conditions of Appendix~\ref{app:support-actor}.
The reference policy depends only on $(j(s),a)$ and is strictly
positive on every such pair.
\item The fixed feature map $\varphi_Q:\mathcal S\times\mathcal A
\to\jR^{d_Q}$ is bounded. Its span contains all cell--action
indicators, all $w_\ell$, and every bounded reward $r_x$.
\item For some $\lambda_0>0$,
$\jE_{\rho(ds)\pi_{\jref}(da\mid s)}
[\varphi_Q\varphi_Q^{\mathsf T}]\succeq\lambda_0 I_{d_Q}$.
\end{enumerate}
Then C2 holds for every finite $\tau>0$, every $x$, and every finite
actor parameter. On an ambient ball $\|\alpha\|\le A$, C1 holds
with
\[
 \lambda_Q=(1-\gamma)c_A\lambda_0,\qquad
 c_A=\exp(-2C_{F,1}A),
\]
where $C_{F,1}$ is a width-uniform bound on $\|D_\alpha F\|$ there
and the tied construction has $F_0=0$.
\end{proposition}
\begin{proof}
At fixed parameters the neural score is constant on each
cell--action pair. The finite-action log normalizer is therefore
constant within each state cell. Both terms of
$\log(\pi_\alpha/\pi_{\jref})$ belong to the cell-indicator span.
The regularized cost $q_{x,\alpha}=-r_x+
\tau\log(\pi_\alpha/\pi_{\jref})$ thus belongs to
$V=\jSpan\varphi_Q$.

For every bounded $v$,
\[
 K_\alpha v(s,a)
 =\sum_{\ell=1}^{M'}w_\ell(s,a)
   \int\sum_{a'}\pi_\alpha(a'\mid s')v(s',a')\,\mu_\ell(ds')
 \in V.
\]
The Neumann series $\sum_{k\ge0}\gamma^kK_\alpha^kq_{x,\alpha}$
converges uniformly and stays in the closed finite-dimensional
space $V$. This proves C2. On the ambient ball,
$|F_\alpha|\le C_{F,1}A$ and
$\pi_\alpha/\pi_{\jref}\ge c_A$. Therefore
$\rho_U\ge c_A\rho\pi_{\jref}$, and
$\nu_U\ge(1-\gamma)\rho_U$ transfers the reference Gram bound
to C1.
\end{proof}

The bounded-feature condition is needed for C1; compactness alone
does not bound an arbitrary measurable feature map. The normalized
network conditions in the proposition supply A1(iii). Reward and
preference regularity and A2 remain separate requirements, verified
for the instance below. This proposition is a sufficient certificate
within our broader assumption-based theorem.

\begin{proposition}[A log-normalizer obstruction under finite-rank dynamics]
\label{crex:prop:obstruction}
Let $\mathcal S=[-1,1]$, $\mathcal A=\{-1,+1\}$,
$\pi_{\jref}=1/2$, and $F_\alpha(s,a)=a\alpha s$, for $\alpha$ in
an open interval containing zero. No fixed finite-dimensional
subspace of bounded functions contains
$\log(\pi_\alpha/\pi_{\jref})$ for every such $\alpha$.
If, in addition, the transition kernel has a fixed finite-rank
mixture as in Proposition~\ref{crex:prop:class}, then, for every
fixed $\tau>0$ and fixed bounded reward, no fixed finite feature
space can represent all corresponding regularized values.
\end{proposition}
\begin{proof}
The log-policy ratio equals
$a\alpha s-\log\cosh(\alpha s)$. If these functions belonged to a
finite-dimensional space $V$, uniform finite differences and
closedness would put all their $\alpha$-derivatives at zero in
$V$. Uniform convergence of these differences follows from
smoothness on a compact neighborhood in $(\alpha,s)$.

Every even Taylor coefficient of $\log\cosh y$ is nonzero.
For completeness, write
$\tanh y=\sum_{k\ge1}(-1)^{k-1}a_k y^{2k-1}$ near zero.
The identity $(\tanh y)'=1-\tanh^2 y$ gives
$a_1=1$ and
$(2k-1)a_k=\sum_{j=1}^{k-1}a_ja_{k-j}>0$ for $k\ge2$.
Integration gives nonzero coefficients
$(-1)^{k-1}a_k/(2k)$ in $\log\cosh y$.
Thus $V$ would contain every $s^{2k}$, $k\ge1$,
contradicting their linear independence.

For the value claim let $W=\jSpan\{w_\ell\}_{\ell=1}^{M'}$.
The finite-rank kernel implies $K_\alpha v\in W$ for every bounded
$v$. If all $Q_\alpha$ belonged to a fixed finite-dimensional
$V$, the Bellman equation would imply
\[
 \tau\log(\pi_\alpha/\pi_{\jref})
 =Q_\alpha-\gamma K_\alpha Q_\alpha+r
 \in V+W+\jSpan\{r\},
\]
contradicting the first part.
\end{proof}

\begin{remark}[The obstruction is not a necessity theorem for finite contexts]
\label{crex:rem:obstruction-scope}
The value conclusion above uses the finite-rank transition condition;
failure of log-ratio membership alone does not imply failure of C2
for an arbitrary policy-dependent Bellman operator. Nor does this
binary-logit example characterize every finite-action policy family.
For instance, with three actions, let
$q(s)=1/2+s/4$ and
$\pi_{\jref}(s)=(q(s)/2,(1-q(s))/2,1/2)$.
Take $v=(1,1,-1)$ and $F_\alpha(s,a)=h(\alpha)v_a$.
The log normalizer is $\log\cosh h(\alpha)$, independent of $s$,
so the log-policy ratio lies in $\jSpan\{1,v\}$ although the policy
varies continuously with $s$. With reset dynamics
$P(\cdot\mid s,a)=\rho$, reward $\tanh(x)s v_a$, and features
$(1,v_a,s v_a)$, C2 follows immediately. Under uniform $\rho$ the
reference Gram matrix is $\operatorname{diag}(1,1,1/3)$, so C1
also holds on bounded actor neighborhoods. Finite policy contexts
are consequently a useful sufficient construction, not an additional
necessary restriction on the general theorem.
\end{remark}

\begin{remark}[Retained dimension and exact A2 calculation]
\label{crex:rem:dimension}
For a cell-constant initialization feature $Z_0$, each conditional
covariance has rank at most $|\mathcal A|-1$. Hence
\[
 \operatorname{rank}\mathcal G_0
 \le M(|\mathcal A|-1),\qquad
 \text{A2 implies }r\le M(|\mathcal A|-1).
\]
When the reset cell masses are known, the certificate is the explicit
finite sum
\[
 \mathcal G_0
 =\sum_{j=1}^M\rho(\mathcal S_j)
   \jCov_{a\sim\pi_{\jref}(\cdot\mid j)}(Z_0(j,a)).
\]
No pilot samples are required to evaluate this sum for the example
below. If the cell masses are unknown, they must instead be bounded
or estimated; finite encoding alone does not remove that statistical
requirement. The example attains the dimension bound:
$r=2=M(|\mathcal A|-1)$.
\end{remark}

\subsubsection{The controlled process and preference model}
\label{crex:sec:process}

Fix any discount factor $\gamma\in(0,1)$. Set
\[
 \mathcal S=[-1,1],\qquad \mathcal A=\{-1,+1\},\qquad
 \rho=\jUnif[-1,1].
\]
Use the Borel sigma algebra on $\mathcal S$. Define
\[
 I_-=[-1,0),\qquad I_+=[0,1],\qquad
 i(s)=\begin{cases}-1,&s<0,\\+1,&s\ge0,\end{cases}
\]
and let $\mu_-$ and $\mu_+$ be the uniform probability measures on
$I_-$ and $I_+$, respectively. The endpoint convention is immaterial
for these measures and fixes all functions pointwise.
Define the transition kernel
\begin{equation}
 P(ds'\mid s,a)
 =\left(\frac12+\frac{as}{4}\right)\mu_+(ds')
  +\left(\frac12-\frac{as}{4}\right)\mu_-(ds').
 \label{crex:eq:transition}
\end{equation}
Both mixture weights lie in $[1/4,3/4]$. Thus this is a measurable
probability kernel with continuous state distributions. Its mixture
weights depend continuously on $s$ and depend on the action.
The reset law is $\rho=(\mu_-+\mu_+)/2$.

The upper parameter is $x\in\jR$. Write $u(x)=\tanh x$ and set
\begin{equation}
 r_x(s,a)=u(x)\,as.
 \label{crex:eq:reward}
\end{equation}
Only the reward model and the fitted preference loss depend on $x$;
the transition kernel, reset law, and reference policy remain fixed.

For the upper objective, independently sample two trajectories
$\xi^1,\xi^2$, each with $K=2$ state--action pairs, using the current
actor, the reset law, and \eqref{crex:eq:transition}. Put
\begin{align}
 D(\boldsymbol\xi)
 &=\sum_{t=0}^{1}(a_t^1s_t^1-a_t^2s_t^2),
 &\Delta_x&=u(x)D(\boldsymbol\xi), \notag\\
 \mathbb P(Y=1\mid\boldsymbol\xi)
 &=\sigma(\kappa D(\boldsymbol\xi)),
 &\sigma(v)&=(1+e^{-v})^{-1}, \label{crex:eq:feedback}\\
 L(x;\boldsymbol\xi,Y)
 &=\log(1+e^{\Delta_x})-Y\Delta_x.
 \label{crex:eq:loss}
\end{align}
Fix $\kappa\in(0,1)$, with $\kappa=1/2$ in the numerical illustration.
Here the teacher reward is the fixed function $r^\dagger(s,a)=\kappa as$.
Conditional on the trajectories, the label law has no dependence on
$x$ or on the actor parameter. The loss has no additional explicit
actor dependence. Fresh tuples and labels can be drawn independently.

\subsubsection{A tied neural actor with an explicit initialization gap}
\label{crex:sec:actor}

Use $\pi_{\jref}(a\mid s)=1/2$. For every integer $n\ge1$ and retained
parameter $\alpha=(\alpha_-,\alpha_+)\in\jR^2$, define
\begin{align}
 h_n(t)&=\sqrt n\,\tanh\!\left(\frac{t}{2\sqrt n}\right),&
 F_{n,\alpha}(s,a)&=a h_n(\alpha_{i(s)}), \label{crex:eq:score}\\
 \pi_{n,\alpha}(a\mid s)
 &=\frac{\exp\{a h_n(\alpha_{i(s)})\}}
         {2\cosh(h_n(\alpha_{i(s)}))}.
 \label{crex:eq:policy}
\end{align}
Write $\pi_{n,\alpha}(a\mid i)$ for this region-constant probability.
The two regions have independently trainable action preferences. In
particular, the policy is state dependent whenever
$h_n(\alpha_-)\ne h_n(\alpha_+)$.

\begin{lemma}[Realization by our tied architecture]
\label{crex:lem:architecture}
The score in \eqref{crex:eq:score} is an exact one-hidden-layer
specialization of the normalized antithetic actor, with input dimension
$2$, hidden width $2n$, and retained dimension $2$. It satisfies the
actor part of A1 uniformly in $n$ and satisfies A2 with
$\lambda_{\mathrm{eff},L}=1/8$.
\end{lemma}

\begin{proof}
Let $e_-,e_+$ be the coordinate vectors in $\jR^2$ and use the bounded
input $\psi(s,a)=a e_{i(s)}$. Index the $2n$ hidden units by $(i,k)$,
where $i\in\{-1,+1\}$ and $k=1,\ldots,n$. For a
$2n$-by-$2$ weight matrix $W$, use the normalized branch
\[
 G_{n,W}(s,a)
 =\frac1{\sqrt{2n}}\mathbf 1_{2n}^{\mathsf T}
   \tanh\!\left(\frac{W\psi(s,a)}{\sqrt2}\right).
\]
The activation acts coordinatewise and the readout is fixed.
Initialize $W_0=0$ and define the fixed chart $B_n$ by
\[
 (B_n\alpha)_{(i,k),j}
 =\frac{\alpha_i}{\sqrt n}\,\mathbf 1_{\{j=i\}}.
\]
Then $\|B_n\alpha\|_F^2=\alpha_-^2+\alpha_+^2$, so $B_n$ is an
isometry. Take the two branches
$W^\pm(\alpha)=\pm B_n\alpha/\sqrt2$ and their antithetic score
$F=(G_{n,W^+}-G_{n,W^-})/\sqrt2$. At an input in region $i$, exactly
$n$ units have nonzero preactivation, equal to
$\pm a\alpha_i/(2\sqrt n)$. Oddness of $\tanh$ gives
\[
 G_{n,W^\pm}(s,a)
 =\pm a\sqrt{\frac n2}\tanh\!\left(\frac{\alpha_i}{2\sqrt n}\right),
\]
and therefore gives exactly \eqref{crex:eq:score}.
The initialization satisfies the normalized spectral requirement,
and all hidden weights vary through the specified fixed chart.

For $k=1,\ldots,4$,
\begin{equation}
 h_n^{(k)}(t)
 =2^{-k}n^{(1-k)/2}
   \tanh^{(k)}\!\left(\frac{t}{2\sqrt n}\right).
 \label{crex:eq:derivatives}
\end{equation}
All derivatives of $\tanh$ in this display are bounded on $\jR$.
Since $n^{(1-k)/2}\le1$, the required operator norms of
$D_\alpha^kF_{n,\alpha}$ are bounded independently of $n$, states,
actions, and points in any fixed ambient ball.
Also $F_{n,0}=0$, $D_\alpha^2F_{n,0}=0$, and $h_n'(0)=1/2$.
Thus the initialization feature is
\[
 Z_{n,0}(s,a)=D_\alpha F_{n,0}(s,a)=\frac a2e_{i(s)}.
\]
Its conditional mean under the uniform reference policy is zero, and
its conditional covariance is $e_i e_i^{\mathsf T}/4$ in region $i$.
Both regions have reset probability $1/2$, whence
\begin{equation}
 \mathcal G_{n,0}
 :=\jE_{s\sim\rho}\jCov_{a\sim\pi_{\jref}(\cdot\mid s)}
      (Z_{n,0}(s,a))
 =\frac18 I_2.
 \label{crex:eq:actor-gap}
\end{equation}
\end{proof}

\begin{lemma}[Reward and upper-loss regularity]
\label{crex:lem:regularity}
The reward and sample loss satisfy the remaining parts of A1 and all
of G on the unrestricted upper domain $\jR$, with
\begin{equation}
 B_r=1,\quad G_r=1,\quad L_r=2,\qquad
 B_L=\log2+4,\quad G_L=4,\quad L_L=12.
 \label{crex:eq:global-constants}
\end{equation}
\end{lemma}
\begin{proof}
The bounds $|u|\le1$, $|u'|\le1$, and $|u''|\le2$, together with
$|as|\le1$, give the reward constants. The reward is $C^\infty$ in $x$,
with bounded derivatives through order four uniformly in $(s,a)$.

Since $|D|\le4$, we have
$|\Delta_x|\le4$, $|\partial_x\Delta_x|\le4$, and
$|\partial_x^2\Delta_x|\le8$. For each realized $Y\in\{0,1\}$,
\begin{align*}
 0\le L&\le\log2+|\Delta_x|,\\
 \partial_xL&=(\sigma(\Delta_x)-Y)\partial_x\Delta_x,\\
 \partial_x^2L
 &=\sigma'(\Delta_x)(\partial_x\Delta_x)^2
   +(\sigma(\Delta_x)-Y)\partial_x^2\Delta_x.
\end{align*}
Use $|\sigma-Y|\le1$ and $0\le\sigma'\le1/4$ to obtain
\eqref{crex:eq:global-constants}. Composition with the smooth
softplus function, with $|D|\le4$, also supplies the samplewise
$C^4$ bounds required by A1. Equation \eqref{crex:eq:feedback}
provides G's parameter-independent conditional feedback law;
the sampling construction provides its fixed tuple size, finite
trajectory length, and fresh sample access.
\end{proof}

\begin{remark}[A concrete two-hidden-layer variant]
\label{crex:rem:depth-two}
Cell constancy is preserved at every fixed hidden depth, but A2 must
still be checked. In particular, initializing all hidden weights at
zero with $\tanh$ gives zero parameter gradients at depth at least
two, and therefore fails A2.

Here is an explicit alternative that trains both hidden layers and
has a gap independent of width. Give each hidden layer width $2n$,
with $n$ units for each region. Use the same input and fixed all-ones
readout as above. Set $W^1_0=0$ and let $W^2_0$ be block diagonal,
each $n$-by-$n$ block having every entry $\sqrt{2/n}$.
Then $\|W^2_0/\sqrt{2n}\|_{\mathrm{op}}=1$.
Define a product-space chart by
\[
 (B_n^1\alpha)_{(i,k),j}
 =\frac{\alpha_i}{\sqrt{2n}}\mathbf1_{\{j=i\}},\qquad
 (B_n^2\alpha)_{(i,k),(j,\ell)}
 =\frac{\alpha_i}{\sqrt2\,n}\mathbf1_{\{j=i\}}.
\]
Each chart component has squared Frobenius norm
$\|\alpha\|^2/2$, so the combined chart is an isometry.
Use $W^{\ell,\pm}=W^\ell_0\pm B_n^\ell\alpha/\sqrt2$,
our layer normalizations, and its antithetic readout.
Writing $t_i=\alpha_i/(2\sqrt{2n})$, direct substitution gives
\begin{equation}
 F^{(2)}_{n,\alpha}(s,a)=aH_n(\alpha_{i(s)}),\qquad
 H_n(\alpha_i)=\frac{\sqrt n}{2}
 \left[
  \tanh((1+t_i)\tanh t_i)+
  \tanh((1-t_i)\tanh t_i)
 \right].
 \label{crex:eq:depth-two}
\end{equation}
Thus $H_n'(0)=1/(2\sqrt2)$ and
$\mathcal G^{(2)}_{n,0}=I_2/16$, for every $n$.
The normalized initialization and the isometric chart supply the
width-uniform derivative bounds. By uniform continuity of the
first derivative, $H_n$ is strictly increasing on a common
neighborhood of zero. Choosing $\tau$ sufficiently large puts
the target logits $i\tanh(x)/(2\tau)$ in its local image.
The local response is then
$\alpha_i^*(x)=H_n^{-1}(i\tanh(x)/(2\tau))$.
The same five critic features and all Bellman-closure arguments
remain valid. The induced response policy, and hence the upper
objective and the numerical computations below, are identical to
those for the one-hidden-layer parameterization whenever this
local response is used.
\end{remark}

\subsubsection{Five fixed critic features}
\label{crex:sec:critic}

Define
\begin{equation}
 z(s)=\sqrt{12}\left(s-\frac{i(s)}2\right),\qquad
 \varphi_Q(s,a)=
 \begin{pmatrix}e_{(i(s),a)}\\ a z(s)\end{pmatrix}\in\jR^5,
 \label{crex:eq:features}
\end{equation}
where the first four coordinates are ordered as
$(-1,-1),(-1,+1),(+1,-1),(+1,+1)$.
These features are fixed independently of $x,\alpha,n$, the critic
scale $m_Q$, and target accuracy. On either region,
\begin{equation}
 |z(s)|\le\sqrt3,\qquad
 \jE_{\mu_i}z=0,\qquad \jE_{\mu_i}z^2=1.
 \label{crex:eq:centered-moments}
\end{equation}
Consequently $\|\varphi_Q(s,a)\|_2\le2$.

For a fixed actor width, suppress $n$ in the transition-operator
notation and set
\begin{align}
 (K_\alpha v)(s,a)
 &=\jE_{s'\sim P(\cdot\mid s,a)}
    \sum_{a'\in\mathcal A}\pi_{n,\alpha}(a'\mid s')v(s',a'),
 \notag\\
 \rho_U(ds,da)&=\rho(ds)\pi_{n,U}(da\mid s),&
 \nu_U&=(1-\gamma)\sum_{k=0}^{\infty}\gamma^k\rho_U K_U^k.
 \label{crex:eq:occupancy}
\end{align}

\begin{lemma}[C1: uniform feature excitation]
\label{crex:lem:C1}
For any fixed ambient actor radius $A>0$, C1 holds throughout
$\|U\|_2\le A$, uniformly in $n\ge1$, with
\begin{equation}
 d_Q=5,\qquad F_Q=2,\qquad
 \lambda_Q=\frac{1-\gamma}{2(1+e^A)}.
 \label{crex:eq:critic-constants}
\end{equation}
\end{lemma}
\begin{proof}
The inequality $|\tanh t|\le|t|$ gives
$|h_n(U_i)|\le |U_i|/2\le A/2$, so
\[
 \pi_{n,U}(a\mid i)\ge p_A:=\frac1{1+e^A}.
\]
Within each region the policy is constant in $s$. Therefore the
zero mean in \eqref{crex:eq:centered-moments} makes the cross moments
between the first four features and the fifth feature vanish.
The reset Gram matrix is exactly
\[
 \jE_{\rho_U}[\varphi_Q\varphi_Q^{\mathsf T}]
 =\operatorname{diag}\!\left(
    \left\{\tfrac12\pi_{n,U}(a\mid i)\right\}_{i,a},\,1
   \right)
 \succeq\frac{p_A}{2}I_5.
\]
Since $\nu_U\ge(1-\gamma)\rho_U$ as measures, integration of the
positive semidefinite matrix $\varphi_Q\varphi_Q^{\mathsf T}$
gives \eqref{crex:eq:critic-constants}. This covers every smaller
base-actor neighborhood used in \cref{thm:main}.
\end{proof}

\begin{lemma}[C2: exact realizability of the regularized value]
\label{crex:lem:C2}
For every finite $\tau>0$, $x\in\jR$, $n\ge1$, and
$\alpha\in\jR^2$, the regularized lower action value
\begin{align}
 q_{x,\alpha}(s,a)
 &=-r_x(s,a)+\tau\log
       \frac{\pi_{n,\alpha}(a\mid s)}{\pi_{\jref}(a\mid s)},\notag\\
 Q^{\mathrm{true}}_{x,\alpha}
 &=(\jId-\gamma K_\alpha)^{-1}q_{x,\alpha}
 \label{crex:eq:true-value}
\end{align}
belongs to $\jSpan\varphi_Q$. On $\|\alpha\|_2\le A$,
\begin{equation}
 \|q_{x,\alpha}\|_\infty\le1+\tau A,\qquad
 \|Q^{\mathrm{true}}_{x,\alpha}\|_\infty
 \le\frac{1+\tau A}{1-\gamma},
 \label{crex:eq:value-envelope}
\end{equation}
uniformly in $x$ and $n$.
\end{lemma}
\begin{proof}
First,
\begin{equation}
 as=\frac{a\,i(s)}2+\frac{a z(s)}{\sqrt{12}}.
 \label{crex:eq:reward-span}
\end{equation}
The first term is constant on each region--action pair and the second
is a multiple of the fifth feature. Thus $r_x\in\jSpan\varphi_Q$.
Also
\begin{equation}
 \log\frac{\pi_{n,\alpha}(a\mid s)}{\pi_{\jref}(a\mid s)}
 =a h_n(\alpha_{i(s)})-\log\cosh(h_n(\alpha_{i(s)})),
 \label{crex:eq:KL-span}
\end{equation}
which is constant on each region--action pair.
In particular $q_{x,\alpha}=\varphi_Q^{\mathsf T}c$ with coefficients
\[
 c_{(i,a)}
 =-\frac{u(x)ai}{2}
   +\tau\bigl[a h_n(\alpha_i)-\log\cosh(h_n(\alpha_i))\bigr],
 \qquad c_5=-\frac{u(x)}{\sqrt{12}}.
\]

For any bounded measurable $v$, define
\[
 m_i(v,\alpha)
 =\int_{I_i}\sum_{a'\in\mathcal A}
     \pi_{n,\alpha}(a'\mid s')v(s',a')\,\mu_i(ds').
\]
The transition kernel gives the exact identity
\begin{equation}
 (K_\alpha v)(s,a)
 =\frac{m_+(v,\alpha)+m_-(v,\alpha)}2
  +\frac{as}{4}\bigl[m_+(v,\alpha)-m_-(v,\alpha)\bigr].
 \label{crex:eq:Bellman-closure}
\end{equation}
Both $1$ and $as$ belong to $\jSpan\varphi_Q$, so this span is
invariant under $K_\alpha$. The Markov operator is a contraction
in the supremum norm. Hence
$\sum_{k\ge0}\gamma^kK_\alpha^kq_{x,\alpha}$ converges uniformly.
Every partial sum lies in the finite-dimensional, and therefore
closed, span of $\varphi_Q$. Its limit is the value in
\eqref{crex:eq:true-value}, which proves C2.

Finally, on the ambient ball,
$|h_n(\alpha_i)|\le A/2$ and
$0\le\log\cosh(h_n(\alpha_i))\le |h_n(\alpha_i)|$.
Thus the absolute log-policy ratio is at most $A$.
The reward bound and the discounted resolvent bound give
\eqref{crex:eq:value-envelope}.
\end{proof}

The Bellman-closure step is the same structural principle used for
value representation in linear MDPs \cite{jin2020linear}.
Here the additional membership of the log-policy ratio in the fixed
feature span is explicitly verified in \eqref{crex:eq:KL-span}.
The unregularized transition has rank-two features
$(w_+,w_-)=(1/2+as/4,1/2-as/4)$ and measure vector $(\mu_+,\mu_-)$.
The five-feature augmentation also represents the actor's KL cost.
The hypotheses of Proposition~\ref{crex:prop:class} specialize to
\[
 \begin{aligned}
 M=M'&=2,\quad
 P=w_+\mu_++w_-\mu_-,\\
 \psi(s,a)&=a e_{i(s)},\quad
 \pi_{\jref}(a\mid i)=1/2,\\
 \|\varphi_Q\|&\le2,\quad
 1,\ as,\ \{\mathbf1_{\{(i(s),a)=(j,b)\}}\}_{j,b}
     \in\jSpan\varphi_Q,\\
 w_\pm,\ r_x&\in\jSpan\varphi_Q,\quad
 \jE_{\rho\pi_{\jref}}\varphi_Q\varphi_Q^{\mathsf T}
   =\operatorname{diag}(1/4,1/4,1/4,1/4,1).
 \end{aligned}
\]

\begin{corollary}[A fixed coefficient envelope]
\label{crex:cor:coefficients}
The coefficient vector $w_{x,\alpha}$ in
$Q^{\mathrm{true}}_{x,\alpha}=\varphi_Q^{\mathsf T}w_{x,\alpha}$
is unique and, on the ambient ball, satisfies
\[
 \|w_{x,\alpha}\|_2\le\frac{2(1+\tau A)}{1-\gamma}.
\]
\end{corollary}
\begin{proof}
At the reference policy, the reset Gram matrix is
$\operatorname{diag}(1/4,1/4,1/4,1/4,1)\succeq I_5/4$.
This proves linear independence of the features. Applying the same
Gram inequality to $Q^{\mathrm{true}}=\varphi_Q^{\mathsf T}w$ yields
$\|w\|_2^2/4\le\jE_{\rho_0}(Q^{\mathrm{true}})^2
\le\|Q^{\mathrm{true}}\|_\infty^2$.
\end{proof}

\subsubsection{An explicit nonconstant lower response}
\label{crex:sec:response}

Use the normalized lower objective~\eqref{eq:main-lower},
\begin{equation}
 g_{\tau,n}(x,\alpha)
 =(1-\gamma)\jE_{n,\alpha}
   \sum_{t=0}^{\infty}\gamma^t q_{x,\alpha}(s_t,a_t).
 \label{crex:eq:lower-objective}
\end{equation}

\begin{proposition}[Selected response and true critic in closed form]
\label{crex:prop:response}
If $\tau>1/2$, then, for every $x\in\jR$ and $n\ge1$,
\eqref{crex:eq:lower-objective} has the unique global minimizer
\begin{equation}
 \alpha_i^*(x)
 =2\sqrt n\,\jatanh\!\left(\frac{i\,u(x)}{2\tau\sqrt n}\right),
 \qquad i\in\{-1,+1\}.
 \label{crex:eq:response}
\end{equation}
Its value and true regularized action value are
\begin{align}
 g_{\tau,n}(x,\alpha^*(x))
 &=-\tau\log\cosh\!\left(\frac{u(x)}{2\tau}\right),
 \label{crex:eq:minimum}\\
 Q^{\mathrm{true}}_{x,\alpha^*(x)}(s,a)
 &=-\frac{u(x)}{\sqrt{12}}\,a z(s)
   -\frac{\tau}{1-\gamma}
       \log\cosh\!\left(\frac{u(x)}{2\tau}\right).
 \label{crex:eq:explicit-critic}
\end{align}
Moreover,
\begin{equation}
 \sup_{n\ge1,x\in\jR}\|\alpha^*(x)\|_2
 \le2\sqrt2\,\jatanh\!\left(\frac1{2\tau}\right),
 \qquad
 \left.\partial_x\alpha_i^*(x)\right|_{x=0}=\frac{i}{\tau}.
 \label{crex:eq:response-bounds}
\end{equation}
\end{proposition}

\begin{proof}
Every state law generated from $\rho$ is a mixture of $\mu_-$ and
$\mu_+$, because this is true initially and every transition in
\eqref{crex:eq:transition} has that form. The normalized discounted
state occupancy therefore has the form
$d_\alpha=d_-(\alpha)\mu_-+d_+(\alpha)\mu_+$, where
$d_-+d_+=1$ and each $d_i\ge(1-\gamma)/2>0$.

Write $h_i=h_n(\alpha_i)$ and
$p_b(a)=e^{ab}/(2\cosh b)$. Under $p_{h_i}$,
$\jE a=\tanh h_i$. Since $\jE_{\mu_i}s=i/2$, the average one-step
cost in region $i$ is
\[
 \ell_i(x,h_i)
 =-\frac{i\,u(x)}2\tanh h_i
  +\tau\bigl[h_i\tanh h_i-\log\cosh h_i\bigr].
\]
Let $b_i=i\,u(x)/(2\tau)$. Direct calculation of binary relative
entropy gives
\[
 \ell_i(x,h_i)
 =\tau\jKL(p_{h_i}\|p_{b_i})
  -\tau\log\cosh\!\left(\frac{u(x)}{2\tau}\right).
\]
Consequently
\begin{equation}
 g_{\tau,n}(x,\alpha)
 =\tau\sum_{i\in\{-1,+1\}}d_i(\alpha)\,
      \jKL(p_{h_i}\|p_{b_i})
  -\tau\log\cosh\!\left(\frac{u(x)}{2\tau}\right).
 \label{crex:eq:lower-KL-decomposition}
\end{equation}
The dependence of $d_i$ on the actor is retained in this identity.
Both weights are strictly positive. Thus equality with the lower
bound occurs exactly when $h_i=b_i$ for both regions.
Since $h_n$ is strictly increasing with range
$(-\sqrt n,\sqrt n)$ and $|b_i|<1\le\sqrt n$ when $\tau>1/2$,
these equalities have the unique solution
\eqref{crex:eq:response}. This also proves \eqref{crex:eq:minimum}.

At this response, \eqref{crex:eq:reward-span} and
\eqref{crex:eq:KL-span} give
\[
 q_{x,\alpha^*}(s,a)
 =-\frac{u(x)}{\sqrt{12}}a z(s)
  -\tau\log\cosh\!\left(\frac{u(x)}{2\tau}\right).
\]
The function $a z(s)$ has zero mean under each
$\mu_i(ds)\pi_{n,\alpha^*}(da\mid i)$. Hence
$K_{\alpha^*}(a z)=0$ and $K_{\alpha^*}1=1$.
The discounted resolvent therefore gives
\eqref{crex:eq:explicit-critic} directly.

For $0\le t<1$, the power series for inverse hyperbolic tangent gives
\[
 \sqrt n\,\jatanh(t/\sqrt n)
 =\sum_{k=0}^{\infty}\frac{t^{2k+1}}{(2k+1)n^k}
 \le\jatanh t.
\]
Apply this with $t=|u(x)|/(2\tau)$ to obtain the norm bound.
Differentiating \eqref{crex:eq:response} gives
\[
 \partial_x\alpha_i^*(x)
 =\frac{i\,\jsech^2 x}
        {\tau\bigl(1-u(x)^2/(4\tau^2n)\bigr)},
\]
which proves the final claim.
\end{proof}

For any fixed target radius $R_0>0$, the bound
\eqref{crex:eq:response-bounds} is smaller than $R_0$ whenever
\[
 \tau>\frac1{2\tanh(R_0/(2\sqrt2))}.
\]
Thus $\tau$ can be chosen large enough to meet our
geometry and initialization requirements while placing the entire
response family inside any prescribed smaller actor buffer.
This same $\tau$ is then fixed before selecting critic constants.
For $x\ne0$ the response gives different policies in the two regions,
and \eqref{crex:eq:explicit-critic} varies with the continuous state
within each region.

\subsubsection{An interior upper stationary point and a nonzero chain term}
\label{crex:sec:upper}

Let $f_n(x,\alpha)=\jE L(x;\boldsymbol\xi,Y)$ and
$\Phi_\tau(x)=f_n(x,\alpha^*(x))$. The conditional mean sample loss is
\begin{equation}
 \ell_\kappa(u,d)
 =\log(1+e^{ud})-ud\,\sigma(\kappa d)
 =\log(2\cosh(ud/2))-\frac{ud}{2}\tanh(\kappa d/2).
 \label{crex:eq:mean-loss}
\end{equation}
It is even in $d$. For arbitrary actor coordinates, the exact score is
\[
 \partial_{\alpha_j}\log\pi_{n,\alpha}(a\mid s)
 =\mathbf1_{\{i(s)=j\}}h_n'(\alpha_j)
      [a-\tanh(h_n(\alpha_j))].
\]
Since the dynamics and conditional feedback law are fixed,
\begin{equation}
 f_\alpha
 =\jE\left[
  L\sum_{\ell=1}^{2}\sum_{t=0}^{1}
    \nabla_\alpha\log\pi_{n,\alpha}(a_t^\ell\mid s_t^\ell)
 \right].
 \label{crex:eq:score-identity}
\end{equation}

\begin{lemma}[An exact one-dimensional quadrature formula]
\label{crex:lem:quadrature}
At the lower response put $u=\tanh x$ and
$p=\sigma(u/\tau)$. Define
\begin{align}
 B_4(y)&=\frac16\sum_{j=0}^{4}(-1)^j\binom4j(y-j)_+^3,
 &&0\le y\le4,\notag\\
 J_k(u)&=\int_0^4 B_4(y)\,
       \ell_\kappa(u,y-4+k)\,dy,
 &&k=0,\ldots,4. \label{crex:eq:J}
\end{align}
Here $B_4$ is the density of the sum of four independent uniform
variables on $[0,1]$. Write $v=p(1-p)$ and
\begin{equation}
 (\omega_0,\omega_1,\omega_2,\omega_3,\omega_4)
 =(v^2,\ 2v-4v^2,\ 1-4v+6v^2,\ 2v-4v^2,\ v^2).
 \label{crex:eq:mixture-weights}
\end{equation}
Then the exact population objective and gradient decomposition are
\begin{align}
 \Phi_\tau(x)&=\sum_{k=0}^{4}\omega_k(p)J_k(u),\notag\\
 \mathsf A_\tau(x):=f_x(x,\alpha^*(x))
 &=(1-u^2)\sum_{k=0}^{4}\omega_k(p)J_k'(u),\notag\\
 \mathsf C_\tau(x):=D_x\alpha^*(x)^{\mathsf T}
                         f_\alpha(x,\alpha^*(x))
 &=\frac{(1-u^2)p(1-p)}{\tau}
     \sum_{k=0}^{4}\omega_k'(p)J_k(u),\notag\\
 \Phi_\tau'(x)&=\mathsf A_\tau(x)+\mathsf C_\tau(x).
 \label{crex:eq:upper-derivative}
\end{align}
In the derivative of $J_k$, the integrand is
$[\sigma(ud)-\sigma(\kappa d)]d$, with $d=y-4+k$.
\end{lemma}
\begin{proof}
At the response the policy logits are
$h_n(\alpha_i^*)=iu/(2\tau)$. Write each state as $s_t=i_tR_t$,
where $R_t=|s_t|$, and define $\epsilon_t=a_ti_t$.
At reset, $R_0$ is uniform on $[0,1]$ and independent of $i_0$.
Every subsequent transition first selects a region and then draws
a fresh uniform magnitude independent of the past. Conditional on
every region and magnitude, the response policy has
$\mathbb P(\epsilon_t=+1)=p$, independent of that region, magnitude,
and the past. Thus the products $a_ts_t=\epsilon_tR_t$ are
independent across time, with independent uniform $R_t$ and
independent signs of probability $p$.

This independence concerns the products $a_ts_t$, not the raw
states conditioned on an entire future region path. It accounts
for the dependence of the transition weights on the preceding
state and action.
For two independent length-two trajectories,
\[
 D=\epsilon_1R_1+\epsilon_2R_2-\epsilon_3R_3-\epsilon_4R_4.
\]
Conditional on the four signs in this sum, if $k$ signs are
positive then $D$ has the law of a sum of four independent
$[0,1]$ uniforms minus $4-k$. Replacing a negatively signed
uniform $-R$ by $(1-R)-1$ proves this claim.
The probability generating polynomial of $k$ is
$[(1-p)+pz]^2[p+(1-p)z]^2$; its coefficients are exactly
\eqref{crex:eq:mixture-weights}. This proves the objective formula.
Differentiating $u(x)$ and $p(x)$ gives
$u'=1-u^2$ and $p'=(1-u^2)p(1-p)/\tau$.
Differentiation of the integrals is justified on their compact
domains. The derivative through $u$ is the direct partial derivative,
and the derivative through $p$ is exactly the trajectory-law
contribution in \eqref{crex:eq:score-identity}.
\end{proof}

\begin{proposition}[A finite interior minimizer with a nonzero correction]
\label{crex:prop:interior}
For every $\kappa\in(0,1)$ and $\tau>1/2$, let
$x^\dagger=\jatanh\kappa$. Then
\[
 \Phi_\tau'(0)<0,\qquad
 \Phi_\tau'(x^\dagger)>0.
\]
Consequently $\Phi_\tau$ has an interior minimizer on
$[0,x^\dagger]$. At every stationary point
$\bar x\in(0,x^\dagger)$,
\begin{equation}
 \mathsf A_\tau(\bar x)<0,\qquad
 \mathsf C_\tau(\bar x)=-\mathsf A_\tau(\bar x)>0,\qquad
 \frac{|\mathsf C_\tau(\bar x)|}{|\mathsf A_\tau(\bar x)|}=1.
 \label{crex:eq:nonzero-chain}
\end{equation}
The chain term vanishes at $x=0$.
\end{proposition}
\begin{proof}
At $x=0$, the loss equals $\log2$ for every tuple and every label,
so $f_\alpha(0,\alpha)=0$ and $\mathsf C_\tau(0)=0$. The direct
derivative equals
$\jE[(1/2-\sigma(\kappa D))D]$, which is strictly negative:
the integrand is negative whenever $D\ne0$, an event of
probability one here.

At $x=x^\dagger$, $u=\kappa$ and the direct derivative is zero.
The conditional mean loss is the binary entropy
$H(\sigma(\kappa d))$, an even function strictly decreasing in
$|d|>0$. To compare the integrals in \eqref{crex:eq:J},
let $X$ be the sum of four independent uniforms minus two.
Its density $b$ is symmetric and decreasing in $|X|$.
For example, $B_4(y)=y^3/6$ on $[0,1]$ and
$B_4(y)=[y^3-4(y-1)^3]/6$ on $[1,2]$ is increasing there,
and symmetry supplies the other half of the density.
For any $c>0$, differentiating the convolution with the entropy
function $H_\kappa$ gives
\[
 \frac{d}{dc}\jE H_\kappa(X+c)
 =\int_0^\infty H_\kappa'(z)
          [b(z-c)-b(z+c)]\,dz<0.
\]
Indeed $H_\kappa'(z)<0$ for $z>0$ and the bracket is nonnegative,
and strictly positive on a set of positive measure.
It follows that $J_2(\kappa)>J_1(\kappa)>J_0(\kappa)$ and
$J_{4-k}(\kappa)=J_k(\kappa)$.

Holding $u=\kappa$ fixed, write the objective as a polynomial in
$v=p(1-p)$. Its derivative with respect to $v$ is
\[
 4v[J_0-J_2]+(4-16v)[J_1-J_2]<0.
\]
Here $0<v<1/4$, $p>1/2$, and $dv/dp=1-2p<0$.
Since $p'(x^\dagger)>0$, the response contribution is strictly
positive. This proves the second endpoint sign.

A minimum on the compact interval exists, and the strict endpoint
derivative signs exclude either endpoint. At every interior
stationary point, $0<u<\kappa$, so
$[\sigma(uD)-\sigma(\kappa D)]D<0$ almost surely.
Thus the direct term is strictly negative and stationarity forces
the positive, nonzero cancellation in \eqref{crex:eq:nonzero-chain}.
\end{proof}

The existence proof holds for every $\tau>1/2$, including any
larger fixed $\tau$ selected by our geometry and
containment constants. The threshold $1/2$ is sufficient for the
explicit response formula; it does not replace that operative
algorithmic choice of $\tau$.

\begin{remark}[Preference-risk stationarity and the teacher objective]
\label{crex:rem:preference-risk}
Let $p^\dagger=\sigma(\kappa D)$, $p_x=\sigma(\tanh(x)D)$,
and let $H_{\rm B}$ denote binary entropy. Conditional cross-entropy gives
\begin{equation}
 \Phi_\tau(x)=\jE_{\boldsymbol\xi\sim(\mathbb P^2_{\alpha^*(x)})^{\otimes2}}
 \left[H_{\rm B}(p^\dagger)+
 \jKL\bigl(\operatorname{Ber}(p^\dagger)\|\operatorname{Ber}(p_x)\bigr)\right].
 \label{crex:eq:risk-decomposition}
\end{equation}
At $x^\dagger=\jatanh\kappa$, the divergence and its derivative vanish,
but the expected entropy can vary through the response-dependent
trajectory law. Proposition~\ref{crex:prop:interior} gives
$\Phi_\tau'(x^\dagger)>0$ and an interior minimizer
$\bar x\in(0,x^\dagger)$.

The normalized return under the teacher reward $r^\dagger(s,a)=\kappa as$
is, at the selected response,
\begin{equation}
 J^\dagger(x):=(1-\gamma)\jE_{\alpha^*(x)}
       \sum_{t\ge0}\gamma^t r^\dagger(s_t,a_t)
 =\frac{\kappa}{2}\tanh\!\left(\frac{\tanh x}{2\tau}\right).
 \label{crex:eq:teacher-return}
\end{equation}
Indeed, conditional on either region, the state is uniform and
$\jE[a\mid i]=i\tanh(\tanh(x)/(2\tau))$, so every time step has
the displayed mean reward. This return is strictly increasing in $x$;
thus $J^\dagger(\bar x)<J^\dagger(x^\dagger)$.
The example establishes a nonzero response correction and also
separates optimization of the stated on-policy risk from reward
identification or teacher-return maximization.
\end{remark}

\paragraph{A fully specified population illustration.}
Take $\gamma=1/2$, $\kappa=1/2$, $n=1$, and $\tau=2$.
Deterministic quadrature of \eqref{crex:eq:J} gives an interior
minimizer
\begin{equation}
 \bar x\approx0.5414613800,\qquad
 x^\dagger\approx0.5493061443,\qquad
 \Phi_2(\bar x)\approx0.6559732584.
 \label{crex:eq:numerical-root}
\end{equation}
At $\bar x$, the direct and response terms are approximately
$-0.001204848558$ and $+0.001204848558$, respectively.
Table~\ref{crex:tab:gradient} reports both absolute magnitudes and
their ratio. The ratio equals one at a stationary point by
cancellation; this identity alone should not be interpreted as a
large correction throughout the domain.

\begin{table}[tb]
\centering\small
\begin{tabular}{rrrrr}
\toprule
$x$ & $\mathsf A_2(x)$ & $\mathsf C_2(x)$ &
$\Phi_2'(x)$ & $|\mathsf C_2|/|\mathsf A_2|$\\
\midrule
$0$        & $-0.15561048$ & $0$          & $-0.15561048$ & $0$\\
$0.25$     & $-0.07068856$ & $0.00057120$ & $-0.07011736$ & $0.008081$\\
$0.50$     & $-0.00814360$ & $0.00116961$ & $-0.00697399$ & $0.143623$\\
$\bar x$   & $-0.00120485$ & $0.00120485$ & $\approx0$   & $1$\\
$x^\dagger$& $0$           & $0.00120937$ & $0.00120937$ & ---\\
$0.75$     & $0.02048602$  & $0.00113742$ & $0.02162344$ & $0.055522$\\
$1.00$     & $0.02633981$  & $0.00079848$ & $0.02713829$ & $0.030315$\\
\bottomrule
\end{tabular}
\caption{Population gradient decomposition for $\kappa=1/2,\tau=2$.
The ratio at $x^\dagger$ is undefined because its denominator is zero.}
\label{crex:tab:gradient}
\end{table}

\begin{figure}[tb]
\centering
\includegraphics[width=\linewidth]{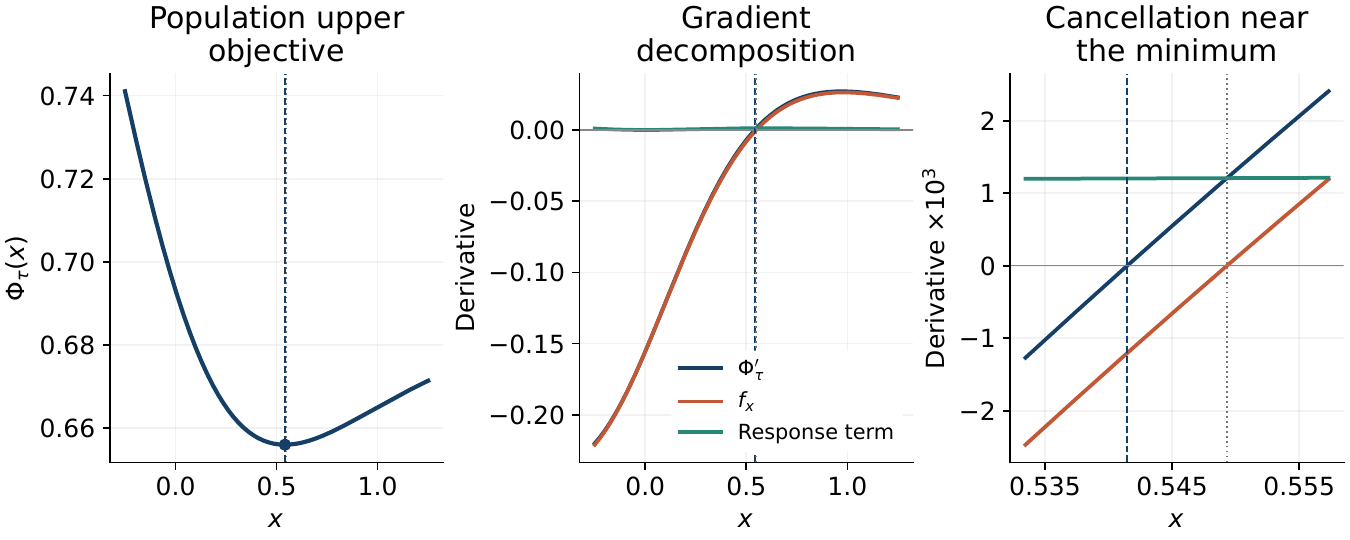}
\caption{Population quantities evaluated by deterministic quadrature.
Left: upper objective. Middle: full, direct, and response derivatives.
Right: the same derivatives near the minimum, with the vertical scale
multiplied by $10^3$. The dashed line is $\bar x$ and the dotted line
is $x^\dagger$. These curves illustrate the response mechanism;
they are not stochastic convergence results.}
\label{crex:fig:upper}
\end{figure}

The density $B_4$ is polynomial between consecutive integers.
Gauss--Legendre quadrature with 12, 24, and 48 nodes on each of its
four pieces agreed at the checked points to less than $10^{-13}$
in the objective and both gradient terms. An independent
four-dimensional tensor quadrature with explicit enumeration of
the 16 sign patterns agreed at $\bar x$ to the same tolerance.
These are numerical consistency checks, rather than formal
interval-error certificates; the existence and nonzero-chain
claims are established analytically above.

For larger fixed regularization the correction remains nonzero,
but becomes smaller, as Table~\ref{crex:tab:tau} illustrates.
In this symmetric example $\mathsf C_\tau=O(\tau^{-2})$ on compact
$x$ intervals: $p-1/2=O(\tau^{-1})$ and
$\omega_k'(p)=O(p-1/2)$ in
\eqref{crex:eq:upper-derivative}.
The $\tau=2$ plot is a population illustration, not a certification
of every conservative constant in the stochastic containment theorem.

\begin{table}[tb]
\centering\small
\begin{tabular}{rrr}
\toprule
$\tau$ & Interior minimizer $\bar x$ & $\mathsf C_\tau(\bar x)$\\
\midrule
$2$   & $0.5414613800$ & $1.204848558\times10^{-3}$\\
$4$   & $0.5473047717$ & $3.064773816\times10^{-4}$\\
$10$  & $0.5489840685$ & $4.927812863\times10^{-5}$\\
$100$ & $0.5493029200$ & $4.932401058\times10^{-7}$\\
\bottomrule
\end{tabular}
\caption{The finite stationary point and its nonzero correction
persist as $\tau$ increases, with $\kappa=1/2$.}
\label{crex:tab:tau}
\end{table}

\ifdefined\FloatBarrier\FloatBarrier\fi
\subsubsection{Joint conclusion and exact critic targets}
\label{crex:sec:joint}

\begin{proposition}[Joint continuous-state instance]
\label{crex:prop:joint}
Fix an ambient actor radius $A>0$ and $\gamma\in(0,1)$.
Use \eqref{crex:eq:transition}--\eqref{crex:eq:loss}, the actor
\eqref{crex:eq:score}--\eqref{crex:eq:policy}, and the feature map
\eqref{crex:eq:features}. Choose sufficiently large finite $\tau$
as required by our actor geometry and containment
construction, and then keep it fixed. The resulting instance
satisfies A1, A2, G, C1, and C2 simultaneously, uniformly over actor
widths $2n$ and over the prescribed actor neighborhoods contained in
$\|\alpha\|_2\le A$. The reward and upper domains are unrestricted in
$x\in\jR$. The actor has retained dimension $2$, the critic has fixed
feature dimension $5$, and the constants in
\eqref{crex:eq:actor-gap}, \eqref{crex:eq:global-constants}, and
\eqref{crex:eq:critic-constants} do not depend on critic scale or
target accuracy.
\end{proposition}
\begin{proof}
Combine Lemmas~\ref{crex:lem:architecture},
\ref{crex:lem:regularity}, \ref{crex:lem:C1}, and
\ref{crex:lem:C2}. The critic verifications hold on the full ambient
ball, so they also cover base and perturbed actors and the local
update segments used by the theorem. Proposition~\ref{crex:prop:response}
shows that the selected response can be placed strictly inside the
chosen geometry neighborhood. No assumption about an independently
accurate critic initialization is used in these verifications.
\end{proof}

\begin{remark}[Scope and the nonlinear trained critic]
\label{crex:rem:scope}
The example has a finite-rank transition kernel and a policy that uses
a two-region state encoding. State differentiability is not required
by A1; the measurable encoding $i(s)$ is therefore allowed. The
transition probabilities and rewards retain within-region continuous
state dependence. This establishes joint satisfiability within the
stated model class.

C2 is a statement about the fixed linear span
$\jSpan\varphi_Q$. We train
\[
 Q_{m_Q,\theta}(s,a)
 =\sqrt{m_Q}\tanh\!\left(
      \frac{\varphi_Q(s,a)^{\mathsf T}\theta}{\sqrt{m_Q}}
    \right),
\]
whose finite nonlinear scale need not yield exact representation of
the true value. For example, on $\|\theta\|_2\le R_Q$ the feature bound
above gives the elementary estimate
\[
 \bigl\|Q_{m_Q,\theta}-\varphi_Q^{\mathsf T}\theta\bigr\|_\infty
 \le\frac{8R_Q^3}{3m_Q}.
\]
Our paired-TD stability and tracking arguments are still
used to obtain its derived squared paired approximation floor and
interaction bound. Exact feature-span membership is the primitive
realizability condition verified by this example.
\end{remark}

\paragraph{Exact value coefficients for arbitrary actors.}
The one-dimensional response formula is not needed to compute the
true value away from the response. Let the cell order be that of
\eqref{crex:eq:features}. For $j\in\{-1,+1\}$ define coefficient
vectors $v_j,m_j(\alpha)\in\jR^5$ by
\[
 (v_j)_{(i,a)}=\frac12+\frac{jai}{8},\qquad
 (v_j)_5=\frac{j}{4\sqrt{12}},\qquad
 (m_j)_{(i,a)}=\mathbf1_{\{i=j\}}\pi_{n,\alpha}(a\mid i),
 \qquad (m_j)_5=0.
\]
Then $w_j(s,a)=\varphi_Q(s,a)^{\mathsf T}v_j$ and
$\jE_{\mu_j\pi_\alpha}\varphi_Q=m_j(\alpha)$.
Define
\begin{equation}
 T_\alpha=\sum_{j\in\{-1,+1\}}v_j m_j(\alpha)^{\mathsf T},
 \qquad
 K_\alpha(\varphi_Q^{\mathsf T}b)
   =\varphi_Q^{\mathsf T}T_\alpha b.
 \label{crex:eq:five-matrix}
\end{equation}
If $c_{x,\alpha}$ is the coefficient vector for $q$ in the proof of
Lemma~\ref{crex:lem:C2}, the true-value coefficients are exactly
\begin{equation}
 w_{x,\alpha}
 =(I_5-\gamma T_\alpha)^{-1}c_{x,\alpha}.
 \label{crex:eq:five-solve}
\end{equation}
The inverse exists because the discounted Bellman operator is
invertible on the invariant feature span and the features are
linearly independent.

\paragraph{The finite nonlinear TD target is a different root.}
For a base actor $U$, the state occupancy weights can themselves
be computed from a two-by-two matrix. Its rows are
\[
 \overline P_U(i,+)=\frac12+\frac{i\tanh(h_n(U_i))}{8},\qquad
 \overline P_U(i,-)=1-\overline P_U(i,+).
\]
With row-vector convention,
\[
 \begin{aligned}
 d_U&=(1-\gamma)(1/2,1/2)(I_2-\gamma\overline P_U)^{-1},\\
 \nu_U(ds,da)&=\sum_i d_U(i)\mu_i(ds)\pi_{n,U}(a\mid i).
 \end{aligned}
\]
For finite critic scale, the relevant mean field remains
\begin{equation}
 \mathsf H_{m_Q}(x,U,\alpha,\theta)
 =\jE_{\nu_U}\left[
  (Q_{m_Q,\theta}-q_{x,\alpha}
      -\gamma K_\alpha Q_{m_Q,\theta})
      \nabla_\theta Q_{m_Q,\theta}
 \right].
 \label{crex:eq:nonlinear-root}
\end{equation}
All integrals in this field reduce to finite action sums and
one-dimensional within-region integrals. Our interior
root $\theta^*_{m_Q}(x,U,\alpha)$ can therefore be computed by a
five-dimensional nonlinear root solve with controlled quadrature.
For $p_{\mathrm{alg}}=(x,U,S)$ the paired target is
\begin{equation}
 Z_{m_Q}(p_{\mathrm{alg}})
 =\left(
 \theta^*_{m_Q}(x,U,U),\
 \frac{\theta^*_{m_Q}(x,U,U+cS)
       -\theta^*_{m_Q}(x,U,U)}{c}
 \right).
 \label{crex:eq:paired-root}
\end{equation}
The linear solve \eqref{crex:eq:five-solve} computes the exact
value coefficients, not this nonlinear TD root. Monitoring the
theorem's $\|z_t-Z_{m_Q}(p_t)\|_W$ therefore requires the nonlinear
solves, with tolerances small enough after division by $c$.

\paragraph{Population evaluation away from the lower response.}
The local perturbed response solves
$g_\alpha(x,\alpha_c^*)+c f_\alpha(x,\alpha_c^*)=0$, with
$S_c^*=(\alpha_c^*-\alpha_0^*)/c$.
Away from the unperturbed response, preference expectations must
retain the transition likelihoods; the independent-product reduction
in \cref{crex:lem:quadrature} is specific to the response policy.
The exact population calculations provide evaluation quantities and
do not replace the sampling estimators in Algorithm~\ref{alg:main-trained}.

\subsection{Conditional global lower responses on a prescribed domain}
\label{app:gpl-global-response}
This subsection proves Corollary~\ref{cor:gpl-global-response} and
specifies the scope of its additional assumption. Fix the finite actor
architecture $\mathbf m$ and the same fixed regularization parameter
$\tau$ as in Theorem~\ref{thm:main}. Write
$a(x)=\alpha^*_{0,\mathbf m}(x)$ and
\[
 g^*_{\tau,\mathbf m,\mathcal D}(x)
 :=\min_{\beta\in\mathcal D}g_{\tau,\mathbf m}(x,\beta),\qquad
 \mathcal A^*_{\tau,\mathbf m,\mathcal D}(x)
 :=\argmin_{\beta\in\mathcal D}g_{\tau,\mathbf m}(x,\beta).
\]
The minimum exists by continuity and compactness. The actor map remains
defined on $\R^r$; $\mathcal D$ specifies the comparison domain for
global lower optimality.

\begin{proof}[Proof of Corollary~\ref{cor:gpl-global-response}]
Proposition~\ref{glob:geometry}, with $c=0$, gives
$a(x)\in B_{R_{\rm geom}/4}\subset\mathcal D$ and
$\nabla_\alpha g_{\tau,\mathbf m}(x,a(x))=0$ for every $x$.
Assumption~\ref{ass:gpl-domain} therefore yields
\begin{equation}
 0\le g_{\tau,\mathbf m}(x,a(x))-g^*_{\tau,\mathbf m,\mathcal D}(x)
 \le\frac{\|\nabla_\alpha g_{\tau,\mathbf m}(x,a(x))\|^2}
               {2\mu_{\rm PL}}=0.
 \label{eq:gpl-identification-proof}
\end{equation}
Thus $a(x)\in\mathcal A^*_{\tau,\mathbf m,\mathcal D}(x)$.
The selected global-response objective
$\Phi^{\rm sel}_{\tau,\mathbf m,\mathcal D}(x)=f_{\mathbf m}(x,a(x))$
equals $\Phi_{\tau,\mathbf m}(x)$ everywhere. Smoothness follows from
the existing response construction, and their gradients agree.
Theorem~\ref{thm:main}, applied to the same algorithm and output,
therefore gives expected squared gradient norm $O(\epsilon)$ with
$\widetilde O(\epsilon^{-2})$ segments and transitions, and its
unchanged label and numerical KL-solve costs.
\end{proof}

\paragraph{Domain, constants, and algorithmic scope.}
The prescribed $\mathcal D$ is independent of $x$, perturbation $c$,
critic scale $m_Q$, and accuracy $\epsilon$. A meaningful extension
compares against actors beyond the neighborhood already covered by
derived strong convexity. The additional PL condition uses the true
fixed-$\tau$ lower objective and its ambient Euclidean gradient,
including at boundary points of $\mathcal D$. It is not a
projected-gradient condition and need not certify arbitrary constrained
stationary points. The proof evaluates it at the interior, zero-gradient
response. Compactness and inclusion of that response suffice for the
identification argument; convexity is a convenient domain convention.

No projection onto $\mathcal D$ or new global lower solver is introduced.
All initializations, local containment and failure-output rules, sampling
blocks, parameter choices, and finite KL solves remain those of
Theorem~\ref{thm:main}. In particular, the argument imposes no PL condition
on the perturbed objective $h_c=g_{\tau,\mathbf m}+cf_{\mathbf m}$ and
requires no critic realizability or training beyond the neighborhoods in
C1--C2. Those conditions continue to justify the estimator locally.
The extra assumption changes the interpretation of the selected response,
not the algorithm or its distribution of outputs.

For a fixed architecture, $\mu_{\rm PL}$ may depend on $\mathbf m$,
$\mathcal D$, and $\tau$. It enters no step-size or sample-size choice
in this corollary, so no new inverse-PL factor is introduced into the
inherited interaction bound. Positivity separately at each $x$ would
already suffice in \eqref{eq:gpl-identification-proof}; uniformity in $x$
is the stated stronger condition. If widths are instead selected as
$\mathbf m=\mathbf m(\epsilon)$, impose the condition for every
architecture used on an accuracy-independent domain and retain the target
notation $\Phi^{\rm sel}_{\tau,\mathbf m(\epsilon),\mathcal D}$.
The argument does not identify targets at different widths, and does
not need a width-uniform PL constant. The pointwise certificate
\[
 g_{\tau,\mathbf m}(x,\alpha)-g^*_{\tau,\mathbf m,\mathcal D}(x)
 \le (2\mu_{\rm PL})^{-1}
          \|\nabla_\alpha g_{\tau,\mathbf m}(x,\alpha)\|^2,
 \qquad \alpha\in\mathcal D,
\]
would require a separate residual estimate to give a finite-sample
lower-suboptimality bound for an approximate actor output.

\paragraph{Selection independence and optimization target.}
PL alone does not generally guarantee uniqueness of a lower minimizer.
The corollary retains the specified smooth selection $a(x)$. If, in
addition, $\mathcal A^*_{\tau,\mathbf m,\mathcal D}(x)$ is a singleton
for every $x$, that selection is the unique global lower response.
A weaker sufficient condition for a selection-independent upper objective
is upper-loss invariance:
\begin{equation}
 f_{\mathbf m}(x,\alpha)=f_{\mathbf m}(x,\beta)
 \quad\text{for all }\alpha,\beta\in
       \mathcal A^*_{\tau,\mathbf m,\mathcal D}(x),\quad\text{for every }x.
 \label{eq:gpl-upper-invariance}
\end{equation}
Then every choice of global lower solution induces the same scalar
objective, which is smooth because it equals $\Phi_{\tau,\mathbf m}$,
even if some selections themselves are nonsmooth. Without either
condition or another selection argument, the corollary does not identify
its objective with the optimistic or pessimistic upper objective obtained
by minimizing or maximizing $f_{\mathbf m}$ over the lower solution set.

The conclusion is upper stationarity for a selected global
\emph{regularized} lower response within the declared actor class and
domain. Matching another bilevel target additionally requires matching
the policy class, domain, regularization, and selection convention.
It gives no unregularized-response comparison or guarantee from arbitrary
actor initialization. In particular, $\tau$ stays at the fixed value
required for the original geometry; the analysis does not permit
$\tau\downarrow0$. PL only for the unregularized lower objective would
not justify \eqref{eq:gpl-identification-proof}.

\paragraph{Why a full-space PL assumption is unavailable here.}
The bounded activations and fixed readout in
\eqref{eq:branch-energy} and \eqref{eq:activation-envelope} imply, at
every branch parameter,
$|G_{\mathbf m,\vartheta}(s,a)|\le\sqrt{m_L}M_{L,0}$.
Consequently the antithetic energy \eqref{eq:deep-energy} obeys
\begin{equation}
 |F_{\mathbf m,\alpha}(s,a)|\le B_{F,\mathbf m}
 :=\sqrt{2m_L}M_{L,0}<\infty
 \qquad(\alpha\in\R^r).
 \label{eq:gpl-energy-bound}
\end{equation}
For the exponential tilt \eqref{eq:deep-policy},
\eqref{eq:log-partition} gives
$|A_{\mathbf m,s}(\alpha)|\le B_{F,\mathbf m}$, and hence
$|\log(\pi_{\mathbf m,\alpha}/\piRef)|\le2B_{F,\mathbf m}$.
The expected log ratio at each state is a nonnegative KL divergence.
The reward bound and normalized discounting in \eqref{eq:main-lower}
therefore yield
\begin{equation}
 -B_r\le g_{\tau,\mathbf m}(x,\alpha)
 \le B_r+2\tau B_{F,\mathbf m}\qquad\text{for all }x,\alpha.
 \label{eq:gpl-lower-bound}
\end{equation}
This bound needs fixed finite widths, but no uniform positive lower bound
on reference action probabilities across states. For fixed $x$, the
objective is $C^2$ on $\R^r$: bounded inputs, bounded activation
derivatives, and finite layer weights give uniform actor and score
derivative bounds on each compact parameter set. Differentiated
length-$t$ trajectory likelihoods have polynomial-in-$t$ bounds, which
are summable with $\gamma^t$, justifying twice differentiating the
discounted expectation locally.

\begin{lemma}[Bounded objectives obstruct full-space uniform PL]
\label{lem:gpl-bounded-obstruction}
Let $v:\R^r\to\R$ be $C^2$ and bounded above, with
$v_*:=\inf_{z\in\R^r}v(z)>-\infty$. If there is $\mu>0$ such that
$\tfrac12\|\nabla v(z)\|^2\ge\mu[v(z)-v_*]$ for every $z\in\R^r$,
then $v$ is constant.
\end{lemma}
\begin{proof}
If $v$ is nonconstant, choose $z_0$ with $v(z_0)>v_*$ and consider the
maximal solution of $\dot z=\nabla v(z)$ from $z_0$. Local existence
and uniqueness follow from $C^2$ regularity. With $M=\sup v<\infty$,
\[
 \int_0^t\|\nabla v(z(u))\|^2\,du
 =v(z(t))-v(z_0)\le M-v(z_0),\qquad
 \|z(t)-z(s)\|\le\sqrt{t-s}\sqrt{M-v(z_0)}
\]
for $0\le s<t$ in the solution interval. If the maximal forward time
were finite, the second estimate would give a finite limit for $z(t)$
there; local existence at that limit would extend the solution.
Thus it exists for all $t\ge0$. But
$e(t)=v(z(t))-v_*>0$ satisfies
$e'(t)=\|\nabla v(z(t))\|^2\ge2\mu e(t)$, so
$e(t)\ge e(0)e^{2\mu t}$. This contradicts $e(t)\le M-v_*$.
\end{proof}

Apply Lemma~\ref{lem:gpl-bounded-obstruction} to
$v(\alpha)=g_{\tau,\mathbf m}(x,\alpha)$ at any fixed $x$.
Equation~\eqref{eq:gpl-lower-bound} supplies both bounds, while
Proposition~\ref{glob:geometry}'s positive local Hessian makes $v$
nonconstant for $r\ge1$. A positive full-space uniform Euclidean PL
constant is therefore impossible for this construction. Replacing
$\mathcal D$ by $\R^r$ in Assumption~\ref{ass:gpl-domain} would make
the conditional extension vacuous in these nontrivial instances.

\paragraph{Compatibility on compact domains.}
The obstruction does not apply when ascent can leave the comparison
domain. For illustration, $v(a)=\tanh^2(a)$ satisfies on any
$[-A,A]$, $0<A<\infty$,
\begin{equation}
 \tfrac12|v'(a)|^2
 =2\operatorname{sech}^4(a)v(a)
 \ge2\operatorname{sech}^4(A)
       [v(a)-\min_{b\in[-A,A]}v(b)].
 \label{eq:gpl-tanh-example}
\end{equation}
It is smooth, globally bounded, and nonconstant, with positive curvature
near its minimum and a positive PL constant on each fixed compact
interval. This illustrates compatibility with bounded nonlinearities;
it does not verify Assumption~\ref{ass:gpl-domain} for the RL examples.

\paragraph{A separate condition for upper optimality.}
Set $V=\Phi^{\rm sel}_{\tau,\mathbf m,\mathcal D}$. If a separately
specified upper domain $\mathcal X$ contains $x_{\rm out}$ almost surely
and, for $V_*:=\inf_{x\in\mathcal X}V(x)>-\infty$, one additionally has
\[
 \tfrac12\|\nabla V(x)\|^2\ge\mu_{\rm up}[V(x)-V_*]
 \quad(x\in\mathcal X),\qquad \mu_{\rm up}>0,
\]
then taking expectations gives
$\E[V(x_{\rm out})-V_*]=O(\epsilon/\mu_{\rm up})$.
Neither the upper PL condition nor output membership in such a domain
is asserted here. They do not follow from lower PL. In fact,
Corollary~\ref{glob:Phi} gives a bounded $C^2$ upper objective, so
Lemma~\ref{lem:gpl-bounded-obstruction} also precludes positive
full-space upper PL when that objective is nonconstant.

\FloatBarrier

\providecommand{\ContinuousAssets}{empirical_assets/continuous}
\section{Continuous-state experimental details}
\label{app:continuous-validation}
This section specifies the sampled continuous-state experiment and its
population diagnostics. The structural construction and its C1--C2
verification appear in Appendix~\ref{crex:example}, especially
Lemmas~\ref{crex:lem:C1} and~\ref{crex:lem:C2}. We use the two-hidden-layer
actor in Remark~\ref{crex:rem:depth-two}.  The code was ran on a 12700k processor with an RTX 5090 and 32GB RAM.

\subsection{Instance, parameterization, and reference calculations}
\label{app:continuous-instance}
Let $I_-=[-1,0)$, $I_+=[0,1]$, and $\mu_i=\operatorname{Unif}(I_i)$.
The reset distribution is uniform on $[-1,1]$, and the transition law is
\[
 P(ds'\mid s,a)=\left(\tfrac12+\tfrac{as}{4}\right)\mu_+(ds')
             +\left(\tfrac12-\tfrac{as}{4}\right)\mu_-(ds'),
 \qquad a\in\{-1,+1\}.
\]
We use $\gamma=1/2$, reward $r_x(s,a)=\tanh(x)as$, and reference-KL
coefficient $\tau=2$ with a uniform reference policy. The lower objective
is the normalized discounted cost with
$q_{x,\alpha}=-r_x+\tau\log(\pi_\alpha/\pi_{\rm ref})$.
Two independent length-two state--action trajectories define
$D=\sum_{t=0}^1(a_t^1s_t^1-a_t^2s_t^2)$; the fixed feedback kernel is
$Y\mid D\sim\operatorname{Bernoulli}(\sigma(D/2))$. The preference loss is
\[
 L(x;D,Y)=\log(1+\exp(\tanh(x)D))-Y\tanh(x)D.
\]
The upper objective $\Phi(x)=f(x,\alpha_0^*(x))$ averages this loss over
trajectories generated by the lower response. There is no upper projection
or added quadratic upper regularizer.

The tied actor has two hidden layers of width eight and two trainable chart
coordinates. Its logits are $aH_4(\alpha_{i(s)})$, where
\[
 H_4(v)=\tanh((1+t)\tanh t)+\tanh((1-t)\tanh t),
 \qquad t=v/(2\sqrt8).
\]
Both hidden layers vary with these coordinates. The critic is
$Q_\theta(s,a)=\sqrt{m_Q}\tanh(\varphi(s,a)^\top\theta/\sqrt{m_Q})$.
The five fixed features comprise four region--action indicators and
$a\sqrt{12}(s-i(s)/2)$, with region sign $i(s)\in\{-1,+1\}$.
Thus $m_Q$ varies the critic's nonlinearity while its parameter dimension
remains five. Training samples continuous states without a state grid;
the two-region structure permits separate deterministic reference calculations.

The learned variants use old-coordinate updates, independent random streams
for the update blocks, no critic warmup, and no population targets.
Exact responses, gradients, and TD roots are evaluated after training or
in explicitly identified population studies.

\subsubsection{Connection to the theoretical sample complexity}
\label{app:continuous-scaling-theory}
The ordinate in Figure~\ref{fig:continuous-headline}(a) matches the
stationarity criterion of Theorem~\ref{thm:main-headline}. For a successful
run and an independent index $J$ uniform on $\{0,\ldots,T-1\}$,
\[
 \E_J\!\left[|\Phi'(x_J)|^2\mid x_0,\ldots,x_{T-1}\right]
 =\frac1T\sum_{t=0}^{T-1}|\Phi'(x_t)|^2
 =\overline G_T.
\]
Thus averaging the displayed run-level statistic across training seeds
estimates the theorem's uniform-output squared-gradient criterion;
the prescribed fallback accounts for failed runs as described above.
The theoretical relation
\[
 \E\|\nabla\Phi(x_{\rm out})\|^2=O(\varepsilon),
 \qquad N=\widetilde{\mathcal O}(\varepsilon^{-2})
\]
has leading polynomial dependence $N^{-1/2}$ when expressed as
squared stationarity versus interaction budget, up to logarithmic factors.
This is the reference slope shown in the main-text plot.

The experimental schedule also follows the theorem's powers of the
iteration budget. Writing $\varepsilon_T\asymp T^{-1/2}$ gives
\[
 \eta_x,\eta_y,\beta,c^2,m_Q^{-2}\asymp T^{-1/2},
 \qquad \frac{\Psi_0}{\eta_xT}=O(T^{-1/2}),
\]
for a uniformly bounded initial potential $\Psi_0$.
These are the terms in the stationarity bound
\eqref{eq:outline-final-rate}, with the remaining truncation and
certificate terms controlled at the corresponding accuracy in the theorem.
With fixed batches and geometric rollout lengths of bounded mean,
charged interactions grow proportionally to $T$ in expectation, apart
from the initial metric cost. Accordingly, the fitted slopes near
$-0.505$ and nearly constant $\sqrt N\,\overline G_T$ provide finite-budget
support for the predicted interaction scaling. An eightfold increase in
budget predicts a reduction by approximately $\sqrt8=2.828$ in this
metric; the observed reductions are 2.856 and 2.859.

\subsubsection{Role of the two critic initializations}
\label{app:continuous-initialization-comparison}
The two curves use the same actor and reward initialization, matched
training seeds, interaction budgets, and update schedules. They differ
in the initial critic coordinates. The zero-initialization curve starts
from $u_0=v_0=0$, which represents the exact values at
$x_0=U_0=S_0=0$. The displaced-initialization curve starts from
$\|u_0\|=2$ and $\|v_0\|=4$, so its initial critics are inaccurate.
Both settings subsequently train the critics through sampled TD updates
while the reward and actor parameters evolve.

Theorem~\ref{thm:main-headline} permits any deterministic critic
initialization in $\mathcal K_Q$. Initial critic error contributes to
$\Psi_0$ in~\eqref{eq:outline-final-rate}; with a displacement fixed
across budgets, its contribution $\Psi_0/(\eta_xT)$ retains the
$T^{-1/2}$ dependence. The theorem therefore allows
initialization-dependent constants with the same interaction exponent.
Comparing the two curves checks whether the predicted decay is also
observed when training begins with inaccurate value estimates.
Their similar slopes support this aspect of the guarantee, while the
critic-error measurements in Section~\ref{app:continuous-training}
directly document recovery from the displaced initialization.

\subsection{Critic initialization and recovery}
\label{app:continuous-training}
We evaluate persistent critic learning from zero and displaced initializations
using the full update in both cases. The fixed-budget study uses $T=2000$,
batch size 256 in each sampled block, geometric cap $H=32$, $c=0.05$,
$m_Q=256$, TD step $\beta=0.2$, and actor and upper steps
$\eta_y=\eta_x=0.02$.
The initial metric uses 4722 reset states; critic radii are $R_u=4$ and
$R_v=8$. The two initializations use matched seeds 100--109, with ten runs
per initialization. All 20 runs complete.

At $x_0=U_0=S_0=0$, zero critics represent the exact initial values.
The displaced initialization has $\|u_0\|=2$ and $\|v_0\|=4$, in fixed
directions shared across seeds. Both settings train the critics from the
first iteration, without critic warmup or population targets in the updates.

For evaluation, population TD roots at the current reward and actor
parameters supply $(u^*,v^*)$. The weighted parameter tracking error is
$100\|u-u^*\|^2+\|v-v^*\|^2$, evaluated at 26 saved reference checkpoints
per run. Table~\ref{tab:continuous-training} reports initial and final
seed-averaged errors for the displaced initialization. Value errors are measured against
the true regularized values. The base metric is the squared sup-norm
error; the paired metric is the squared sup-norm of the endpoint-error
difference after division by $c$.

\begin{table}[htbp]
\centering\small
\begin{tabular}{@{}lrr@{}}
\toprule
Critic error metric & Initial mean & Final mean\\
\midrule
Weighted parameter tracking & $416$ & $0.1904$\\
Squared base value error & $5.8796$ & $1.141\times10^{-3}$\\
Squared normalized paired value error & $23.7719$ & $4.437\times10^{-4}$\\
\bottomrule
\end{tabular}
\caption{Critic recovery from displaced initialization. Entries are means
over ten training seeds; population reference calculations are used only
for evaluation.}
\label{tab:continuous-training}
\end{table}

The reductions in both base and normalized paired value errors show
recovery from initially inaccurate critics on this continuous-state
instance while the reward and actor parameters evolve.

\FloatBarrier
\providecommand{\EndToEndAssets}{experiment_results/continuous_state_end_to_end}
\section{End-to-end continuous-state benchmark details}
\label{app:end-to-end-details}
\label{app:implementation}
\label{app:empirical-requirements}
This section describes Walker Walk and Highway-fast. They both have
continuous observations; Highway-fast has a finite action set.
The separate continuous-state theory-aligned study has its own protocol.
These practical SAC experiments evaluate end-to-end learning performance.

\subsection{Cohort, uncertainty, and interaction accounting}
Both figures use 5 training seeds. Each seed contributes
the mean of ten held-out evaluation episodes at each checkpoint.
Lines average these five means; bands show their sample standard deviation
with denominator $5-1$, not a confidence interval or a pooled episode SD.
Each preference run uses 100,000 main interactions and 500 labels.
Walker Walk additionally uses 9,216 auxiliary response interactions per run.
Highway-fast auxiliary interactions are listed below. Known-reward SAC uses
100,000 main interactions and no preference labels or auxiliary responses.
The horizontal axis counts total training interactions: main plus cumulative auxiliary response transitions. For each common evaluation checkpoint, the plotted cost is averaged across the five seeds.

\begin{table}[htbp]
\centering
\caption{Highway-fast auxiliary interactions, in addition to 100,000 main
interactions per run.}
\begin{tabular}{lrrr}
\toprule
Seed & Secant & Gaur et al. & Known-reward SAC\\
\midrule
1 & 1682 & 1800 & 0\\
3 & 1712 & 1607 & 0\\
4 & 1800 & 1734 & 0\\
5 & 1612 & 1591 & 0\\
6 & 1704 & 1578 & 0\\
\bottomrule
\end{tabular}
\end{table}

\begin{figure}[htbp]
\centering
\includegraphics[width=0.49\linewidth]{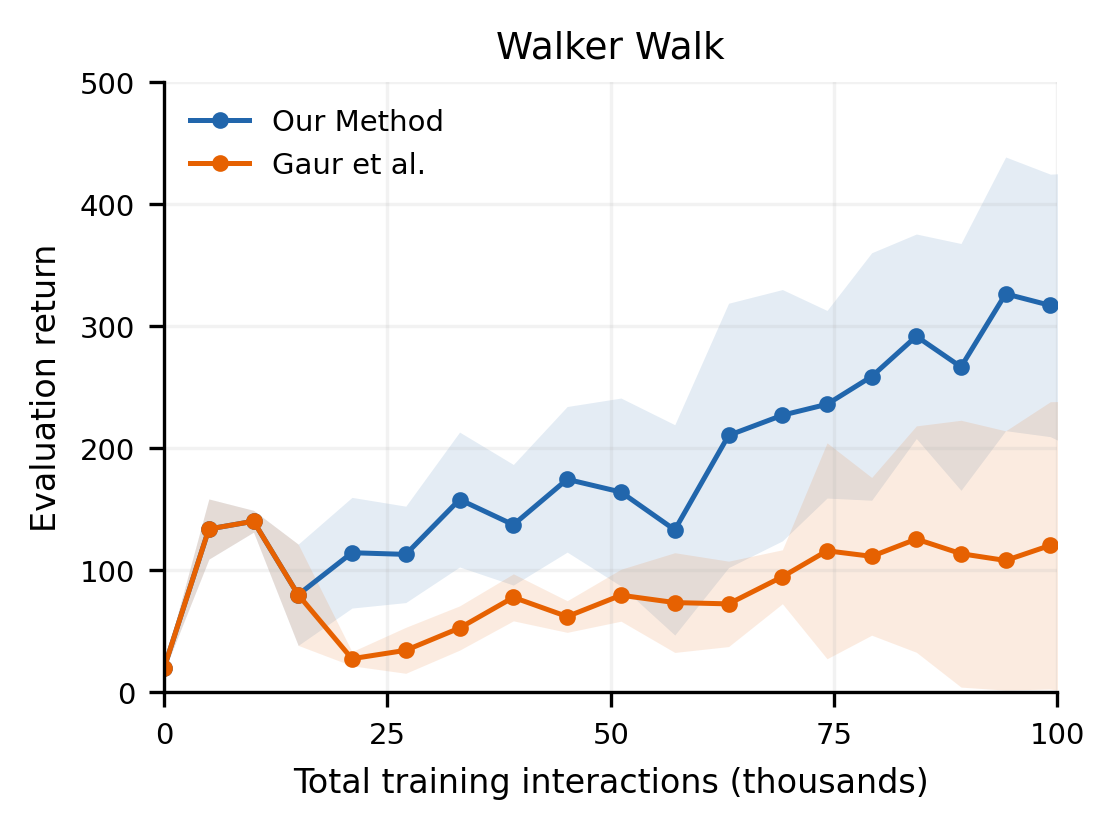}\hfill
\includegraphics[width=0.49\linewidth]{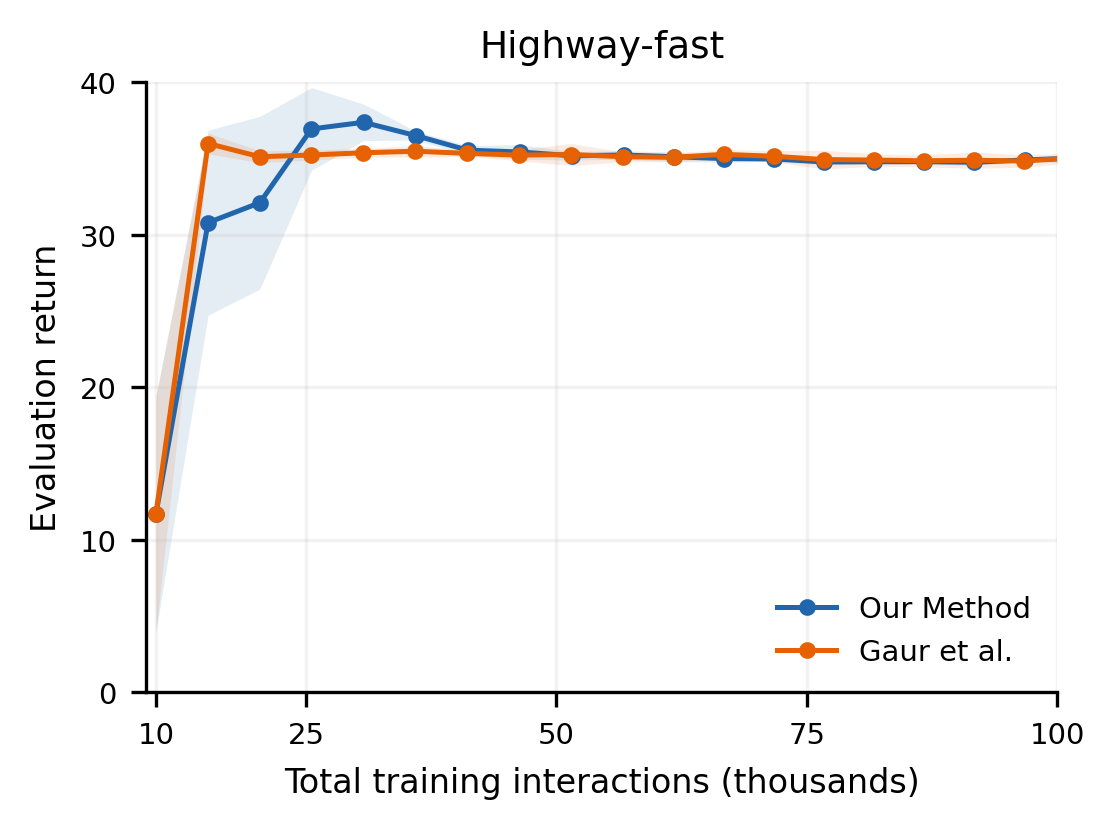}
\caption{End-to-end deterministic evaluation returns: mean and one sample
SD across five matched training seeds, each contributing a ten-episode mean.
The horizontal axis includes cumulative auxiliary response interactions;
checkpoint costs are averaged across seeds. Walker Walk starts at its first
measured checkpoint, and Highway-fast is shown after exploration warm-up.}
\label{fig:end-to-end}
\end{figure}

\subsection{End-to-end  implementation}
\paragraph{Environments.}
Walker Walk uses DeepMind Control, a 24-dimensional observation, a bounded
six-dimensional continuous action, and 1000-step episodes with per-step task
reward in $[0,1]$. Highway-fast uses \texttt{highway-fast-v0}, with 25
continuous observation features: presence, position $(x,y)$, and velocity
$(v_x,v_y)$ for five vehicles. Its five actions are lane left, idle, lane
right, faster, and slower. The configuration uses three lanes, 20 traffic
vehicles, vehicle density 1, policy frequency 1 Hz, simulation frequency
5 Hz, and at most 50 policy steps per episode. Collisions terminate episodes;
\texttt{offroad\_terminal} is false. Task rewards are normalized. Returns are
undiscounted sums of true environment rewards.

\paragraph{Actor and critics.}
Walker Walk uses a tanh-squashed diagonal-Gaussian actor with two 1024-unit
ReLU layers, outputting action means and log standard deviations; the latter
are clamped to $[-5,2]$. Its two Q networks each map concatenated state/action
through two 1024-unit ReLU layers to a scalar. Highway-fast uses a categorical
actor with two 1024-unit ReLU layers and five logits. Each of its two Q
networks maps the state through two 1024-unit ReLU layers to five action
values. Discrete SAC computes the action expectation exactly. Highway
actions are represented by five-dimensional one-hot vectors in the reward
model.

\paragraph{Reward model and labels.}
Both preference methods use an ensemble of three reward networks, each with
three 256-unit LeakyReLU layers and a scalar tanh output. Inputs concatenate
state and action. Reward Adam learning rate is $3\times10^{-4}$; preference
batch size and reward updates per query block are both 50. Segment length is
50 for Walker Walk and 10 for Highway-fast. Synthetic labels use segment
task rewards, $\beta=-1$, $\gamma_{\rm teacher}=1$, zero mistake and skip
probabilities, and equality-threshold coefficient 0.1. This coefficient
scales the teacher's equality margin; it is not an independent probability
of assigning an equal label. Acquisition uses disagreement sampling from a
ten-times-larger candidate batch.

\begin{table}[ht]
\centering
\caption{Common SAC and collection settings. Feedback settings apply to the
two preference methods. Evaluation schedules are specified below.}
\label{tab:sac-hp}
\small
\begin{tabularx}{\linewidth}{@{}Xl@{}}
\toprule
Setting & Value \\
\midrule
Main environment interactions / replay capacity & 100,000 / 100,000 \\
Random / intrinsic warm-up interactions & 1000 / 9000 \\
Discount / SAC minibatch size & 0.99 / 1024 \\
Actor and critic learning rates / Adam betas & $5\times10^{-4}$ / $(0.9,0.999)$ \\
Temperature learning rate / initial temperature & $10^{-4}$ / 0.1 (learned) \\
Actor / critic-target update frequencies & 1 / 2 \\
Target Polyak coefficient & 0.005 \\
Feedback interval / total label budget & 5000 / 500 \\
Critic reset updates after warm-up & 100 \\
Held-out evaluation episodes per checkpoint & 10 \\
Training device & CUDA \\
\bottomrule
\end{tabularx}
\end{table}

\paragraph{Response and upper-update settings.}
Secant uses perturbation $c=0.1$ and chain coefficient 1. Each response
branch receives 100 shadow-SAC updates and supplies two auxiliary rollout
episodes, capped at 256 steps on Walker Walk and 50 on Highway-fast.
The 10k feedback block performs preference bootstrapping; nine subsequent
blocks apply the upper correction, beginning at 15k. Across those blocks
there are 1800 shadow-SAC updates (two branches per block).

The \cite{gaur2025sample} comparator uses the upper objective $G+\widehat V_{\rm base}/\sigma$,
with $\sigma=0.05$ on Walker Walk and $\sigma=0.08$ on Highway-fast. This is the upper level loss used for the empirical results in Sectio F of the work.
The rest of the pipeline, including both response branches, the perturbation
$c=0.1$, their update counts, and auxiliary rollout collection, is retained.
Thus $c$ configures the retained response construction; it is not substituted
for $\sigma$ in the Gaur upper loss. The value contribution uses the base
response's normalized, truncated discounted learned-reward return and is
averaged across reward-ensemble members. The positive sign is retained,
including for negative learned returns. $\sigma$ was chosen via grid search over $[0.01,0.2]$.

\paragraph{Evaluation and display.}
Evaluation uses 10 held-out environment seeds: the Gaussian mean with squashing on Walker Walk and the categorical
argmax on Highway-fast. Walker Walk uses saved actors at 25k, 50k, 75k, and
100k. The Walker curve begins at its first measured checkpoint, 25k main interactions; no artificial origin is included. Highway-fast evaluations are recorded at the first episode
boundary after each 5k milestone (less than 50 steps after the nominal
milestone); the exact step is retained in the data. The figure includes
10k--100k main-interaction checkpoints, displayed at their total training
costs and omitting the earlier exploration period. The preference methods'
shared warm-up produces identical early policies within each seed; their
upper updates first differ at 15k. Lines join available evaluations without
smoothing, jitter, or synthetic intermediate measurements.

\paragraph{Reproduction material.}
The accompanying code regenerates figures and tables from the archived
episode returns and verifies the recorded source hashes. Exact source
versions and the support snapshot are supplied for reconstructing each
historical training run. The code README distinguishes rebuilding reported
figures from rerunning GPU training with the recorded environment.

\FloatBarrier

\FloatBarrier
\end{document}